\documentclass[10pt,twocolumn,letterpaper]{article}
\usepackage[pagenumbers]{cvpr} 
\usepackage{listings}
\usepackage{xspace}
\usepackage{multicol}
\usepackage{comment}

\usepackage[hang,flushmargin]{footmisc} 

\newcommand{\algComment}[1]{{~~\textit{\color{black!35}#1}}}
\newcommand{\algRed}[1]{{{\color{red!85!black}#1}}}
\renewcommand{\paragraph}[1]{\vspace{2pt}\noindent\textbf{#1}\;} 
\usepackage[skip=1pt plus2pt, indent=11pt]{parskip}
\usepackage{tabularx}
\usepackage{makecell}
\usepackage{multirow}
\usepackage{booktabs}
\usepackage{siunitx} 
\usepackage{fbox}
\usepackage{overpic}
\usepackage{floatpag} 
\usepackage{pifont} 
\usepackage{array} 
\usepackage{ragged2e}
\usepackage{wrapfig}

\def\bx{{\boldsymbol{x}}}
\def\bX{{\mathbf{X}}} 
\def\by{{\boldsymbol{y}}}

\def\btheta{{\boldsymbol{\theta}}}

\def\bpsi{{\boldsymbol{\psi}}}

\newcommand{\modulo}{\bmod\,}
\makeatletter
\newcommand{\shorteq}{\mathrel{\mkern0.2mu\mathpalette\shorteq@\relax\mkern0.2mu}}
\newcommand{\shorteq@}[2]{\scalebox{0.5}[1]{$\m@th#1=$}}

\makeatother
\usepackage{bbold} 
\usepackage{nicefrac}
\newcommand\lldots{\ifmmode$\lldots$\else\thinspace\makebox[1em][c]{.\hfil.\hfil.}\fi} 
\newcommand{\mnist}{{\sc{mnist}}\xspace}
\newcommand{\fmnist}{{\sc{fashion-mnist}}\xspace}
\newcommand{\fashionm}{{\sc{fashion-m.}}\xspace}
\newcommand{\cifar}{{\sc{cifar}}\xspace}
\newcommand{\cifarTen}{{\sc{cifar-10}}\xspace}
\newcommand{\svhn}{{\sc{svhn}}\xspace}

\usepackage{algorithm}

\definecolor{customblue}{HTML}{1A1AFF}
\definecolor{customred}{HTML}{FF4D4D}
\definecolor{darkYellow}{rgb}{0.8, 0.6, 0}

\newcommand{\coloredBullet}{\raisebox{.85pt}{\tikz \node[draw=customred, fill=customblue, circle, inner sep=0.7pt] {};}}

\usepackage[framemethod=TikZ]{mdframed}
\definecolor{lightGray}{gray}{0.97}
\definecolor{midGray}{gray}{0.40}
\definecolor{lightYellow}{RGB}{254,254,233}
\definecolor{darkGray}{rgb}{0.45,0.45,0.45}
\definecolor{darkerGray}{rgb}{0.3,0.3,0.3}
\mdfdefinestyle{citationFrame}{%
  linecolor=midGray,
  linewidth=0.5pt,
  roundcorner=3pt,
  skipabove=5pt,
  skipbelow=0pt,
  innertopmargin=5pt,
  innerbottommargin=5pt,
  innerrightmargin=7pt,
  innerleftmargin=7pt,
  leftmargin=0pt,
  rightmargin=0pt,
  backgroundcolor=lightYellow
}

\mdfdefinestyle{takeawayFrame}{%
  linecolor=midGray,
  linewidth=0.5pt,
  roundcorner=3pt,
  skipabove=7pt,
  skipbelow=0pt,
  innertopmargin=5pt,
  innerbottommargin=5pt,
  innerrightmargin=7pt,
  innerleftmargin=7pt,
  leftmargin=0pt,
  rightmargin=0pt,
  backgroundcolor=lightYellow,
  userdefinedwidth=\linewidth
}

\makeatletter
\def\NAT@def@citea{\def\@citea{\NAT@separator\,}} 
\makeatother

\definecolor{cvprblue}{rgb}{0.21,0.49,0.74}
\definecolor{myGray}{rgb}{0.85,0.85,0.85}
\usepackage[pagebackref,breaklinks,colorlinks,citecolor=cvprblue]{hyperref}
\usepackage{colortbl} 

\def\paperID{2199}
\def\confName{CVPR}
\def\confYear{2025}
\title{Do We Always Need the Simplicity Bias?\\Looking for Optimal Inductive Biases in the Wild}

\author{%
    Damien Teney\\
    Idiap Research Institute
    \and
    Liangze Jiang\\
    EPFL 
    \and
    Florin Gogianu\\
    Bitdefender
    \and
    Ehsan Abbasnejad\\
    University of Adelaide\vspace{5pt}
}

\begin{document}

\maketitle

\setlength{\textfloatsep}{12pt plus 0pt minus 4pt} 

\begin{abstract} 
Neural architectures tend to fit their data with relatively simple functions.
This ``simplicity bias'' is widely regarded as key to their success. 
This paper explores the limits of this principle.
Building on recent findings that the simplicity bias stems from ReLU activations~\cite{teney2024neural},
we introduce a method to meta-learn new activation functions
and inductive biases better suited to specific tasks.

\vspace{1pt}
\noindent\textbf{Findings.}
We identify multiple tasks where the simplicity bias is inadequate and ReLUs suboptimal.
In these cases, we learn new activation functions that perform better
by inducing a prior of higher complexity.
Interestingly, these cases correspond to domains where neural networks have historically struggled:
tabular data,
regression tasks,
cases of shortcut learning,
and algorithmic grokking tasks.
In comparison, the simplicity bias
induced by ReLUs
proves adequate on image tasks
where the best learned activations are nearly identical to ReLUs and GeLUs.

\vspace{1pt}
\noindent\textbf{Implications.}
Contrary to popular belief, the simplicity bias
of ReLU networks
is not universally useful.
It is near-optimal for image classification,
but other inductive biases are sometimes preferable. 
We showed that activation functions can control these inductive biases,
but future tailored architectures might provide further benefits.
Advances are still needed
to 
characterize a model's inductive biases
beyond ``complexity'',
and their adequacy with 
the data.
\end{abstract}


\section{Introduction}
\label{sec:intro}

\paragraph{When and why NNs generalize is yet to be understood.}
Neural networks (NNs) have proven more effective than other machine learning models.
However, we still miss a complete explanation of their generalization abilities. 
A better understanding 
could help
address 
failures from
shortcut learning~\cite{geirhos2020shortcut,teney2021evading}
to distribution shifts~\cite{jiang2024ood,teney2024id},
biases, and spurious correlations in language models for example~\cite{gallegos2024bias, harutyunyan2024context, singhal2023long}.
Understanding 
conditions for generalization would also enable the design of architectures
and data preparation from first principles, rather than trial and error.


\begin{figure}[t]
  \centering
  \includegraphics[width=\columnwidth]{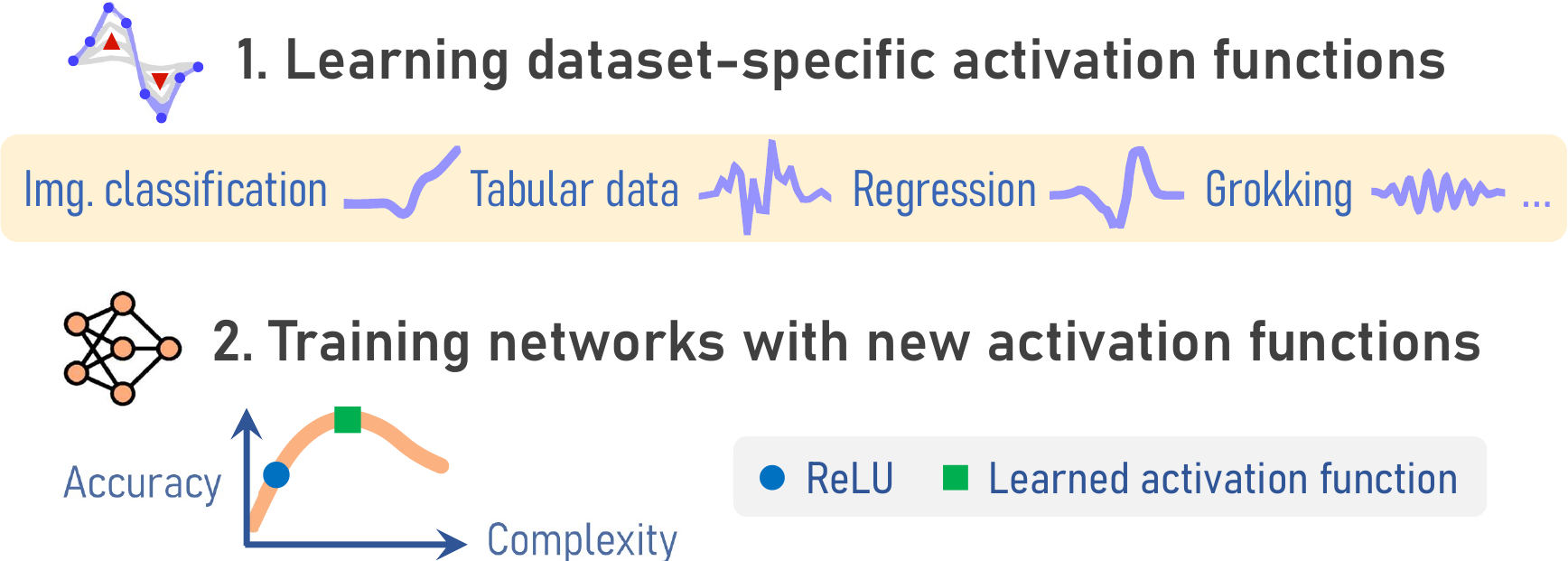}
  \vspace{-8pt}
  \caption{\label{fig:teaser}%
  (1)~We modulate the inductive bias of neural architectures by learning novel activation functions  that improve generalization on specific datasets.
  (2)~With this tool, we study the relation between model accuracy and complexity.
  We identify tasks where the simplicity bias of ReLU architectures is suboptimal.}
  \vspace{3pt}
\end{figure}

\paragraph{This paper studies inductive biases}
i.e.\ the assumptions made by learning algorithms to generalize beyond training data~\cite{mitchell1980need}.%
\footnote{Inductive biases can formalized as a prior over the space of functions~\cite{mingard2021sgd}.}
A vast literature examines the inductive biases of
architectures~\cite{cohen2016inductive},
optimizers~\cite{neyshabur2014search},
losses~\cite{hui2020evaluation},
regularizers~\cite{kukavcka2017regularization}, etc.
The \textbf{simplicity bias}
is one aspect of the inductive biases of NNs
that makes them fit their training data
with simple%
\footnote{Simplicity
can be
formalized using
Kolmogorov complexity 
or its approximations, e.g.\
frequency, compressibility, sensitivity, etc.~\cite{teney2024neural,dingle2018input,hahn2021sensitivity,dherin2022neural}}
functions~\cite{arpit2017closer,poggio2018theory}.
Despite wide belief that the simplicity bias could be due to
SGD~\cite{arora2019implicit,hermann2020shapes,lyu2021gradient,tachet2018learning},
work on untrained networks showed that
it can be explained with architectures alone~\cite{de2019random,goldblum2023no,mingard2019neural,valle2018deep}.
ReLUs also
seem critical to induce the simplicity bias in typical architectures~\cite{teney2024neural}.

\paragraph{Limits of the simplicity bias.}
The simplicity bias is an intuitive explanation
for the ability to generalize on real-world data.
It embodies
Occam's razor~\cite{mingard2023deep}
and assumes that data-generating processes in the real world are simple.
Additionally, a prior for simplicity is supported by results in algorithmic information theory~\cite{dingle2018input}
stating essentially that
``\textit{a bias in the distribution of target functions must be towards \underline{low} complexity}''.
However, this only means that simplicity is a good prior on average,
but not necessarily the best choice on any task or dataset.
This matches the no-free lunch theorem~\cite{wolpert2002supervised}
according to which
no inductive bias is universally useful.
Therefore, this paper asks the following.

\begin{mdframed}[style=citationFrame,userdefinedwidth=
\linewidth,align=center,skipabove=5pt,skipbelow=0pt]
    \bfseries
    Are there practical applications of machine learning where the simplicity bias is detrimental?
    In these cases, what do the optimal inductive biases look like?
\end{mdframed}\vspace{-3pt}
For example, shortcut learning is one situation where the simplicity bias is already known to be detrimental~\cite{shah2020pitfalls,teney2021evading}.

\paragraph{Searching for optimal inductive biases by learning activation functions.}
Prior work~\cite{teney2024neural} showed that
ReLU activations are critical to obtain the simplicity bias in typical architectures.
Hence we build a new tool to modulate the simplicity bias 
by learning dataset-specific activation functions.
It uses bi-level optimization and a spline parametrization to learn activations free of any prior,
such as constraints of smoothness or monotonicity
(unlike prior work~\cite{alexandridis2024adaptive,apicella2021survey,scardapane2019kafnets}).
This 
(1)~enables the discovery of entirely new activation functions
and inductive biases
that improve generalization (\hyperref[fig:teaser]{Figure~\ref{fig:teaser}})
and
(2)~highlights the suboptimality of the simplicity bias by comparing the
accuracy and complexity of models with ReLUs vs.\ learned activations.

\paragraph{Findings.}
We examine four domains that we hypothesized to be impaired by the simplicity bias:
tabular data, regression tasks, cases of shortcut learning, and algorithmic tasks.
Our intuition is that they require learning 
functions with high sensitivity or sharp transitions.
For each domain, we collect existing datasets
then train and analyze models without and with learned activation functions.
In all cases, we obtain better generalization with dataset-specific activations,
and the improvements are attributable to learning higher-complexity solutions.
In comparison, this analysis also shows that classical image datasets
(\mnist, \cifar, \fmnist, \svhn)
are extremely well suited to the inductive biases of ReLUs.
The best learned activations 
are then strikingly similar to variants like GeLUs~\cite{hendrycks2016gaussian}.

\paragraph{Summary of contributions.}
\begin{itemize}[itemsep=2pt,topsep=0pt]
\item A new method to discover dataset-specific activation functions
optimized for generalization.

\item 
An examination of~$\,>$20 datasets
showing that the simplicity bias of ReLU architectures can be suboptimal.
(1)~For \textbf{\hyperref[sec:regression]{regression}} tasks and \textbf{\hyperref[sec:tabular]{tabular}} data, new learned activations greatly improve accuracy
by helping learn complex functions.
(2)~For \textbf{\hyperref[sec:img]{image classification}}, 
the process rediscovers smooth variants of ReLUs,
suggesting a near-optimal choice
for these popular tasks.
(3)~In cases of \textbf{\hyperref[sec:shortcut]{shortcut learning}},
we show that different learned activations 
can steer the learning towards different image features.
(4)~For 
\textbf{\hyperref[sec:grokking]{grokking}} tasks, new learned activations can eliminate the phenomenon,
supporting the explanation as a mismatch between data and architectures.
We also measure a positive \textbf{\hyperref[sec:transfer]{transferability}} of learned activations across related tasks.

\item An analysis showing that improvements
with learned activations
correlate with the learning of complex functions.

\end{itemize}

\paragraph{Implications.}
All cases where the simplicity bias proved suboptimal are in domains where NNs have historically struggled.
We now connect them to a common explanation.
This implies that architectures tailored to some specific
domains may still have a place besides scaling up models and data.
Conversely, the suitability of ReLUs to image classification
suggests that researchers successfully
converged by trial and error to designs well tuned to popular tasks.

\section{Methods}
\label{sec:method}

This section introduces tools to analyze trained models
and to learn new dataset-specific activation functions.
\subsection{Visualizing a Model's Function}
\label{sec:paths}
A neural network
implements
a function 
$f_\btheta:\mathbb{R}^{d_\textrm{in}}\!\rightarrow\!\mathbb{R}^{d_\textrm{out}}$
of parameters $\btheta$ (weights and biases)
that maps an input $\bx\in\mathbb{R}^{d_\textrm{in}}$
to an output $\by\in\mathbb{R}^{d_\textrm{out}}$.
For a regression task, $\by\!\in\!\mathbb{R}$ is the predicted value.
For a classification task, $\by$ is a vector of logits
passed through a sigmoid or softmax to obtain class probabilities.
Because $d_\textrm{in}$ can be large, $f$ can be difficult to visualize and analyze.
A workaround is to examine $f$ over 1D or 2D slices of the input space~%
\cite{fridovich2022spectral,teney2024neural}.
To obtain a slice in a region of plausible data,
we use the training data~$\mathcal{T}$.
For a 1D slice (linear path),
we sample
$\bx_1,\bx_2\sim\mathcal{T}$
then define the path
$\bX_{\bx_1,\bx_2} = [\, (1\!-\!\lambda)\,\bx_1 + \lambda\,\bx_2, \;\lambda \in [0,1]\,]$.
We proceed analogously with three points for a 2D slice.
We sample $\lambda$ regularly in [0,1] such that
$\bX$ is a finite sequence of points.
Then~$f$ is evaluated on these points
to give a 1D sequence or 2D grid of values that are convenient to display and analyze 
(\hyperref[fig:tabBoundaries]{Figure~\ref{fig:tabBoundaries}c}).
When~$d_\textrm{out}\!>\!1$ (multi-class task), we examine one random dimension of $f$'s output at a time.

\subsection{Measuring a Model's Complexity}
We wish to quantify the complexity of the function~$f$
implemented by a model trained on data~$\mathcal{T}$.
Prior work 
used Fourier decompositions~\cite{fridovich2022spectral,teney2024neural}
but this requires a delicate 
implementation.
We found a reliable alternative with the total variation (TV) of $f$
averaged over many 1D paths:%
\setlength{\abovedisplayskip}{2pt}
\setlength{\belowdisplayskip}{2pt}
\begin{equation}\label{eq:tv1}
    \operatorname{TV}(f, \mathcal{T}) ~=~
    {\mathbb{E}}_{\bx_1,\bx_2\;\sim\mathcal{T}}
    \int_{\bx_1}^{\bx_2} \big| f'(\bx) \big| \;d\bx \;.
\end{equation}
with $f'$ the first derivative.
We estimate (\ref{eq:tv1}) using a path as defined in \hyperref[sec:paths]{Section~\ref{sec:paths}}.
We name the points in such a path 
$\bX_{\bx_a,\bx_z} := [ \bx_a, \bx_b, \bx_c, ... \, \bx_y, \bx_z ]$. We then have: 
\setlength{\abovedisplayskip}{6pt}
\setlength{\belowdisplayskip}{6pt}
\begin{equation}
\begin{split}
    \label{eq:tv2}
    \operatorname{TV}(f, \mathcal{T}) ~\approx~
    {\mathbb{E}}_{\bx_a,\bx_z\;\sim\mathcal{T}}~~
    & ~| f(\bx_b) - f(\bx_a) |\\[-3pt]
    & \!\!\!\!\!+ | f(\bx_c) - f(\bx_b) | \, + \,\ldots\\[-3pt]
    & \!\!\!\!\!+ | f(\bx_z) - f(\bx_y) |\,.
\end{split}
\end{equation}
Appendix~\ref{sec:tv} shows that (\ref{eq:tv2}) correlates closely with
a Fourier-based measure of complexity:
the higher the TV, the higher the complexity.
Yet, it is straightforward to implement
and discriminative across small and large values.

\subsection{Meta-Learning Activation Functions}
Our goal is to optimize the inductive biases of a neural network
and recent work~\cite{teney2024neural} showed
that the activation functions are the most important component.
The typical approach to learn
activations~\cite{alexandridis2024adaptive,apicella2019simple,apicella2021survey,bingham2020evolutionary,chelly2024trainable,ducotterd2024improving,jagtap2020adaptive,scardapane2019kafnets,sutfeld2020adaptive} (see \hyperref[sec:relatedWork]{Related Work})
replaces them with a small shared ReLU MLP
that implements
an $\mathbb{R}\!\rightarrow\!\mathbb{R}$ function.
Its parameters are 
optimized along the network's.
However this cannot discover truly novel activations
because the embedded ReLU MLP has itself a simplicity bias
and activations are optimized together with the model.
We propose instead:
\begin{itemize}[itemsep=1pt,topsep=-1pt]
\item[-] an \textbf{unbiased} parametrization of the activations as splines,
\item[-] a bi-level optimization to learn \textbf{reusable} activations,
\item[-] an episodic training to \textbf{optimize for generalization} rather than
simply to fit the training data.
\end{itemize}

\phantomsection
\label{sec:splines}
\paragraph{Parametrization as splines.}
We want a space of activation functions free of priors
such as the smoothness and monotonicity enforced in prior work~\cite{apicella2019simple,chelly2024trainable}.
We implement an activation 
$g_\bpsi:\mathbb{R}\!\rightarrow\!\mathbb{R}$
as a linear spline
with control points defined by $\bpsi$.
We define $n_\textrm{c}$ points spread regularly in an interval $[a,b]$,
typically $\sim\!50$ points in $[-5,+5]$.
Then $g$ represents piecewise linear segments
interpolating values specified in the learned parameters
$\bpsi := [\,g_\bpsi(a), \ldots g_\bpsi(b))\,] \in\mathbb{R}^{n_\textrm{c}}$.
$g$ can represent simple and complex functions,
including smooth curves, periodic functions, sharp transitions, etc.

\paragraph{Bi-level optimization \& episodic training.}
Our goal is to get an activation function that can be
reused like any other in subsequent training runs.
This differs from prior work (e.g.\ \cite{alexandridis2024adaptive})
that continuously updates the activation during training:
the final one may not be suitable to start training with.
Our solution is a bi-level meta-learning loop.
An inner loop trains the model with a fixed activation function. 
An outer loop trains the activation function to maximize generalization.
Each outer step simulates a new learning task or \textit{episode}.
This means (1)~initializing the model with different weights and
(2)~using different subsets of data for training and validation.
With suitable choices, this can simulate
in-~or out-of-distribution conditions (see \hyperref[sec:shortcut]{Section~\ref{sec:shortcut}}).
Without episodes, the learned activation could overfit to a particular model initialization for example, and would not generalize 
in subsequent training runs.
The method is outlined as \hyperref[alg]{Algorithm~\ref{alg}}.
Its implementation is
discussed in \hyperref[sec:learningAfsDetails]{Appendix~\ref{sec:learningAfsDetails}}.

\newenvironment{algOuter}[1][]{%
  \mdfsetup{%
     frametitle={%
       \tikz[baseline=(current bounding box.east),outer sep=0pt]
        \node[anchor=east,rectangle,fill=black!2]
        {\strut #1};}}%
   \mdfsetup{innertopmargin=-5pt,linecolor=black!50,
             innerleftmargin=1pt,innerrightmargin=1pt,
             leftmargin=0pt,rightmargin=0pt,
             backgroundcolor=black!2,
             linewidth=0.5pt,topline=true,
             frametitleaboveskip=\dimexpr-\ht\strutbox\relax,}
   \begin{mdframed}[]\relax%
   }{\end{mdframed}}

\newenvironment{algInner}[1][]{%
  \mdfsetup{%
     frametitle={%
       \tikz[baseline=(current bounding box.east),outer sep=0pt]
        \node[anchor=east,rectangle,fill=black!2]
        {\strut #1};}}%
   \mdfsetup{innertopmargin=-5pt,linecolor=black!50,
             innerleftmargin=1pt,innerrightmargin=1pt,
             leftmargin=5pt,rightmargin=5pt,
             backgroundcolor=black!2,
             linewidth=0.5pt,topline=true,
             frametitleaboveskip=\dimexpr-\ht\strutbox\relax,}
   \begin{mdframed}[]\relax%
   }{\end{mdframed}}

\begin{mdframed}[style=citationFrame,userdefinedwidth=
\linewidth,align=center,skipabove=6pt,skipbelow=0pt]
    \begin{wrapfigure}{l}{0.4\baselineskip}
      \vspace{-\baselineskip} 
      \vspace{-22pt}\includegraphics[trim=-15pt 405pt 0pt -600pt, height=2.1\baselineskip]{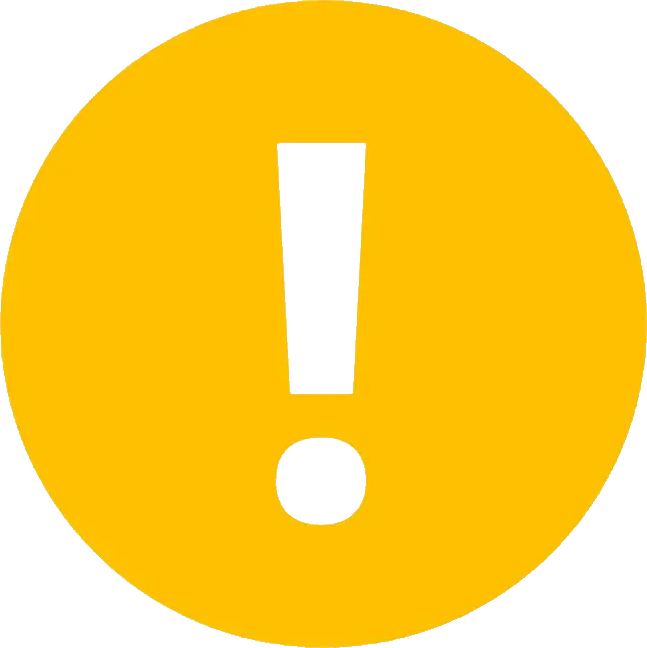}
    \end{wrapfigure}
    \textit{Inductive bias} and \textit{simplicity bias} are not
    interchangeable.
    Our method
    optimizes toward better {generalization}.
    {Simplicity} is only one aspect of
    the trained models that we analyze post-hoc
    (e.g.\ \hyperref[fig:imgPlots]{Figure~\ref{fig:imgPlots}}).
\end{mdframed}\vspace{-4pt}

\setlength{\textfloatsep}{0pt} 

\begin{algorithm}[th]
    \caption{Meta-learning an activation function (AF).}
    \label{alg}
    \vspace{3pt}
    \phantom{I}\textbf{Input}: training data $\mathcal{T}$; untrained neural model $f_{\btheta,\bpsi}$\\[2pt]
    \phantom{I}Initialize $\bpsi$ with zeros \algComment{Parametrization of AF}\\[1pt]
    \phantom{I}$n_\textrm{tr} \leftarrow 0$ \algComment{Number of inner-loop iterations}
    \vspace{3pt}
    \begin{algOuter}[\textbf{while} $n_\textrm{tr} < n_\textrm{tr}^\textrm{max}$%
    \algComment{Outer loop: train AF}]
        \phantom{I}Increment $n_\textrm{tr}$\\[2pt]
        \phantom{I}Sample the episode's tr. ($\mathcal{T}'$) and val.\ ($\mathcal{V}$) sets from~$\mathcal{T}$%
        \\[3pt]
        \phantom{I}Initialize $\btheta$ randomly \algComment{Model weights and biases}
        \vspace{1pt}
        \begin{algInner}[\textbf{for} $n_\textrm{tr}$ steps \algComment{Inner loop: train model with fixed AF}]
            \phantom{I}Eval.\ loss on~\algRed{$\mathcal{T}'$}: $L\leftarrow
            \Sigma_{(\bx,\by)\,\algRed{\in\mathcal{T}'}} \,
            \mathcal{L}\big(
            f_{\btheta,\bpsi}(\bx, \by)
            \big)$\\[3pt]
            \phantom{I}Gradient step on weights/biases:
            $\btheta \leftarrow\operatorname{GD}(\btheta, \nabla_\btheta L)$
            \vspace{1pt}
        \end{algInner}
        \vspace{2pt}
        \phantom{I}Eval.\ loss on~\algRed{$\mathcal{V}$}:\,
            $L\leftarrow
            \Sigma_{(\bx,\by)\,\algRed{\in\mathcal{V}}} \,
            \mathcal{L}\big(
            f_{\btheta,\bpsi}(\bx, \by)
            \big)$\\[2pt]
        \phantom{I}Gradient step on AF:\,
        $\bpsi \leftarrow\operatorname{GD}(\bpsi, \nabla_\bpsi L)$\\[2pt]
        \phantom{I}\textbf{if} performance on $\mathcal{V}$ worsens \textbf{then} break\algComment{Early stopping}
    \end{algOuter}
    \phantom{I}\textbf{Output}: optimized AF $\bpsi$
    \vspace{2pt}
\end{algorithm}

\section{Tasks and Results}
\label{sec:tasks}

We now examine tasks that we hypothesized to be
ill-suited to the simplicity bias of ReLU architectures.
The intuition is that the target function to learn
(e.g.\ optimal classifier)
contains
sharp transitions (regression tasks, tabular datasets),
or repeating patterns (algorithmic tasks) that contradict the ReLUs' simplicity bias.
For each task, we examine existing datasets
with the tools from \hyperref[sec:method]{Section~\ref{sec:method}}.
In all cases, we find
benefits from architectures whose inductive biases favor more complex functions.
Additional details and results are provided in Appendix~\ref{sec:results2}.

\vspace{2pt}
\subsection{Image Classification Tasks}
\vspace{-2pt}
\label{sec:img}
\paragraph{Background.}
We start with classical datasets
to validate our methodology:
\mnist, \fmnist, \svhn, \cifarTen
\cite{lecun1998gradient,krizhevsky2009learning,netzer2011reading,xiao2017fashion}.
They are representative of the vision tasks
that guided the development of deep learning.
\textbf{Our hypothesis} is therefore that the inductive biases of modern architectures
and ReLUs are well suited to these datasets.

\paragraph{Setup.}
For each dataset,
we learn 
activation functions with \hyperref[alg]{Algorithm~\ref{alg}}.
We experiment with two initializations of the spline parameters: as zeros and so as to mimic a ReLU.
The goal of the latter is to explore the space of functions similar to ReLUs.
Because of the difficulty of the optimization, 
the algorithm is likely to converge to a \emph{local} optimum similar to ReLUs if there is one.
We also experiment with the sharing of the activation function.
By default, a single function is shared across the network.
Alternatively, we learn a different activation function per layer.
This provides more ways to affect the model's inductive biases.
Our base architecture is a 3-layer MLP
(details in Appendix~\ref{sec:expDetails}).

\begin{figure}[t!]
  \vspace{-8pt}
  \centering
  \renewcommand{\tabcolsep}{0em}
  \renewcommand{\arraystretch}{1}
  \scriptsize
  \begin{tabular}{ccl}
    \scriptsize~~~~(a)~\textbf{Image classification}&
    \scriptsize~~~(b)~\textbf{Image regression}&\\[-0pt]
    \makecell[t]{\raisebox{.00\columnwidth}{
      \begin{overpic}[height=0.25\columnwidth, trim=0pt 45.5pt 0pt 0pt, clip]{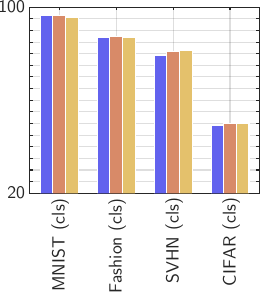}
        \put(6, 38){\makebox(0,0)[r]{\rotatebox{90}{\scriptsize{Accuracy (\%)}}}}
      \end{overpic}
    }
    }&
    \makecell[t]{
      \begin{overpic}[height=0.251\columnwidth, trim=0pt 47.5pt 0pt 0pt, clip]{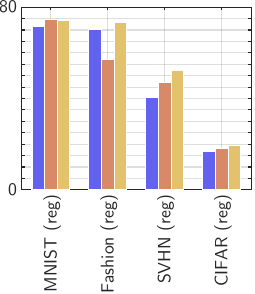}
      \end{overpic}
    }&
    \makecell[bl]{
      ~~\scriptsize\textcolor[HTML]{6161EF}{\raisebox{0.0em}{\scalebox{1.2}[1]{\ding{110}}}}
      ReLU activations\\
      ~~\scriptsize\textcolor[HTML]{E68F69}{\raisebox{0.0em}{\scalebox{1.2}[1]{\ding{110}}}}
      Learned act., ReLU init.\\
      ~~\scriptsize\textcolor[HTML]{F3CC6E}{\raisebox{0.0em}{\scalebox{1.2}[1]{\ding{110}}}}
      Learned act., zero init.\\\vspace{.045\columnwidth}
    }\\[-2pt]
    \rotatebox{45}{\tiny MNIST}~
    \rotatebox{45}{\tiny Fashion}~
    \rotatebox{45}{\tiny SVHN}~
    \rotatebox{45}{\tiny CIFAR}&
    \rotatebox{45}{\tiny MNIST}~
    \rotatebox{45}{\tiny Fashion}~
    \rotatebox{45}{\tiny SVHN}~
    \rotatebox{45}{\tiny CIFAR}
  \end{tabular}
  \vspace{-9pt}
  \caption{\label{fig:imgActBars}
    Test accuracy on image datasets.
    (a)~For \textbf{classification} tasks, all models
    perform similarly, suggesting that the inductive biases of ReLUs are well suited to these datasets.
    (b)~For \textbf{regression} tasks, models with learned activations perform better, especially from
    an initialization as zeros, which enables the discovery of completely novel activation functions.
  }
  \vspace{-6pt}
\end{figure}

\paragraph{Results.}
We compare in \hyperref[fig:imgActBars]{Figure~\ref{fig:imgActBars}a} 
the accuracy of models with ReLUs vs.\ learned activation functions.
Differences are small.
The learned activations only improve slightly on \svhn and \cifar.
This suggests that the inductive biases of ReLUs
are generally well suited to these datasets.

\setlength{\textfloatsep}{12pt plus 0pt minus 4pt}

\begin{figure}[t!]
  \renewcommand{\tabcolsep}{0.0em}
  \renewcommand{\arraystretch}{1.1}
  \scriptsize
  \begin{tabular}{rcccc}
    ~ & \textcolor{darkGray}{ReLU} & \multicolumn{3}{c}{\textcolor{darkGray}{Learned activation functions}}\\[-1pt]
    ~ & \textcolor{darkGray}{activations} & \textcolor{darkGray}{ReLU init.} & \textcolor{darkGray}{Zero init.} & \textcolor{darkGray}{Layer-specific}\\[2pt]
    \makecell[br]{\textcolor{darkGray}{(a)} \mnist as a~~~~\\\bfseries classification~~~~\\task~~~~\\\vspace{0.48em}}&
    \makecell{\includegraphics[height=.185\columnwidth]{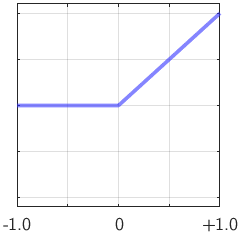}}&
    \makecell{\includegraphics[height=.185\columnwidth]{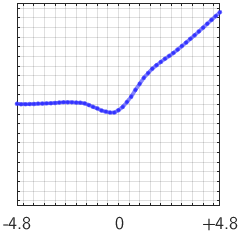}}& 
    \makecell{\includegraphics[height=.185\columnwidth]{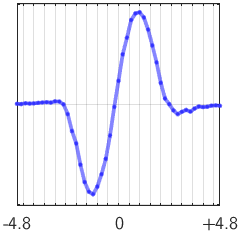}}&
    \makecell{\includegraphics[height=.185\columnwidth]{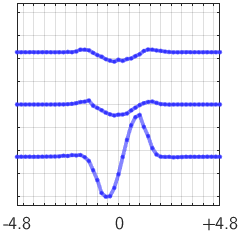}}\\[-1pt] 
    ~ & \multicolumn{4}{c}{
        \begin{tikzpicture}\draw[-, color=darkGray] (0,0) -- (0.74\columnwidth,0) node[midway, fill=white, text=darkerGray] {\textbf{\textit{Similar accuracy}}};\end{tikzpicture}}\\[5pt]
    \makecell[br]{\textcolor{darkGray}{(b)} \mnist as a~~~~\\\bfseries regression~~~~\\ task~~~~\\\vspace{0.48em}}&
    \makecell{\includegraphics[height=.185\columnwidth]{visualizations-mnist-ranged-cropped-bestExps/af-file0001-100-1-te-sn0-0032-nTr1.00.png}}&
    \makecell{\includegraphics[height=.185\columnwidth]{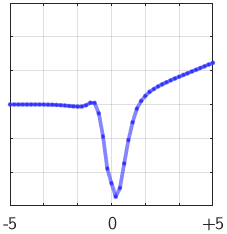}}& 
    \makecell{\includegraphics[height=.185\columnwidth]{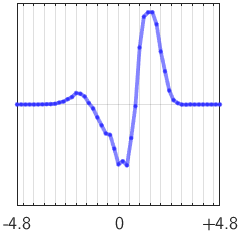}}&
    \makecell{\includegraphics[height=.185\columnwidth]{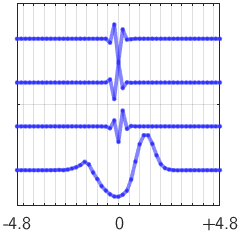}}\\[-1pt] 
    ~ & \multicolumn{4}{c}{
        \begin{tikzpicture}\draw[->, color=darkGray] (0,0) -- (0.74\columnwidth,0) node[midway, fill=white, text=darkerGray] {\textbf{\textit{Increasing accuracy}}};\end{tikzpicture}}
  \end{tabular}
  \vspace{-10pt}
  \caption{\label{fig:imgAct}
  Activation functions learned for \mnist.
  For a \textbf{classification} task, the activation learned from a ReLU 
  resembles the popular GeLUs. 
  For a \textbf{regression} task, the learned activations contain irregularities
  that help a network represent complex functions.
  See \hyperref[fig:imgActAll]{Figure~\ref{fig:imgActAll}}
  for similar results on other datasets.
  }
\end{figure}

\noindent
We examine the learned activations in \hyperref[fig:imgAct]{Figure~\ref{fig:imgAct}a}.
With an initialization as ReLUs,
the optimization converges to a smooth variant remarkably similar to
GeLUs~\cite{bingham2020evolutionary}
which are widely used.
This suggests that the research community has empirically converged on a local optimum
in the space of activation functions.
With an initialization as zeros,
we discover wavelets~\cite{saragadam2023wire}
that are 
unlike common activations but perform as well as ReLUs, i.e.\ another local optimum.

\begin{mdframed}[style=takeawayFrame]
\textbf{Take-away:}
for image classification, learned activations provide very little 
benefit over ReLUs.
Smooth variants of ReLUs are a local optimum in the space of activations.
ReLUs' popularity for such tasks could thus be explained with their proximity to this optimum.
\end{mdframed}\vspace{-3pt}

\begin{figure}[t!]
  \vspace{-8pt}
  \centering
  \scriptsize
  \renewcommand{\tabcolsep}{0.9em}
  \renewcommand{\arraystretch}{1}
  \begin{tabular}{cc}
    \textcolor{darkGray}{(a)} \mnist as \textbf{classification} task &
    \textcolor{darkGray}{(b)} \mnist as \textbf{regression} task \\[2pt]
    \makecell{\includegraphics[height=.16\textwidth]{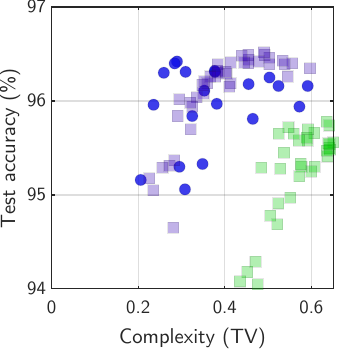}} &
    \makecell{\includegraphics[height=.16\textwidth]{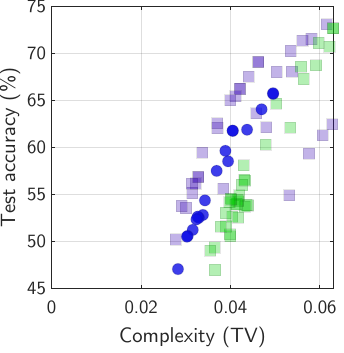}} 
  \end{tabular}
  \vspace{-10pt}
  \caption{
    \label{fig:imgPlots}
    Accuracy vs.\ complexity 
    on image datasets.
    Each marker is a model with different hyperparameters 
    and ReLUs\;%
    (\textcolor[HTML]{3D3DEB}{\raisebox{0.0em}{\scalebox{0.75}[0.75]{\ding{108}}}})
    or learned activations initialized as ReLUs\;%
    (\textcolor[HTML]{C1B4E8}{\raisebox{0.0em}{\scalebox{0.75}[0.75]{\ding{110}}}})    
    or as zeros\;%
    (\textcolor[HTML]{B3F0B3}{\raisebox{0.0em}{\scalebox{0.75}[0.75]{\ding{110}}}}).
    For \textbf{classification} (a),
    ReLUs are close to best.
    Activations optimized from ReLUs only improve the accuracy slightly,
    corresponding to the GeLU-like function in \hyperref[fig:imgAct]{Figure~\ref{fig:imgAct}}.
    For \textbf{regression} (b), new activations (learned from zeros) are best.
    Moreover, accuracy and complexity are clearly
    correlated only for regression.
    This supports the hypothesis that regression is more complex than classification
    and thus benefits from alternatives to the ReLUs' simplicity bias.
    See \hyperref[fig:imgPlotsAll]{Figure~\ref{fig:imgPlotsAll}} 
    for similar results on other datasets.
  }
\end{figure}

\subsection{Regression Tasks}
\label{sec:regression}

\paragraph{Background.}
Regression tasks are known to be difficult for NNs~\cite{stewart2023regression}.
They are often turned into a classification
through discretization~\cite{farebrother2024stop,imani2024investigating}.
Existing explanations that invoke implicit biases of gradient descent are clearly incomplete~\cite{stewart2023regression}.
\textbf{Our hypothesis} is that
regression is difficult because
it often involves irregular decision boundaries~\cite{grinsztajn2022tree}
in opposition to
the typical solutions of ReLU networks~\cite{dherin2022neural}.

\paragraph{Setup.}
We use the same setup and image datasets
as \hyperref[sec:img]{Section~\ref{sec:img}}.
The task is now to directly predict class IDs.
E.g.\ for \mnist this means predicting digit values.
Models are trained with an MSE loss.
To measure accuracy,
we discretize the predictions to the nearest class ID.

\paragraph{Results.}
The first observation from
\hyperref[fig:imgActBars]{Figure~\ref{fig:imgActBars}b}
is that regression is clearly more difficult for NNs than classification
(lower accuracies)
despite the identical underlying task.
Importantly, the learned activations now provide clear improvements, especially when learned from scratch (initialization as zeros).
This confirms the hypothesis that the inductive biases of ReLUs are not well suited to these tasks.

\hyperref[fig:imgAct]{Figure~\ref{fig:imgAct}b}\ shows that the learned activations contain more irregularities for regression than classification.
Prior work~\cite{teney2024neural} showed that this can help models represent complex functions with sharp transitions.
An analysis of the complexity of trained models
(\hyperref[fig:imgPlots]{Figures~\ref{fig:imgPlots}} and \ref{fig:imgPlotsAll})
shows that the accuracy is correlated with complexity for regression but not classification.
And regression models with learned activations implement functions of higher complexity than with ReLUs.
This supports the claim that the improvements arise from 
overcoming the simplicity bias of ReLUs.

\phantomsection 
\label{sec:diffAccSameComplexity}
\paragraph{Complexity is only one dimension}%
of the inductive biases.
The complexity plots for \svhn 
(\hyperref[fig:imgPlotsAll]{Figure~\ref{fig:imgPlotsAll}})
interestingly show that
models with ReLUs and learned activations get different accuracies at the same complexity level.
This shows that our meta learning approach can
search over dimensions of the inductive biases
that are not captured by our complexity measure,
and are yet to be explicitly studied.


\begin{mdframed}[style=takeawayFrame]
\textbf{Take-away:} regression is more difficult for NNs than classification, and the simplicity bias of ReLUs is partly to blame.
Learned activations improve performance by helping networks represent more complex functions.
\end{mdframed}\vspace{-3pt}

\subsection{Tabular Data}
\label{sec:tabular}

\paragraph{Background.}
Tabular data is
any data with few unstructured dimensions,
which often contains low-cardinality variables such as dates or categorical attributes.
This contrasts e.g.\
with images,
which contain many 
correlated, continuous dimensions (pixels).
NNs struggle with tabular datasets
and are often inferior to decision trees~\cite{grinsztajn2022tree,mcelfresh2024neural}.
\textbf{Our hypothesis} is that the inductive biases of
standard
architectures are ill-suited
to such data
because of the simplicity bias.
It makes it difficult to learn functions
where small changes in the input (e.g.\ day of the week)
correspond to abrupt changes in the target 
---~the definition of \textit{sensitivity}, a proxy for complexity~\cite{dherin2022neural}.
This seldom occurs in vision 
where similar images 
correspond to similar labels.


\paragraph{Setup.}
We use 16 real-world classification datasets from
Grinsztajn et al.~\cite{grinsztajn2022tree,grinTabBenchmark}.
Baselines include a linear classifier,
 k-NNs, and boosted decision trees.
Our models are MLPs
with 1--4 hidden layers
(details in \hyperref[detailsTabular]{Appendix~\ref{detailsTabular}}).
We compare learned activations functions with ReLUs and TanHs with a global prefactor,
$\operatorname{tanh}(\alpha x)$ with $\alpha\!\in\!\mathrm{R}^{+}$ tuned on the validation set.
This is a simple option with tunable complexity, albeit with inductive biases of TanHs~\cite{jagtap2020adaptive,teney2024neural}.

We also experiment with learned \textit{input activation functions} (IAFs).
The motivation 
is to learn a different behavior for each input dimension.
Since they carry different information, e.g.\ continuous vs.\ categorical variables,
one could be suited to the simplicity bias while another is not, for example.
IAFs are dimension-specific activation functions
applied directly on the data before a standard MLP.
IAFs are learned like AFs,
from an initialization as the identity 
i.e.\ no effect by default.
They subsume the gated inputs, and Fourier/numerical embeddings from prior work
~\cite{dragoiclosing,fiedler2021simple,gorishniy2022embeddings}.

\begin{figure}[bht!]
  \centering
  \vspace{-4pt}
  \includegraphics[width=.85\columnwidth]{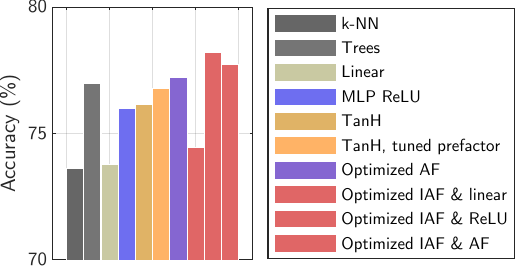}
  \vspace{-8pt}
  \caption{\label{fig:tab}
  Comparison of model types over 16 tabular datasets.
  Vanilla MLPs often perform worse than decision trees,
  but adjusting their inductive biases with learned activation functions (AFs) eliminates this gap.
  The \textit{input activation functions} (IAFs) enable even better performance.
  See \hyperref[fig:tabAllDatasets]{Figure~\ref{fig:tabAllDatasets}} for results per dataset.
  }
\end{figure}

\paragraph{Results.}%
We compare the accuracy of models on the 16 datasets in
\hyperref[fig:tab]{Figures~\ref{fig:tab}} and 
\hyperref[fig:tabAllDatasets]{\ref{fig:tabAllDatasets}}.
Vanilla MLPs generally perform worse than trees.
But adjusting the MLPs' inductive biases with learned prefactors or
activations eliminates the gap.
IAFs perform best, sometimes even surpassing trees.
We analyze below the reasons for these improvements.

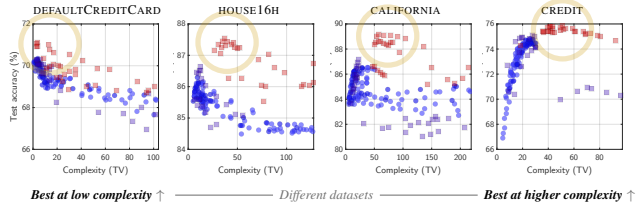
\begin{figure}[ht!]
  \vspace{-8pt}
  \centering
  \renewcommand{\tabcolsep}{0em}
  \renewcommand{\arraystretch}{1}
  \begin{tabularx}{\columnwidth}{*{4}{>{\centering\arraybackslash}X}}
    \tiny~~~~~~\textsc{defaultCreditCard}&
    \tiny~~~\textsc{house16h}&
    \tiny~~~\textsc{california}&
    \tiny~~~\textsc{credit}\\[-1pt]
    \begin{overpic}[trim=9pt 0pt 4pt 0pt, height=.25\columnwidth]{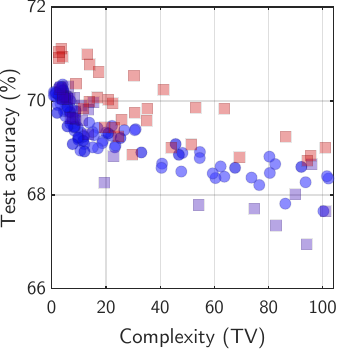}
        \put(0, 65){\begin{tikzpicture}\draw[darkYellow, line width=2pt, opacity=0.25] (0,0) circle (11pt);\end{tikzpicture}}
    \end{overpic}&
    \begin{overpic}[clip, trim=9pt 0pt 3pt 0pt, height=.25\columnwidth]{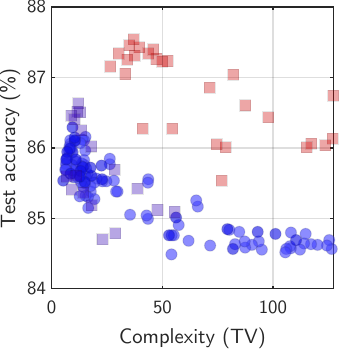}
        \put(14, 66){\begin{tikzpicture}\draw[darkYellow, line width=2pt, opacity=0.25] (0,0) circle (11pt);\end{tikzpicture}}
    \end{overpic}&
    \begin{overpic}[clip, trim=9pt 0pt 3pt 0pt, height=.25\columnwidth]{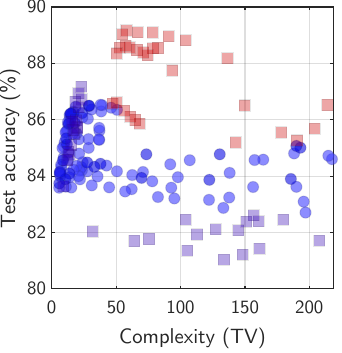}
        \put(16, 70){\begin{tikzpicture}\draw[darkYellow, line width=2pt, opacity=0.25] (0,0) circle (11pt);\end{tikzpicture}}
    \end{overpic}&
    \begin{overpic}[clip, trim=9pt 0pt -9pt 0pt, height=.25\columnwidth]{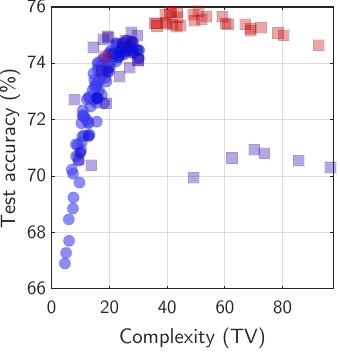}
        \put(31, 74){\begin{tikzpicture}\draw[darkYellow, line width=2pt, opacity=0.25] (0,0) circle (11pt);\end{tikzpicture}}
    \end{overpic}\\[-4pt]
    \multicolumn{4}{r}{\tiny\textcolor{darkGray}{\textit{
        \textcolor{black}{\textbf{Best at low complexity}}~$\uparrow$~~~%
        \raisebox{0.5ex}{\rule{7.4em}{0.4pt}}%
        ~~Different datasets~~%
        \raisebox{0.5ex}{\rule{7.4em}{0.4pt}}%
        ~~\textcolor{black}{\textbf{Best at higher complexity}}~$\uparrow$%
    }}}
  \end{tabularx}
  \vspace{-8pt}
  \caption{\label{fig:tabPlots}
    Test accuracy vs.\ complexity 
    on tabular datasets.
    Each marker represents a model with different hyperparameters, 
    and ReLUs\;%
    (\textcolor[HTML]{5C5CFD}{\raisebox{0.0em}{\scalebox{0.75}[0.75]{\ding{108}}}})
    or learned activations initialized as ReLUs\;%
    (\textcolor[HTML]{B8A6E4}{\raisebox{0.0em}{\scalebox{0.75}[0.75]{\ding{110}}}})    
    or as zeros\;%
    (\textcolor[HTML]{EDA6A6}{\raisebox{0.0em}{\scalebox{0.75}[0.75]{\ding{110}}}}).
    The learned activations perform better in all cases,
    but the accuracy peaks at different complexity levels.
    For some datasets, a low complexity is best
    and ReLUs thus perform quite well
    (leftmost panel, note the smaller Y scale).
    For other datasets, the opposite is true and the improvements with learned activations is larger.
  }
\end{figure}

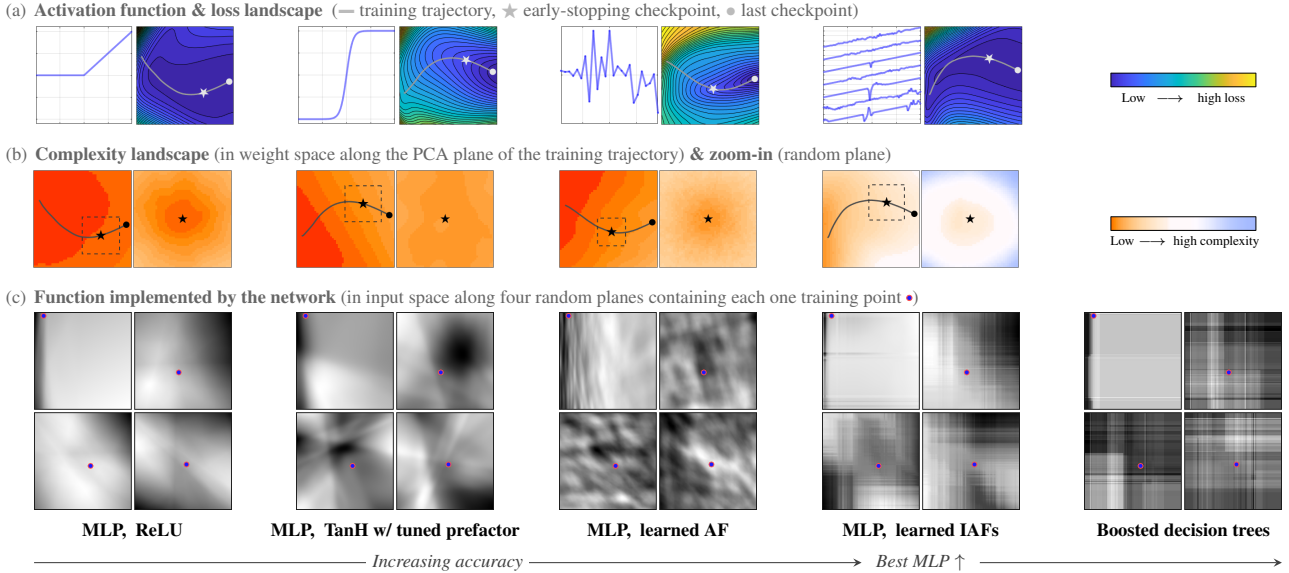
\begin{figure*}[ht!]
\centering
\renewcommand{\tabcolsep}{.25em}
\renewcommand{\arraystretch}{1.5}
\scriptsize
\begin{tabularx}{\textwidth}{*{5}{>{\centering\arraybackslash}X}}

  \multicolumn{5}{l}{%
  \textcolor[HTML]{666666}{%
  (a)~\,\textbf{Activation function \& loss landscape} ~(%
  \textcolor[HTML]{BBBBBB}{\raisebox{0.16em}{\scalebox{1.2}[0.25]{\ding{110}}}}
  training trajectory,
  \textcolor[HTML]{BBBBBB}{$\bigstar$} early-stopping checkpoint,
  \textcolor[HTML]{BBBBBB}{$\bullet$} last checkpoint%
  )}}\\

\makecell{
\includegraphics[clip, trim=2pt 11.1pt 6.5pt 0pt, height=.076\textwidth]{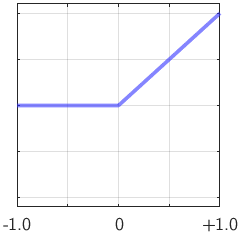}%
 \reflectbox{\rotatebox[origin=c]{180}{\includegraphics[height=.076\textwidth]{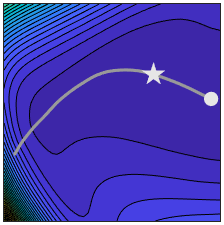}}}%
}&
\makecell{
\includegraphics[clip, trim=2pt 11.1pt 6.5pt 0pt, height=.076\textwidth]{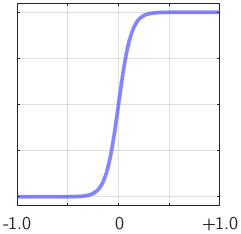}%
\reflectbox{\rotatebox[origin=c]{180}{\includegraphics[height=.076\textwidth]{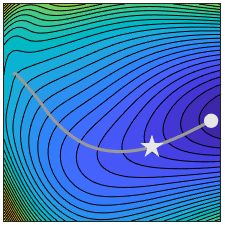}}}%
}&
\makecell{
\includegraphics[clip, trim=2pt 11.1pt 6.5pt 0pt, height=.076\textwidth]{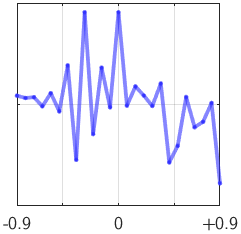}%
\reflectbox{\rotatebox[origin=c]{180}{\includegraphics[height=.076\textwidth]{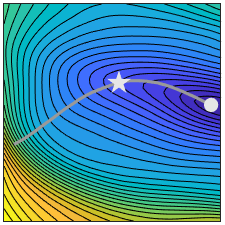}}}%
}& 
\makecell{
\includegraphics[clip, trim=2pt 11.1pt 6.5pt 0pt, height=.076\textwidth]{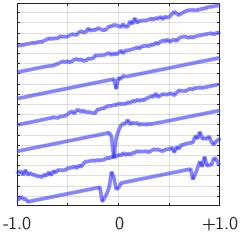}%
\reflectbox{\rotatebox[origin=c]{180}{\includegraphics[height=.076\textwidth]{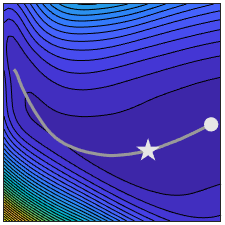}}}%
}&
\makecell[cb]{
        \rotatebox[origin=b]{270}{\fbox{\includegraphics[clip, trim=3pt 3pt 3pt 3pt, width=.01\textwidth, height=.11\textwidth]{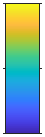}}}\\
        \tiny~Low~~~$\longrightarrow$~~~~high loss
        \vspace{1.3em}
}\\[1pt]

\multicolumn{4}{l}{%
\textcolor[HTML]{666666}{%
(b)~\,\textbf{Complexity landscape} (in weight space along the PCA plane of the training trajectory)
\textbf{\& zoom-in}
(random plane)
}}\\
\makecell{
\includegraphics[height=.076\textwidth]{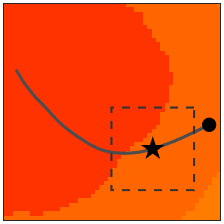}%
\includegraphics[height=.076\textwidth]{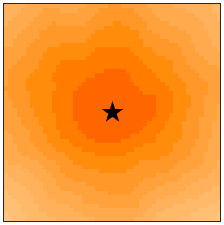}%
}&
\makecell{
\includegraphics[height=.076\textwidth]{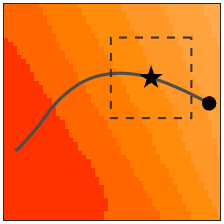}%
\includegraphics[height=.076\textwidth]{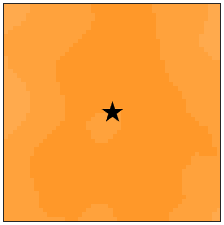}%
}&
\makecell{
\includegraphics[height=.076\textwidth]{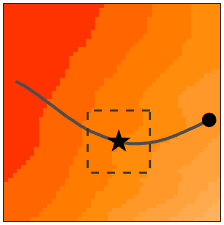}%
\includegraphics[height=.076\textwidth]{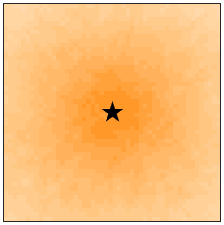}%
}&
\makecell{
\includegraphics[height=.076\textwidth]{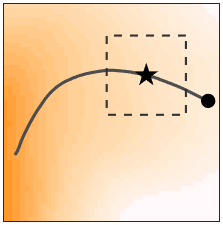}%
\includegraphics[height=.076\textwidth]{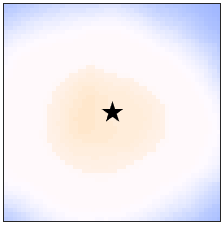}%
}&
\makecell[cb]{
        \rotatebox[origin=b]{270}{\fbox{\includegraphics[clip, trim=3pt 3pt 3pt 3pt, width=.01\textwidth, height=.11\textwidth]{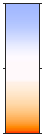}}}\\
        \tiny~~Low~$\longrightarrow$~~high complexity
        \vspace{1.3em}
}\\[1pt]

\multicolumn{4}{l}{%
\textcolor[HTML]{666666}{%
(c)~\,\textbf{Function implemented by the network} (in input space along four random planes
containing each one training point~\coloredBullet) }}\\
\makecell{\includegraphics[height=.15\textwidth]{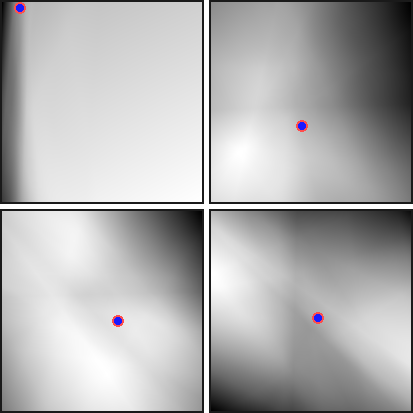}}&
\makecell{\includegraphics[height=.15\textwidth]{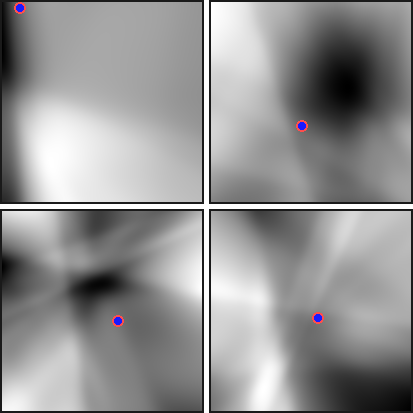}}&
\makecell{\includegraphics[height=.15\textwidth]{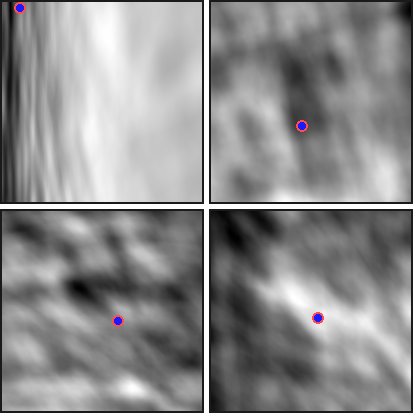}}&
\makecell{\includegraphics[height=.15\textwidth]{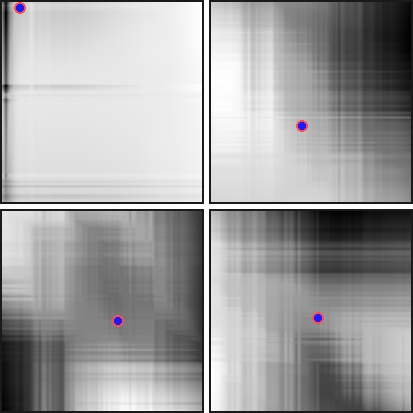}}&
\makecell{\includegraphics[height=.15\textwidth]{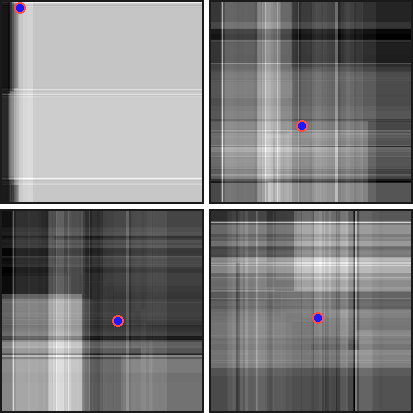}}
\\[-2pt]
\textbf{MLP,~~ReLU}&
\textbf{MLP,~~TanH w/ tuned prefactor}&
\textbf{MLP,~~learned AF}&
\textbf{MLP,~~learned IAFs}&
\textbf{Boosted decision trees}
  \end{tabularx}
  \begin{tikzpicture}
    \draw[-, color=darkgray, line width=0.4pt, arrows=-{stealth}] (0,0) -- (0.63\textwidth,0)
    node[midway, fill=white, text=darkgray] {\textit{Increasing accuracy}}
    node[fill=white, text=darkgray] at (0.675\textwidth, 0) {\textit{Best MLP} \bfseries$\uparrow$};
    \draw[->, color=darkgray, line width=0.4pt, arrows=-{stealth}] (0.72\textwidth,0) -- (0.95\textwidth,0);
  \end{tikzpicture}\\
  \vspace{-8pt}
  \caption{\label{fig:tabBoundaries}
  Models trained on the \textsc{electricity}~\cite{grinTabBenchmark}
  tabular dataset.
  ReLU MLPs perform worst (left).
  TanHs induce sharper transitions in the \textbf{network's function (c)}.
  So does the learned \textbf{activation function (a)} which is itself very irregular. 
  The 
  input activation functions (IAFs) perform best and
  mimic the axis-aligned 
  boundaries of trees (bottom-right).
  The \textbf{complexity landscapes (b)} show that complexity
  increases to the highest level in the best model (IAFs).
  The zoom-in shows that
  the ambient complexity 
  is also much higher than in other models.
  This means that it is inherently more likely to represent complex functions
  since they are more abundant in parameter space.
  }
  \vspace{-9pt}
\end{figure*}

\paragraph{Learned activation functions close the gap to decision trees by mimicking their inductive bias.}
We visualize in
\hyperref[fig:tabBoundaries]{Figure~\ref{fig:tabBoundaries}c}
the functions implemented by different models,
plotting their output
over 
slices of the input space
(\hyperref[sec:paths]{Section~\ref{sec:paths}}).
ReLUs produce the smoothest 
function
while TanHs and learned activations induce sharper patterns. 
Notably, the IAFs
induce sharp axis-aligned decision boundaries
that are also characteristic of trees, with which they share a high accuracy.
Axis-aligned transitions
are the consequence of IAFs applied \emph{independently} to each dimension.
Sharp transitions originate from the complex shape of the learned activation function (\hyperref[fig:tabBoundaries]{Figure~\ref{fig:tabBoundaries}a})
which
is possible
thanks to the unbiased spline parametrization
(\hyperref[sec:splines]{Section~\ref{sec:splines}}).
  
\setlength{\fboxrule}{0.2pt} 
\setlength{\fboxrule}{0.4pt} 

\paragraph{The simplicity/complexity bias is a property of the architecture.}
We visualize \textit{complexity landscapes} of MLPs
in \hyperref[fig:tabBoundaries]{Figure~\ref{fig:tabBoundaries}b}.
Similarly to standard \textit{loss} landscapes~\cite{li2018visualizing},
we plot model complexity over 2D slices of the parameter space.
A first global view over a plane aligned with the training trajectory
shows that complexity steadily increases during training
for all models~\cite{kalimeris2019sgd,rahaman2019spectral,xu2019frequency}
but does so to the highest level for the best model (IAFs).
A second view zooms in on each optimized solution
in a random 2D plane.
This examines the effect of \emph{arbitrary} perturbations to the parameters.%
    \footnote{This resembles an analysis of untrained models
    \cite{de2019random,mingard2019neural,teney2024neural,valle2018deep}
    but focuses on relevant regions on the parameter space, near optimized models.}
It shows that the ambient complexity 
of perturbed solutions of the best model
is much higher than the solution itself,
and than with less accurate models.
This means that this architecture is more likely to represent complex functions
because they are more abundant in parameter space~\cite{mingard2021sgd,scimeca2021shortcut}.
\textbf{This is why the simplicity bias can be overcome}:
it results from architecture choices
and not from an inevitable ``implicit bias'' of SGD~\cite{shah2020pitfalls,soudry2018implicit,tsoy2024simplicity,xu2019training}.


\paragraph{Different tabular datasets require different inductive biases.}
We examine 
the relation between accuracy and complexity in
\hyperref[fig:tabPlots]{Figure~\ref{fig:tabPlots}}.
The accuracy peaks at different complexity levels for different datasets.
For some, a low complexity is best and
ReLU MLPs
perform well.
For others, a higher complexity is best and the improvements with learned activations are larger.
This supports the hypothesis that improvements
over ReLU MLPs come from overcoming their simplicity bias.
The variance across datasets is also unsurprising
since they have little in common besides their low dimensionality
(full results in \hyperref[detailsTabular]{Appendix~\ref{sec:results2Tabular}}).

\begin{figure}[ht!]
  \centering
  \includegraphics[width=.63\columnwidth]{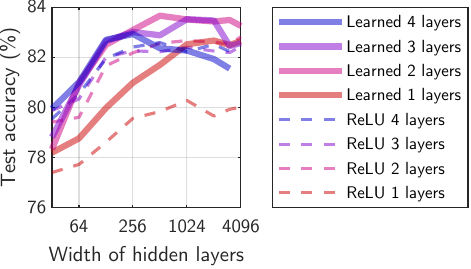}
  \vspace{-7pt}
  \caption{\label{fig:tabDim}
  The learned activation functions surpass ReLUs, often with fewer layers.
  They can also be reused with different network widths
  (\textsc{covertype}~\cite{grinTabBenchmark} tabular dataset,
  see \hyperref[fig:tabNLayers]{Figure~\ref{fig:tabNLayers}} for others).
  \vspace{-5pt}
  }
\end{figure}

\paragraph{Effect of width and depth.}
We show in
\hyperref[fig:tabDim]{Figure~\ref{fig:tabDim}}
that the learned activations can be reused in networks of different widths than they were trained for. The accuracy varies with width similarly as with ReLUs.
\citet{teney2024neural} indeed showed that a model's width affects its capacity but not its inductive biases. Therefore width does not interfere with the effects of the learned activations.
\hyperref[fig:tabDim]{Figures~\ref{fig:tabDim}} and ~\ref{fig:tabNLayers} also show that good performance can be achieved with fewer layers than with ReLUs.
Learned activations might thus have utility in model compression and distillation.

\begin{mdframed}[style=takeawayFrame]
\textbf{Take-away:} 
many tabular datasets are ill-suited to ReLU models
because they require learning a complex function.
Learned activations improve accuracy
by implementing sharp axis-aligned decision boundaries
that mimic the inductive biases of decision trees.
\end{mdframed}\vspace{-3pt}

\begin{figure*}[t!]
  \renewcommand{\tabcolsep}{0.2em}
  \renewcommand{\arraystretch}{1}
  \centering
  \begin{tabular}{ccccc}
    \makecell[b]{
    \includegraphics[height=.0365\textwidth]{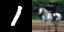}\\[-2pt]
    \includegraphics[height=.0365\textwidth]{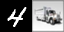}\\[3.5pt]
    \footnotesize Ambiguous\\[-3pt]\footnotesize training images}&
    \makecell[b]{
    \includegraphics[height=.084\textwidth]{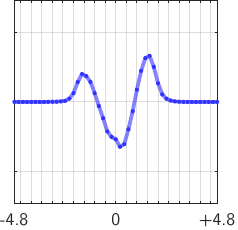}\\
    \footnotesize Optimized\\[-3pt]\footnotesize for \mnist}&
    \makecell[b]{
    \includegraphics[height=.084\textwidth]{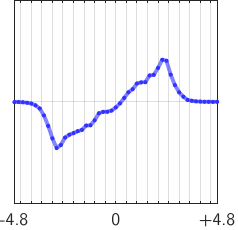}\\
    \footnotesize Optimized\\[-3pt]\footnotesize for \cifar}&
    \makecell[br]{\includegraphics[height=.12\textwidth]{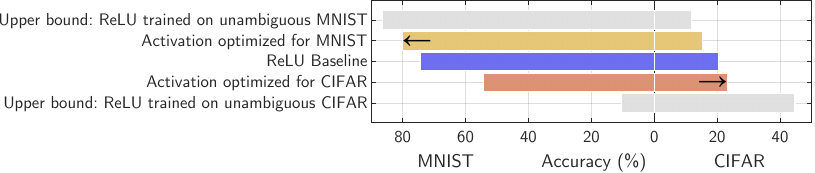}\vspace{-1pt}}&
    \makecell[br]{\includegraphics[clip, trim=0pt 0pt 225pt 0pt, height=.12\textwidth]{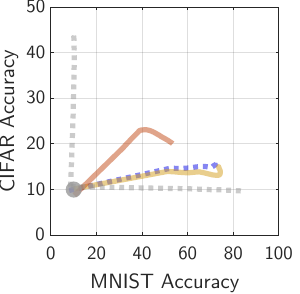}\vspace{0.02pt}}
  \end{tabular}
  \vspace{-4pt}
  \caption{\label{fig:shortcut}
  Experiments on shortcut learning with \mnist/\cifar collages.
  The ReLU baseline 
  (\textcolor[HTML]{6E6EF0}{\raisebox{0.0em}{\scalebox{1.2}[0.85]{\ding{110}}}})
  relies mostly on simple \mnist features.
  We learn two activation functions that shift the preference towards different features
  ($\leftarrow$/$\rightarrow$).
  Training trajectories (right) clearly differ
  with the activation optimized for \cifar~%
  (\textcolor[HTML]{E1A58A}{\raisebox{0.17em}{\scalebox{1.2}[0.3]{\ding{110}}}}),
  for \mnist~%
  (\textcolor[HTML]{EACF8E}{\raisebox{0.17em}{\scalebox{1.2}[0.3]{\ding{110}}}}),
  or a ReLU~%
  (\textcolor[HTML]{8686F2}{\raisebox{0.17em}{\scalebox{0.5}[0.3]{\ding{110}}}}%
   \textcolor[HTML]{FFFFFF}{\raisebox{0.17em}{\scalebox{0.2}[0.3]{\ding{110}}}}%
   \textcolor[HTML]{8686F2}{\raisebox{0.17em}{\scalebox{0.5}[0.3]{\ding{110}}}}).
  The model at initialization (random weights) is marked with 
  \textcolor[HTML]{ACABAA}{\scalebox{1.3}[1.3]{$\bullet\,$}}.
  }
  \vspace{-10pt}
\end{figure*}

\subsection{Shortcut Learning}
\label{sec:shortcut}

\paragraph{Background.}
Shortcut learning occurs when a model learns
spurious
features
instead of generalizable ones.
It is a
known
consequence of the simplicity bias~\cite{shah2020pitfalls,teney2021evading}
when the training data contains multiple features of different complexity.
\textbf{Our hypothesis} is that
the preference for some features depends on their alignment with the inductive biases.
We will evaluate whether this can be controlled with activation functions.


\paragraph{Setup.}
We use \mnist/\cifar collages~\cite{shah2020pitfalls,teney2021evading,teney2022predicting}, a
classification task
over images combining tiles from \mnist and \cifar-10.
The \textbf{training set} is ambiguous: both tiles are predictive of the labels.
Two unambiguous \textbf{test sets} evaluate reliance on either tile:
one is predictive, the other contains a random class.
We similarly build two \textbf{validation sets}
to learn \textbf{two activation functions} optimized for either tile.
We simulate OOD conditions by setting $\mathcal{V}$ in Algorithm~\ref{alg}.
The models are the fully-connected MLPs used in~\cite{teney2021evading}.



\paragraph{Results.}
\hyperref[fig:shortcut]{Figure~\ref{fig:shortcut}}
shows that a baseline with ReLUs is prone to shortcut learning.
It relies exclusively on \mnist and the accuracy on the \cifar test set is not better than chance (10\%).
In comparison,
using either learned activation
steers the learning towards either tile.
The accuracy shifts towards either of two tiles
as the model prioritizes different features, merely with a change of activation function.
This shows that the simplicity bias is not an inevitable effect of SGD.
Instead, it directly reflects 
the alignment between the chosen architecture and the data.

\paragraph{Training dynamics.}
The accuracy on \cifar
remains below
a model trained on unambiguous \cifar data.
This is because training dynamics are also important.
In \hyperref[fig:shortcut]{Figure~\ref{fig:shortcut} (right)},
we plot the accuracy on the two tiles for the whole training trajectory.
The reliance on different features
varies,
and the model eventually relies primarily on simple ones
with enough iterations (i.e.\ without early stopping).
This calls for future work
combining our findings with the extensive literature on ID\,/\,OOD training dynamics~\cite{jain2024bias,teney2024id,tsoy2024simplicity}.



\begin{mdframed}[style=takeawayFrame]
\textbf{Take-away:} we confirm
that shortcut learning is
a side effect of the simplicity bias.
Different activation functions,
while not completely avoiding shortcut learning,
can steer the learning towards particular input features.
\end{mdframed}\vspace{-3pt}

\subsection{Algorithmic Tasks and Grokking}
\label{sec:grokking}

\paragraph{Background.}
Grokking is a phenomenon where a model first overfits the data 
(i.e.\ high training accuracy, low test accuracy)
then 
shifts to high test accuracy
after many training steps~\cite{power2022grokking}.
This is typically observed on
algorithmic tasks
and architectures from MLPs to transformers.
\textbf{Our hypothesis} is 
that grokking is due to a mismatch between the target function
and the model's inductive biases.
Indeed, typical architectures were not developed for
the algorithmic tasks where grokking is typically observed.
To verify this hypothesis, we will show that endowing an architecture with the right inductive biases, using learned activation functions, can eliminate the phenomenon.
Supporting this hypothesis,
\citet{zhou2024rationale} proposed that grokking
comes from the frequency principle (i.e.\ low frequencies learned first by SGD),
and \citet{kumar2023grokking} showed that
it correlates with a misalignment between features at initialization and the target function.

\begin{figure}[ht!]
  \centering
  \renewcommand{\tabcolsep}{0em}
  \renewcommand{\arraystretch}{1}
  \begin{tabularx}{\columnwidth}{*{6}{>{\centering\arraybackslash}X}}
    \makecell{\scriptsize $a\!+\!b$\\[-4pt]\scriptsize$(\modulo 27)$}&
    \makecell{\scriptsize $ab$\\[-4pt]\scriptsize$(\modulo 27)$}&
    \makecell{\scriptsize $a^2\!+\!ab\!+\!b^2$\\[-4pt]\scriptsize$(\modulo 53)$}&
    \makecell{\scriptsize $a^2 + b^2$\\[-4pt]\scriptsize$(\modulo 27)$}&    
    \makecell{\scriptsize $a^3 + ab$\\[-4pt]\scriptsize$(\modulo 53)$}&
    \makecell{\scriptsize $a.b$\\[-4pt]\scriptsize in $S_4$}\\[8pt]
    \makecell{\includegraphics[width=.075\textwidth]{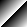}}&
    \makecell{\includegraphics[width=.075\textwidth]{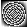}}&
    \makecell{\includegraphics[width=.075\textwidth]{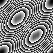}}&
    \makecell{\includegraphics[width=.075\textwidth]{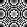}}&
    \makecell{\includegraphics[width=.075\textwidth]{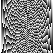}}&
    \makecell{\includegraphics[width=.075\textwidth]{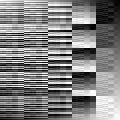}}
  \end{tabularx}
  \vspace{-9pt}
  \caption{\label{fig:grokking}
  Target functions used
  to investigate grokking~\cite{power2022grokking}
  (details in Appendix, \hyperref[fig:grokkingData]{Figure~\ref{fig:grokkingData}}).
  These patterns are very different from the tasks for which typical architectures were developed.
  \vspace{-6pt}
  }
\end{figure}

\paragraph{Setup.}
Following 
\cite{gromov2023grokking,kumar2023grokking,liu2022towards}
we train $1$-hidden layer MLPs on algorithmic tasks,
defined each
by one binary operation
(\hyperref[fig:grokking]{Figures~\ref{fig:grokking}} and~\ref{fig:grokkingData}).
e.g.\ $y\!=\!(x_1\!+x_2)\bmod\,13$.
The operands 
are passed as one-hot vectors
and the task is a classification over possible outputs.
Details in \hyperref[detailsGrokking]{Appendix~\ref{detailsGrokking}}.


\begin{figure}[ht!]
  \raggedleft 
  \vspace{4pt}
  \begin{overpic}[width=.4\columnwidth]{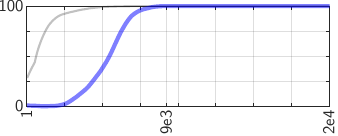}
    \put(52, 45){\makebox(0,0){{\scriptsize\textbf{ReLU baseline}}}}
    \put(-1, 32){\makebox(0,0)[r]{\scriptsize{Accuracy (\%)}}}
    \put(-1, 22){\makebox(0,0)[r]{\scriptsize
      \textcolor[HTML]{777777}{\raisebox{0.22em}{\scalebox{1.4}[0.125]{\ding{110}}}}
      Training
    }}
    \put(-1, 13){\makebox(0,0)[r]{\scriptsize
      \textcolor[HTML]{7F7FFF}{\raisebox{0.18em}{\scalebox{1.4}[0.28]{\ding{110}}}}
      Test
    }}
    \put(73, 2){\makebox(0,0){{\scriptsize{Tr. steps}}}}
  \end{overpic}
  \begin{overpic}[width=.4\columnwidth]{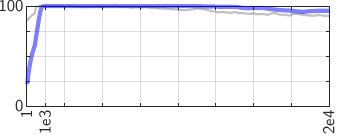}
    \put(52, 45){\makebox(0,0){{\scriptsize\textbf{Learned activation function}}}}
    \put(73, 2){\makebox(0,0){{\scriptsize{Tr. steps}}}}
  \end{overpic}
  \vspace{-5pt}
  \caption{\label{fig:grokking2}
  The learned activations essentially eliminate grokking (delayed convergence).
  On the above task
  (addition~$\bmod~27$),
  our model converges $\sim\!10\times$ faster than ReLUs.
  }
  \vspace{-8pt}
\end{figure}
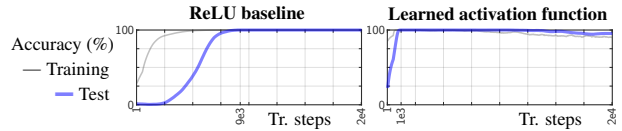

\begin{figure}[ht!]
  \vspace{-4pt}
  \centering
  \renewcommand{\arraystretch}{1}
  \begin{tabularx}{\columnwidth}{*{4}{>{\centering\arraybackslash}X}}
    {\fbox{\includegraphics[width=.18\columnwidth, trim=5 38 5 38, clip]{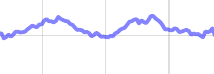}}}&
    {\fbox{\includegraphics[width=.18\columnwidth, trim=5 38 5 38, clip]{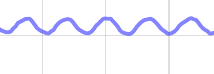}}}&
    {\fbox{\includegraphics[width=.18\columnwidth, trim=5 38 5 38, clip]{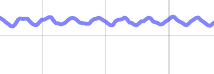}}}&
    {\fbox{\includegraphics[width=.18\columnwidth, trim=5 38 5 38, clip]{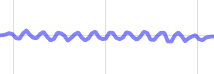}}}\\[-2pt]
    {\scriptsize$(\bmod\,13)$}&
    {\scriptsize$(\bmod\,27)$}&
    {\scriptsize$(\bmod\,29)$}&
    {\scriptsize$(\bmod\,41)$}
  \end{tabularx}
  \vspace{-10pt}
  \caption{\label{fig:grokking3}
  Activations learned for 
  modular addition. 
  The frequency of the sine-like function varies across versions of the task.
  }
  \vspace{-8pt}
\end{figure}

\begin{figure}[ht!]
  \vspace{-4pt}
  \centering  
  \hspace{10pt} 
  \begin{overpic}[height=.22\columnwidth]{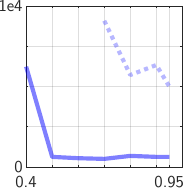}
    \put(1, 74){\makebox(0,0)[r]{\scriptsize{Num. training}}}
    \put(1, 59){\makebox(0,0)[r]{\scriptsize{steps to reach}}}
    \put(1, 43){\makebox(0,0)[r]{\scriptsize{{\tiny$\ge$}~$95\%$ test}}}
    \put(1, 27){\makebox(0,0)[r]{\scriptsize{accuracy}}}
    \put(55, -8){\makebox(0,0){{\scriptsize{Fraction of tr. data}}}}
  \end{overpic}~~~
  \begin{overpic}[height=.22\columnwidth]{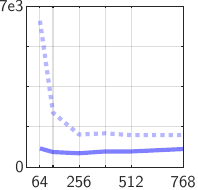}
    \put(55, -8){\makebox(0,0){{\scriptsize{Network width}}}}
  \end{overpic}
  \vspace{4pt}
  \caption{\label{fig:grokking4}
  Models with
  learned activations\;(\textcolor[HTML]{7F7FFF}{\raisebox{0.16em}{\scalebox{1.2}[0.25]{\ding{110}}}})
  converge faster than
  ReLUs\;(\textcolor[HTML]{7F7FFF}{\raisebox{0.16em}{\scalebox{0.5}[0.25]{\ding{110}}}}%
   \textcolor[HTML]{FFFFFF}{\raisebox{0.16em}{\scalebox{0.2}[0.25]{\ding{110}}}}%
   \textcolor[HTML]{7F7FFF}{\raisebox{0.16em}{\scalebox{0.5}[0.25]{\ding{110}}}})  
  across a variety of settings
  (addition~$\bmod\,27$).
  }
\end{figure}

\paragraph{Results.}
We compare models with ReLU vs.\ learned activations
across various tasks,
network widths, and fractions of training data.
We find that the learned, task-specific activations
lead to faster convergence and/or higher test accuracy
(\hyperref[fig:grokking2]{Figures~\ref{fig:grokking2}}--\ref{fig:grokking4}).
On modular addition 
(a common task in the grokking literature)
the learned-activation model converges $\sim\!10\times$ faster than ReLUs.
Curiously, some models with learned activations also
end up overfitting (decreasing test accuracy) with prolonged training.
In contrast, ReLU networks either never generalize (test accuracy $\sim\!0$)
or \textit{grok} and keep a high accuracy indefinitely.
Further investigation is needed to explain this difference.
We examine learned activation functions in \hyperref[fig:grokking3]{Figure~\ref{fig:grokking3}}.
See \hyperref[fig:grokkingData]{Figure~\ref{fig:grokkingFullResults}}
in the Appendix for results on other algorithmic tasks~\cite{power2022grokking}.


\begin{mdframed}[style=takeawayFrame]
\textbf{Take-away:}
learned activations 
eliminate grokking in all our cases,
suggesting, as a cause, the mismatch between the data and the architectures' inductive biases.
\end{mdframed}\vspace{-3pt}%

\section{Do the Activations Transfer Across Tasks?}
\label{sec:transfer}
So far,
we used dataset-specific activation functions
and found that there exist better alternatives to ReLUs.
A practical application
would be the learning of activation functions
suitable to a broad task, or {range} of related datasets.

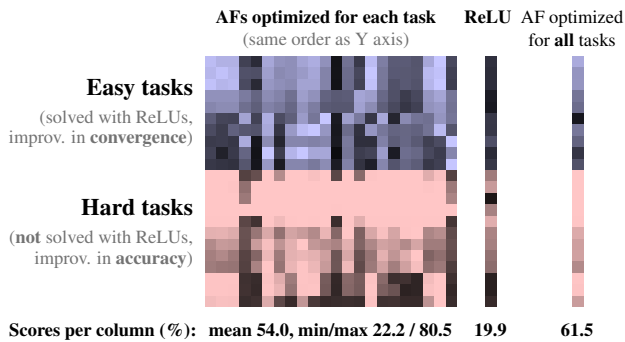
\begin{figure}[h!]
  \raggedleft
  \vspace{12pt}
  \begin{overpic}[height=.4\columnwidth]{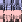}
    \put(-5, 87){\makebox(0,0)[r]{\small{\strut \textbf{Easy tasks}}}}
    \put(-5, 75){\makebox(0,0)[r]{\scriptsize{\strut \textcolor{darkGray}{(solved with ReLUs,}}}}
    \put(-5, 67){\makebox(0,0)[r]{\scriptsize{\strut \textcolor{darkGray}{improv. in \textbf{convergence})}}}}
    \put(-5, 39){\makebox(0,0)[r]{\small{\strut \textbf{Hard tasks}}}}
    \put(-5, 27){\makebox(0,0)[r]{\scriptsize{\strut \textcolor{darkGray}{(\textbf{not} solved with ReLUs,}}}}
    \put(-5, 19){\makebox(0,0)[r]{\scriptsize{\strut \textcolor{darkGray}{improv. in \textbf{accuracy})}}}}
    \put(48, 115){\makebox(0,0){\scriptsize{\strut \textbf{AFs optimized for each task}}}}
    \put(48, 106){\makebox(0,0){\scriptsize{\strut \textcolor{darkGray}{(same order as Y axis)}}}}
    \put(10, -10){\makebox(0,0){\scriptsize{\strut \textbf{Scores per column (\%):~~\,mean 54.0, min/max 22.2 / 80.5}}}}
  \end{overpic}\hspace{11pt}%
  \begin{overpic}[height=.4\columnwidth]{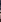}
    \put(1, 115){\makebox(0,0){\scriptsize{\strut \phantom{p}\textbf{ReLU}\phantom{p}}}}
    \put(2, -10){\makebox(0,0){\scriptsize{\strut \textbf{19.9}}}}
  \end{overpic}\hspace{29pt}%
  \begin{overpic}[height=.4\columnwidth]{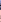}
    \put(0, 115){\makebox(0,0){\scriptsize{\strut AF optimized}}}
    \put(0, 106){\makebox(0,0){\scriptsize{\strut for \textbf{all} tasks}}}
    \put(2, -10){\makebox(0,0){\scriptsize{\strut \textbf{61.5}}}}
  \end{overpic}\hspace{8pt}\phantom{p}
  \vspace{6pt}
  \caption{\label{fig:grokkingTransfer}
  Transfer of AFs (columns) across algorithmic tasks (rows).
  Colors represent the fraction of the best
  \colorbox[HTML]{C5C5FE}{\strut convergence speed}
  or
  \colorbox[HTML]{FEC4C4}{\strut accuracy}
  per task (brighter is better).
  If the activations were over-specialized, the matrix would be diagonal.
  On the contrary, it is densely filled, indicating positive transfer across many tasks.
  }
  \vspace{-8pt}
\end{figure}

As a first step,
we study the specialization of the activations functions (AFs) learned
for the 22 algorithmic tasks from
\hyperref[fig:grokkingData]{Figure~\ref{fig:grokkingData}}
\cite{power2022grokking}.
We evaluate every task/activation combination, yielding the 
$22\!\times\!22$
matrix of
\hyperref[fig:grokkingTransfer]{Figure~\ref{fig:grokkingTransfer}}.
The learned activations do transfer, with
improvements in accuracy and convergence shared across tasks.
We also evaluate an activation learned 
learned on \emph{all tasks} simultaneously.
The accuracy across tasks (i.e.\ per-column average)
reaches $61.5\%$
vs.\ only $19.9\%$ for ReLUs.
and $54.0\%$ on average for tasks-specific solutions.
This procedure can thus improve performance on a \emph{range} of related tasks.
Future work could leverage it
to discover activation functions that improve performance in other specific domains.
See
\hyperref[sec:transferImgReg]{Appendix~\ref{sec:transferImgReg}}
for other transfer experiments using image regression tasks.

\section{Discussion}

We used activation functions as a tool
to show that there exist a variety of inductive biases
that are useful across applications of NNs.
The impossibility of \emph{universal} inductive biases is
well known~\cite{wolpert2002supervised}
but a strong argument has also been made
that deep learning research
is converging towards few architectures with wide applicability~\cite{goldblum2023no}.
This argument rests on the NNs' simplicity bias
being a good match for real-world data~\cite{buchanan2018natural,dingle2018input,lin2017does}.
Our results do not invalidate these assumptions:
NNs are widely applicable
and their simplicity bias is evidently very effective \emph{on average}.
Our results show instead the following.
\begin{enumerate}[itemsep=2pt,topsep=0pt]
\item 
There exist \textbf{real-world tasks where the inductive biases of typical architectures' are suboptimal}.
This explanation connects four domains where NNs historically struggled.
\item 
The \textbf{simplicity bias in modern NNs
depends on particular design choices}, the activation functions in particular.
Research has converged on these choices by trial and error,
in large part by optimizing performance on vision tasks.
Therefore the adequacy of ReLUs for image classification (\hyperref[sec:img]{Section~\ref{sec:img}}) is not accidental.
\end{enumerate}


\paragraph{Relevance to transformers and language models.}
The simplicity bias exists in transformers
\cite{bhattamishra2022simplicity,zhou2023algorithms} and language models~\cite{alkhamissi2024brain,goldblum2023no,teney2024neural}.
Their embedding layer
resembles \textit{input activations functions} (\hyperref[sec:tabular]{Section~\ref{sec:tabular}}).
Could this explain the transformers' remarkable flexibility? I.e.\ a simplicity bias
on
an initial mapping of arbitrary complexity.
\citet{zhong2024algorithmic}
indeed trained embeddings alone
in a random-weight transformer and could learn complex 
tasks.



\paragraph{Limitations and open questions.}
We prioritized breadth by establishing a new connection across
multiple disparate topics in machine learning.
Each section could expand into its own paper
with additional models, datasets, comparisons, etc.
Our findings on shortcut learning
for example
(\hyperref[sec:shortcut]{Section~\ref{sec:shortcut}})
could yield new methods to address distribution shifts,
though no such claim is made here.
Here are the most promising follow-up questions opened by this paper.
\begin{itemize}[itemsep=2pt,topsep=2pt]
\item
\textbf{How to fully characterize inductive biases?}
We focused on simplicity
for its prevalence
in AI~\cite{goldblum2023no}, philosophy~\cite{popper1959logic}, and the natural sciences~\cite{buchanan2018natural}.
But it is only one dimension among many to characterize inductive biases.

\item
\textbf{Can we improve state-of-the-art architectures?}
We used simple MLPs
to isolate the effects of activations functions
since they are central to the simplicity bias~\cite{teney2024neural}.
But other
existing mechanisms (architectural, optimization)
may already tweak or attenuate the simplicity bias.

\item 
\textbf{Can we learn transferable activation functions for other domains?}
We examined transferability 
in \hyperref[sec:transfer]{Sections~\ref{sec:transfer}} and \ref{sec:transferImgReg}.
The results suggest the possibility
of better architectures optimized for specific domains.
Predicting the suitability of an architecture/dataset pair
ex ante (prior to training) would be extremely useful.
This may follow from advances on the first open question above.


\item
\textbf{Are there other detrimental effects of the simplicity bias?}
Any learning algorithm needs inductive biases
to
``fill the gaps'' between training examples.
The better they are, the fewer examples are needed.
Researching what inductive biases are most useful on
real-world tasks might thus hold the key for machine learning to become as data-efficient as humans.
More speculatively,
high-level cognition has 
been argued to require
postulating explanations beyond the data~\cite{deutsch2011beginning,elton2021applying}.
In this regard, simplicity-biased architectures might also hold us back.

\end{itemize}


{
    \small
    \bibliographystyle{ieeenat_fullname}
    \bibliography{main}

\begin{thebibliography}{110}
\providecommand{\natexlab}[1]{#1}
\providecommand{\url}[1]{\texttt{#1}}
\expandafter\ifx\csname urlstyle\endcsname\relax
  \providecommand{\doi}[1]{doi: #1}\else
  \providecommand{\doi}{doi: \begingroup \urlstyle{rm}\Url}\fi

\bibitem[Addepalli et~al.(2022)Addepalli, Nasery, Radhakrishnan, Netrapalli, and Jain]{addepalli2022feature}
Sravanti Addepalli, Anshul Nasery, Venkatesh~Babu Radhakrishnan, Praneeth Netrapalli, and Prateek Jain.
\newblock Feature reconstruction from outputs can mitigate simplicity bias in neural networks.
\newblock In \emph{ICLR}, 2022.

\bibitem[Alexandridis et~al.(2024)Alexandridis, Deng, Nguyen, and Luo]{alexandridis2024adaptive}
Konstantinos~Panagiotis Alexandridis, Jiankang Deng, Anh Nguyen, and Shan Luo.
\newblock Adaptive parametric activation.
\newblock \emph{arXiv preprint arXiv:2407.08567}, 2024.

\bibitem[Alexandridis et~al.(2025)Alexandridis, Deng, Nguyen, and Luo]{alexandridis2025adaptive}
Konstantinos~Panagiotis Alexandridis, Jiankang Deng, Anh Nguyen, and Shan Luo.
\newblock Adaptive parametric activation.
\newblock In \emph{ECCV}, 2025.

\bibitem[AlKhamissi et~al.(2024)AlKhamissi, Tuckute, Bosselut, and Schrimpf]{alkhamissi2024brain}
Badr AlKhamissi, Greta Tuckute, Antoine Bosselut, and Martin Schrimpf.
\newblock Brain-like language processing via a shallow untrained multihead attention network.
\newblock \emph{arXiv preprint arXiv:2406.15109}, 2024.

\bibitem[Apicella et~al.(2019)Apicella, Isgro, and Prevete]{apicella2019simple}
Andrea Apicella, Francesco Isgro, and Roberto Prevete.
\newblock A simple and efficient architecture for trainable activation functions.
\newblock \emph{Neurocomputing}, 2019.

\bibitem[Apicella et~al.(2021)Apicella, Donnarumma, Isgr{\`o}, and Prevete]{apicella2021survey}
Andrea Apicella, Francesco Donnarumma, Francesco Isgr{\`o}, and Roberto Prevete.
\newblock A survey on modern trainable activation functions.
\newblock \emph{Neural Networks}, 2021.

\bibitem[Arora et~al.(2019)Arora, Cohen, Hu, and Luo]{arora2019implicit}
Sanjeev Arora, Nadav Cohen, Wei Hu, and Yuping Luo.
\newblock Implicit regularization in deep matrix factorization.
\newblock \emph{NeurIPS}, 2019.

\bibitem[Arpit et~al.(2017)Arpit, Jastrzebski, Ballas, Krueger, Bengio, Kanwal, Maharaj, Fischer, Courville, Bengio, et~al.]{arpit2017closer}
Devansh Arpit, Stanislaw Jastrzebski, Nicolas Ballas, David Krueger, Emmanuel Bengio, Maxinder~S Kanwal, Tegan Maharaj, Asja Fischer, Aaron Courville, Yoshua Bengio, et~al.
\newblock A closer look at memorization in deep networks.
\newblock In \emph{ICML}. PMLR, 2017.

\bibitem[Bell and Sagun(2023)]{bell2023simplicity}
Samuel~James Bell and Levent Sagun.
\newblock Simplicity bias leads to amplified performance disparities.
\newblock In \emph{Proceedings of the 2023 ACM Conference on Fairness, Accountability, and Transparency}, pages 355--369, 2023.

\bibitem[Bhattamishra et~al.(2022)Bhattamishra, Patel, Kanade, and Blunsom]{bhattamishra2022simplicity}
Satwik Bhattamishra, Arkil Patel, Varun Kanade, and Phil Blunsom.
\newblock Simplicity bias in transformers and their ability to learn sparse boolean functions.
\newblock \emph{arXiv preprint arXiv:2211.12316}, 2022.

\bibitem[Bingham et~al.(2020)Bingham, Macke, and Miikkulainen]{bingham2020evolutionary}
Garrett Bingham, William Macke, and Risto Miikkulainen.
\newblock Evolutionary optimization of deep learning activation functions.
\newblock In \emph{Genetic and Evolutionary Computation Conference}, 2020.

\bibitem[Buchanan(2018)]{buchanan2018natural}
Mark Buchanan.
\newblock A natural bias for simplicity.
\newblock \emph{Nature Physics}, 2018.

\bibitem[Chelly et~al.(2024)Chelly, Finder, Ifergane, and Freifeld]{chelly2024trainable}
Irit Chelly, Shahaf~E Finder, Shira Ifergane, and Oren Freifeld.
\newblock Trainable highly-expressive activation functions.
\newblock \emph{arXiv preprint arXiv:2407.07564}, 2024.

\bibitem[Cohen and Shashua(2016)]{cohen2016inductive}
Nadav Cohen and Amnon Shashua.
\newblock Inductive bias of deep convolutional networks through pooling geometry.
\newblock \emph{arXiv preprint arXiv:1605.06743}, 2016.

\bibitem[De~Palma et~al.(2019)De~Palma, Kiani, and Lloyd]{de2019random}
Giacomo De~Palma, Bobak Kiani, and Seth Lloyd.
\newblock Random deep neural networks are biased towards simple functions.
\newblock \emph{NeurIPS}, 2019.

\bibitem[Deutsch(2011)]{deutsch2011beginning}
David Deutsch.
\newblock \emph{The beginning of infinity: Explanations that transform the world}.
\newblock Penguin, 2011.

\bibitem[Dherin et~al.(2022)Dherin, Munn, Rosca, and Barrett]{dherin2022neural}
Benoit Dherin, Michael Munn, Mihaela Rosca, and David Barrett.
\newblock Why neural networks find simple solutions: The many regularizers of geometric complexity.
\newblock \emph{NeurIPS}, 35, 2022.

\bibitem[Dingle et~al.(2018)Dingle, Camargo, and Louis]{dingle2018input}
Kamaludin Dingle, Chico~Q Camargo, and Ard~A Louis.
\newblock Input--output maps are strongly biased towards simple outputs.
\newblock \emph{Nature communications}, 2018.

\bibitem[Domingos(1999)]{domingos1999role}
Pedro Domingos.
\newblock The role of occam's razor in knowledge discovery.
\newblock \emph{Data mining and knowledge discovery}, 1999.

\bibitem[Dragoi et~al.(2024)Dragoi, Gogianu, and Burceanu]{dragoiclosing}
Marius Dragoi, Florin Gogianu, and Elena Burceanu.
\newblock Closing the gap on tabular data with fourier and implicit categorical features.
\newblock \emph{Submission to ICLR (available on OpenReview)}, 2024.

\bibitem[Dubey et~al.(2022)Dubey, Singh, and Chaudhuri]{dubey2022activation}
Shiv~Ram Dubey, Satish~Kumar Singh, and Bidyut~Baran Chaudhuri.
\newblock Activation functions in deep learning: A comprehensive survey and benchmark.
\newblock \emph{Neurocomputing}, 2022.

\bibitem[Ducotterd et~al.(2024)Ducotterd, Goujon, Bohra, Perdios, Neumayer, and Unser]{ducotterd2024improving}
Stanislas Ducotterd, Alexis Goujon, Pakshal Bohra, Dimitris Perdios, Sebastian Neumayer, and Michael Unser.
\newblock Improving lipschitz-constrained neural networks by learning activation functions.
\newblock \emph{Journal of Machine Learning Research}, 2024.

\bibitem[Elton(2021)]{elton2021applying}
Daniel~C Elton.
\newblock Applying deutsch’s concept of good explanations to artificial intelligence and neuroscience--an initial exploration.
\newblock \emph{Cognitive Systems Research}, 2021.

\bibitem[Farebrother et~al.(2024)Farebrother, Orbay, Vuong, Ta{\"\i}ga, Chebotar, Xiao, Irpan, Levine, Castro, Faust, et~al.]{farebrother2024stop}
Jesse Farebrother, Jordi Orbay, Quan Vuong, Adrien~Ali Ta{\"\i}ga, Yevgen Chebotar, Ted Xiao, Alex Irpan, Sergey Levine, Pablo~Samuel Castro, Aleksandra Faust, et~al.
\newblock Stop regressing: Training value functions via classification for scalable deep rl.
\newblock \emph{arXiv preprint arXiv:2403.03950}, 2024.

\bibitem[Fiedler(2021)]{fiedler2021simple}
James Fiedler.
\newblock Simple modifications to improve tabular neural networks.
\newblock \emph{CoRR}, abs/2108.03214, 2021.

\bibitem[Fridovich-Keil et~al.(2022)Fridovich-Keil, Gontijo~Lopes, and Roelofs]{fridovich2022spectral}
Sara Fridovich-Keil, Raphael Gontijo~Lopes, and Rebecca Roelofs.
\newblock Spectral bias in practice: The role of function frequency in generalization.
\newblock \emph{NeurIPS}, 2022.

\bibitem[Gallant(1988)]{gallant1988there}
Gallant.
\newblock There exists a neural network that does not make avoidable mistakes.
\newblock In \emph{IEEE International Conference on Neural Networks}. IEEE, 1988.

\bibitem[Gallegos et~al.(2024)Gallegos, Rossi, Barrow, Tanjim, Kim, Dernoncourt, Yu, Zhang, and Ahmed]{gallegos2024bias}
Isabel~O Gallegos, Ryan~A Rossi, Joe Barrow, Md~Mehrab Tanjim, Sungchul Kim, Franck Dernoncourt, Tong Yu, Ruiyi Zhang, and Nesreen~K Ahmed.
\newblock Bias and fairness in large language models: A survey.
\newblock \emph{Computational Linguistics}, 2024.

\bibitem[Geirhos et~al.(2020)Geirhos, Jacobsen, Michaelis, Zemel, Brendel, Bethge, and Wichmann]{geirhos2020shortcut}
Robert Geirhos, J{\"o}rn-Henrik Jacobsen, Claudio Michaelis, Richard Zemel, Wieland Brendel, Matthias Bethge, and Felix~A Wichmann.
\newblock Shortcut learning in deep neural networks.
\newblock \emph{Nature Machine Intelligence}, 2020.

\bibitem[Gogianu et~al.(2021)Gogianu, Berariu, Rosca, Clopath, Busoniu, and Pascanu]{gogianu2021spectral}
Florin Gogianu, Tudor Berariu, Mihaela~C Rosca, Claudia Clopath, Lucian Busoniu, and Razvan Pascanu.
\newblock Spectral normalisation for deep reinforcement learning: an optimisation perspective.
\newblock In \emph{ICML}, 2021.

\bibitem[Goldblum et~al.(2023)Goldblum, Finzi, Rowan, and Wilson]{goldblum2023no}
Micah Goldblum, Marc Finzi, Keefer Rowan, and Andrew~Gordon Wilson.
\newblock The no free lunch theorem, kolmogorov complexity, and the role of inductive biases in machine learning.
\newblock \emph{arXiv preprint arXiv:2304.05366}, 2023.

\bibitem[Gorishniy et~al.(2022)Gorishniy, Rubachev, and Babenko]{gorishniy2022embeddings}
Yury Gorishniy, Ivan Rubachev, and Artem Babenko.
\newblock On embeddings for numerical features in tabular deep learning.
\newblock \emph{NeurIPS}, 2022.

\bibitem[Goyal and Bengio(2022)]{goyal2022inductive}
Anirudh Goyal and Yoshua Bengio.
\newblock Inductive biases for deep learning of higher-level cognition.
\newblock \emph{Proceedings of the Royal Society A}, 2022.

\bibitem[Grinsztajn(2022)]{grinTabBenchmark}
L{\'e}o Grinsztajn.
\newblock Tabular data learning benchmark.
\newblock \emph{https://github.com/LeoGrin/tabular-benchmark}, 2022.

\bibitem[Grinsztajn et~al.(2022)Grinsztajn, Oyallon, and Varoquaux]{grinsztajn2022tree}
L{\'e}o Grinsztajn, Edouard Oyallon, and Ga{\"e}l Varoquaux.
\newblock Why do tree-based models still outperform deep learning on typical tabular data?
\newblock \emph{NeurIPS}, 2022.

\bibitem[Gromov(2023)]{gromov2023grokking}
Andrey Gromov.
\newblock Grokking modular arithmetic.
\newblock \emph{arXiv preprint arXiv:2301.02679}, 2023.

\bibitem[Hahn et~al.(2021)Hahn, Jurafsky, and Futrell]{hahn2021sensitivity}
Michael Hahn, Dan Jurafsky, and Richard Futrell.
\newblock Sensitivity as a complexity measure for sequence classification tasks.
\newblock \emph{Transactions of the ACL}, 2021.

\bibitem[Harutyunyan et~al.(2024)Harutyunyan, Darbinyan, Karapetyan, and Khachatrian]{harutyunyan2024context}
Hrayr Harutyunyan, Rafayel Darbinyan, Samvel Karapetyan, and Hrant Khachatrian.
\newblock In-context learning in presence of spurious correlations.
\newblock \emph{arXiv preprint arXiv:2410.03140}, 2024.

\bibitem[Hendrycks and Gimpel(2016)]{hendrycks2016gaussian}
Dan Hendrycks and Kevin Gimpel.
\newblock Gaussian error linear units ({GeLUs}).
\newblock \emph{arXiv preprint arXiv:1606.08415}, 2016.

\bibitem[Hermann and Lampinen(2020)]{hermann2020shapes}
Katherine~L Hermann and Andrew~K Lampinen.
\newblock What shapes feature representations? exploring datasets, architectures, and training.
\newblock \emph{arXiv preprint arXiv:2006.12433}, 2020.

\bibitem[Hosseini et~al.(2018)Hosseini, Xiao, Jaiswal, and Poovendran]{hosseini2018assessing}
Hossein Hosseini, Baicen Xiao, Mayoore Jaiswal, and Radha Poovendran.
\newblock Assessing shape bias property of convolutional neural networks.
\newblock In \emph{Proceedings of the IEEE Conference on Computer Vision and Pattern Recognition Workshops}, 2018.

\bibitem[Hui and Belkin(2020)]{hui2020evaluation}
Like Hui and Mikhail Belkin.
\newblock Evaluation of neural architectures trained with square loss vs cross-entropy in classification tasks.
\newblock \emph{arXiv preprint arXiv:2006.07322}, 2020.

\bibitem[Imani et~al.(2024)Imani, Luedemann, Scholnick-Hughes, Elelimy, and White]{imani2024investigating}
Ehsan Imani, Kai Luedemann, Sam Scholnick-Hughes, Esraa Elelimy, and Martha White.
\newblock Investigating the histogram loss in regression.
\newblock \emph{arXiv preprint arXiv:2402.13425}, 2024.

\bibitem[Jagtap and Karniadakis(2023)]{jagtap2023important}
Ameya~D Jagtap and George~Em Karniadakis.
\newblock How important are activation functions in regression and classification? a survey, performance comparison, and future directions.
\newblock \emph{Journal of Machine Learning for Modeling and Computing}, 4\penalty0 (1), 2023.

\bibitem[Jagtap et~al.(2020)Jagtap, Kawaguchi, and Karniadakis]{jagtap2020adaptive}
Ameya~D Jagtap, Kenji Kawaguchi, and George~Em Karniadakis.
\newblock Adaptive activation functions accelerate convergence in deep and physics-informed neural networks.
\newblock \emph{Journal of Computational Physics}, 2020.

\bibitem[Jain et~al.(2024)Jain, Nobahari, Baratin, and Mannelli]{jain2024bias}
Anchit Jain, Rozhin Nobahari, Aristide Baratin, and Stefano~Sarao Mannelli.
\newblock Bias in motion: Theoretical insights into the dynamics of bias in sgd training.
\newblock \emph{arXiv preprint arXiv:2405.18296}, 2024.

\bibitem[Jiang and Teney(2024)]{jiang2024ood}
Liangze Jiang and Damien Teney.
\newblock {OOD}-chameleon: Is algorithm selection for ood generalization learnable?
\newblock \emph{arXiv preprint arXiv:2410.02735}, 2024.

\bibitem[Kalimeris et~al.(2019)Kalimeris, Kaplun, Nakkiran, Edelman, Yang, Barak, and Zhang]{kalimeris2019sgd}
Dimitris Kalimeris, Gal Kaplun, Preetum Nakkiran, Benjamin Edelman, Tristan Yang, Boaz Barak, and Haofeng Zhang.
\newblock {SGD} on neural networks learns functions of increasing complexity.
\newblock \emph{NeurIPS}, 2019.

\bibitem[Krizhevsky and Hinton(2009)]{krizhevsky2009learning}
Alex Krizhevsky and Geoffrey Hinton.
\newblock Learning multiple layers of features from tiny images.
\newblock Technical report, University of Toronto, 2009.

\bibitem[Kuka{\v{c}}ka et~al.(2017)Kuka{\v{c}}ka, Golkov, and Cremers]{kukavcka2017regularization}
Jan Kuka{\v{c}}ka, Vladimir Golkov, and Daniel Cremers.
\newblock Regularization for deep learning: A taxonomy.
\newblock \emph{arXiv preprint arXiv:1710.10686}, 2017.

\bibitem[Kumar et~al.(2023)Kumar, Bordelon, Gershman, and Pehlevan]{kumar2023grokking}
Tanishq Kumar, Blake Bordelon, Samuel~J Gershman, and Cengiz Pehlevan.
\newblock Grokking as the transition from lazy to rich training dynamics.
\newblock \emph{arXiv preprint arXiv:2310.06110}, 2023.

\bibitem[LeCun et~al.(1998)LeCun, Bottou, Bengio, and Haffner]{lecun1998gradient}
Yann LeCun, L{\'e}on Bottou, Yoshua Bengio, and Patrick Haffner.
\newblock Gradient-based learning applied to document recognition.
\newblock \emph{Proceedings of the IEEE}, 86\penalty0 (11):\penalty0 2278--2324, 1998.

\bibitem[Li and Pathak(2021)]{li2021functional}
Alexander Li and Deepak Pathak.
\newblock Functional regularization for reinforcement learning via learned fourier features.
\newblock \emph{NeurIPS}, 2021.

\bibitem[Li et~al.(2018)Li, Xu, Taylor, Studer, and Goldstein]{li2018visualizing}
Hao Li, Zheng Xu, Gavin Taylor, Christoph Studer, and Tom Goldstein.
\newblock Visualizing the loss landscape of neural nets.
\newblock \emph{NeurIPS}, 2018.

\bibitem[Li et~al.(2024)Li, Wang, and Li]{li2024tree}
Xuan Li, Yun Wang, and Bo Li.
\newblock Tree-regularized tabular embeddings.
\newblock \emph{arXiv preprint arXiv:2403.00963}, 2024.

\bibitem[Lin et~al.(2017)Lin, Tegmark, and Rolnick]{lin2017does}
Henry~W Lin, Max Tegmark, and David Rolnick.
\newblock Why does deep and cheap learning work so well?
\newblock \emph{Journal of Statistical Physics}, 2017.

\bibitem[Liu et~al.(2020)Liu, Cai, and Xu]{liu2020multi}
Ziqi Liu, Wei Cai, and Zhi-Qin~John Xu.
\newblock Multi-scale deep neural network (mscalednn) for solving poisson-boltzmann equation in complex domains.
\newblock \emph{arXiv preprint arXiv:2007.11207}, 2020.

\bibitem[Liu et~al.(2022)Liu, Kitouni, Nolte, Michaud, Tegmark, and Williams]{liu2022towards}
Ziming Liu, Ouail Kitouni, Niklas~S Nolte, Eric Michaud, Max Tegmark, and Mike Williams.
\newblock Towards understanding grokking: An effective theory of representation learning.
\newblock \emph{NeurIPS}, 2022.

\bibitem[Liu et~al.(2024)Liu, Wang, Vaidya, Ruehle, Halverson, Solja{\v{c}}i{\'c}, Hou, and Tegmark]{liu2024kan}
Ziming Liu, Yixuan Wang, Sachin Vaidya, Fabian Ruehle, James Halverson, Marin Solja{\v{c}}i{\'c}, Thomas~Y Hou, and Max Tegmark.
\newblock Kan: Kolmogorov-arnold networks.
\newblock \emph{arXiv preprint arXiv:2404.19756}, 2024.

\bibitem[Lyu et~al.(2021)Lyu, Li, Wang, and Arora]{lyu2021gradient}
Kaifeng Lyu, Zhiyuan Li, Runzhe Wang, and Sanjeev Arora.
\newblock Gradient descent on two-layer nets: Margin maximization and simplicity bias.
\newblock \emph{NeurIPS}, 2021.

\bibitem[Maas et~al.(2013)Maas, Hannun, Ng, et~al.]{maas2013rectifier}
Andrew~L Maas, Awni~Y Hannun, Andrew~Y Ng, et~al.
\newblock Rectifier nonlinearities improve neural network acoustic models.
\newblock In \emph{ICML}, 2013.

\bibitem[Mavor-Parker et~al.(2024)Mavor-Parker, Sargent, Barry, Griffin, and Lyle]{mavor2024frequency}
Augustine~N Mavor-Parker, Matthew~J Sargent, Caswell Barry, Lewis Griffin, and Clare Lyle.
\newblock Frequency and generalisation of periodic activation functions in reinforcement learning.
\newblock \emph{arXiv preprint arXiv:2407.06756}, 2024.

\bibitem[McElfresh et~al.(2024)McElfresh, Khandagale, Valverde, Prasad~C, Ramakrishnan, Goldblum, and White]{mcelfresh2024neural}
Duncan McElfresh, Sujay Khandagale, Jonathan Valverde, Vishak Prasad~C, Ganesh Ramakrishnan, Micah Goldblum, and Colin White.
\newblock When do neural nets outperform boosted trees on tabular data?
\newblock \emph{NeurIPS}, 2024.

\bibitem[Mingard et~al.(2019)Mingard, Skalse, Valle-P{\'e}rez, Mart{\'\i}nez-Rubio, Mikulik, and Louis]{mingard2019neural}
Chris Mingard, Joar Skalse, Guillermo Valle-P{\'e}rez, David Mart{\'\i}nez-Rubio, Vladimir Mikulik, and Ard~A Louis.
\newblock Neural networks are a priori biased towards boolean functions with low entropy.
\newblock \emph{arXiv preprint arXiv:1909.11522}, 2019.

\bibitem[Mingard et~al.(2021)Mingard, Valle-P{\'e}rez, Skalse, and Louis]{mingard2021sgd}
Chris Mingard, Guillermo Valle-P{\'e}rez, Joar Skalse, and Ard~A Louis.
\newblock Is {SGD} a bayesian sampler? well, almost.
\newblock \emph{Journal of Machine Learning Research}, 2021.

\bibitem[Mingard et~al.(2023)Mingard, Rees, Valle-P{\'e}rez, and Louis]{mingard2023deep}
Chris Mingard, Henry Rees, Guillermo Valle-P{\'e}rez, and Ard~A Louis.
\newblock Do deep neural networks have an inbuilt occam's razor?
\newblock \emph{arXiv preprint arXiv:2304.06670}, 2023.

\bibitem[Mitchell(1980)]{mitchell1980need}
Tom~M Mitchell.
\newblock The need for biases in learning generalizations.
\newblock \emph{Rutgers University CS tech report CBM-TR-117}, 1980.

\bibitem[Netzer et~al.(2011)Netzer, Wang, Coates, Bissacco, Wu, and Ng]{netzer2011reading}
Yuval Netzer, Tao Wang, Adam Coates, Alessandro Bissacco, Bo Wu, and Andrew~Y Ng.
\newblock Reading digits in natural images with unsupervised feature learning.
\newblock \emph{NIPS Workshop on Deep Learning and Unsupervised Feature Learning}, 2011.

\bibitem[Neyshabur et~al.(2014)Neyshabur, Tomioka, and Srebro]{neyshabur2014search}
Behnam Neyshabur, Ryota Tomioka, and Nathan Srebro.
\newblock In search of the real inductive bias: On the role of implicit regularization in deep learning.
\newblock \emph{arXiv preprint arXiv:1412.6614}, 2014.

\bibitem[Pezeshki et~al.(2021)Pezeshki, Kaba, Bengio, Courville, Precup, and Lajoie]{pezeshki2021gradient}
Mohammad Pezeshki, Oumar Kaba, Yoshua Bengio, Aaron~C Courville, Doina Precup, and Guillaume Lajoie.
\newblock Gradient starvation: A learning proclivity in neural networks.
\newblock \emph{NeurIPS}, 2021.

\bibitem[Poggio et~al.(2018)Poggio, Kawaguchi, Liao, Miranda, Rosasco, Boix, Hidary, and Mhaskar]{poggio2018theory}
Tomaso Poggio, Kenji Kawaguchi, Qianli Liao, Brando Miranda, Lorenzo Rosasco, Xavier Boix, Jack Hidary, and Hrushikesh Mhaskar.
\newblock Theory of deep learning {III}: the non-overfitting puzzle.
\newblock \emph{CBMM Memo}, 2018.

\bibitem[Popper(1959)]{popper1959logic}
Karl Popper.
\newblock \emph{"7. Simplicity". The logic of scientific discovery}.
\newblock Routledge, 1959.

\bibitem[Power et~al.(2022)Power, Burda, Edwards, Babuschkin, and Misra]{power2022grokking}
Alethea Power, Yuri Burda, Harri Edwards, Igor Babuschkin, and Vedant Misra.
\newblock Grokking: Generalization beyond overfitting on small algorithmic datasets.
\newblock \emph{arXiv preprint arXiv:2201.02177}, 2022.

\bibitem[Puli et~al.(2023)Puli, Zhang, Wald, and Ranganath]{puli2023don}
Aahlad~Manas Puli, Lily Zhang, Yoav Wald, and Rajesh Ranganath.
\newblock Don’t blame dataset shift! shortcut learning due to gradients and cross entropy.
\newblock \emph{NeurIPS}, 2023.

\bibitem[Rahaman et~al.(2019)Rahaman, Baratin, Arpit, Draxler, Lin, Hamprecht, Bengio, and Courville]{rahaman2019spectral}
Nasim Rahaman, Aristide Baratin, Devansh Arpit, Felix Draxler, Min Lin, Fred Hamprecht, Yoshua Bengio, and Aaron Courville.
\newblock On the spectral bias of neural networks.
\newblock In \emph{ICML}. PMLR, 2019.

\bibitem[Ramachandran et~al.(2017)Ramachandran, Zoph, and Le]{ramachandran2017swish}
Prajit Ramachandran, Barret Zoph, and Quoc~V Le.
\newblock Swish: a self-gated activation function.
\newblock \emph{arXiv preprint arXiv:1710.05941}, 2017.

\bibitem[Ramasinghe and Lucey(2022)]{ramasinghe2022beyond}
Sameera Ramasinghe and Simon Lucey.
\newblock Beyond periodicity: Towards a unifying framework for activations in coordinate-{MLPs}.
\newblock In \emph{ECCV}. Springer, 2022.

\bibitem[Ramasinghe et~al.(2022)Ramasinghe, MacDonald, and Lucey]{ramasinghe2022frequency}
Sameera Ramasinghe, Lachlan~E MacDonald, and Simon Lucey.
\newblock On the frequency-bias of coordinate-{MLPs}.
\newblock \emph{NeurIPS}, 2022.

\bibitem[Rosca et~al.(2020)Rosca, Weber, Gretton, and Mohamed]{rosca2020case}
Mihaela Rosca, Theophane Weber, Arthur Gretton, and Shakir Mohamed.
\newblock A case for new neural network smoothness constraints.
\newblock \emph{I Can't Believe It's Not Better (ICBINB) Workshop at NeurIPS}, 2020.

\bibitem[Saragadam et~al.(2023)Saragadam, LeJeune, Tan, Balakrishnan, Veeraraghavan, and Baraniuk]{saragadam2023wire}
Vishwanath Saragadam, Daniel LeJeune, Jasper Tan, Guha Balakrishnan, Ashok Veeraraghavan, and Richard~G Baraniuk.
\newblock Wire: Wavelet implicit neural representations.
\newblock In \emph{CVPR}, 2023.

\bibitem[Scardapane et~al.(2019{\natexlab{a}})Scardapane, Scarpiniti, Comminiello, and Uncini]{scardapane2019learning}
Simone Scardapane, Michele Scarpiniti, Danilo Comminiello, and Aurelio Uncini.
\newblock Learning activation functions from data using cubic spline interpolation.
\newblock \emph{Neural Advances in Processing Nonlinear Dynamic Signals}, 2019{\natexlab{a}}.

\bibitem[Scardapane et~al.(2019{\natexlab{b}})Scardapane, Van~Vaerenbergh, Totaro, and Uncini]{scardapane2019kafnets}
Simone Scardapane, Steven Van~Vaerenbergh, Simone Totaro, and Aurelio Uncini.
\newblock Kafnets: Kernel-based non-parametric activation functions for neural networks.
\newblock \emph{Neural Networks}, 2019{\natexlab{b}}.

\bibitem[Schmidhuber(1997)]{schmidhuber1997discovering}
J{\"u}rgen Schmidhuber.
\newblock Discovering neural nets with low kolmogorov complexity and high generalization capability.
\newblock \emph{Neural Networks}, 1997.

\bibitem[Scimeca et~al.(2021)Scimeca, Oh, Chun, Poli, and Yun]{scimeca2021shortcut}
Luca Scimeca, Seong~Joon Oh, Sanghyuk Chun, Michael Poli, and Sangdoo Yun.
\newblock Which shortcut cues will {DNN}s choose? a study from the parameter-space perspective.
\newblock \emph{arXiv preprint arXiv:2110.03095}, 2021.

\bibitem[Shah et~al.(2020)Shah, Tamuly, Raghunathan, Jain, and Netrapalli]{shah2020pitfalls}
Harshay Shah, Kaustav Tamuly, Aditi Raghunathan, Prateek Jain, and Praneeth Netrapalli.
\newblock The pitfalls of simplicity bias in neural networks.
\newblock \emph{NeurIPS}, 2020.

\bibitem[Shi et~al.(2024)Shi, Zhou, and Gu]{shi2024improved}
Kexuan Shi, Xingyu Zhou, and Shuhang Gu.
\newblock Improved implicit neural representation with fourier reparameterized training.
\newblock In \emph{CVPR}, 2024.

\bibitem[Singhal et~al.(2023)Singhal, Goyal, Xu, and Durrett]{singhal2023long}
Prasann Singhal, Tanya Goyal, Jiacheng Xu, and Greg Durrett.
\newblock A long way to go: Investigating length correlations in rlhf.
\newblock \emph{arXiv preprint arXiv:2310.03716}, 2023.

\bibitem[Sitzmann et~al.(2020)Sitzmann, Martel, Bergman, Lindell, and Wetzstein]{sitzmann2020implicit}
Vincent Sitzmann, Julien Martel, Alexander Bergman, David Lindell, and Gordon Wetzstein.
\newblock Implicit neural representations with periodic activation functions.
\newblock \emph{NeurIPS}, 2020.

\bibitem[Soudry et~al.(2018)Soudry, Hoffer, Nacson, Gunasekar, and Srebro]{soudry2018implicit}
Daniel Soudry, Elad Hoffer, Mor~Shpigel Nacson, Suriya Gunasekar, and Nathan Srebro.
\newblock The implicit bias of gradient descent on separable data.
\newblock \emph{The Journal of Machine Learning Research}, 2018.

\bibitem[Stewart et~al.(2023)Stewart, Bach, Berthet, and Vert]{stewart2023regression}
Lawrence Stewart, Francis Bach, Quentin Berthet, and Jean-Philippe Vert.
\newblock Regression as classification: Influence of task formulation on neural network features.
\newblock In \emph{ICML}. PMLR, 2023.

\bibitem[S{\"u}tfeld et~al.(2020)S{\"u}tfeld, Brieger, Finger, F{\"u}llhase, and Pipa]{sutfeld2020adaptive}
Leon~Ren{\'e} S{\"u}tfeld, Flemming Brieger, Holger Finger, Sonja F{\"u}llhase, and Gordon Pipa.
\newblock Adaptive blending units: Trainable activation functions for deep neural networks.
\newblock In \emph{Intelligent Computing: Proceedings of the Computing Conference}. Springer, 2020.

\bibitem[Tachet et~al.(2018)Tachet, Pezeshki, Shabanian, Courville, and Bengio]{tachet2018learning}
Remi Tachet, Mohammad Pezeshki, Samira Shabanian, Aaron Courville, and Yoshua Bengio.
\newblock On the learning dynamics of deep neural networks.
\newblock \emph{arXiv preprint arXiv:1809.06848}, 2018.

\bibitem[Teney et~al.(2021)Teney, Abbasnejad, Lucey, and Hengel]{teney2021evading}
Damien Teney, Ehsan Abbasnejad, Simon Lucey, and Anton van~den Hengel.
\newblock Evading the simplicity bias: Training a diverse set of models discovers solutions with superior ood generalization.
\newblock \emph{arXiv preprint arXiv:2105.05612}, 2021.

\bibitem[Teney et~al.(2022)Teney, Peyrard, and Abbasnejad]{teney2022predicting}
Damien Teney, Maxime Peyrard, and Ehsan Abbasnejad.
\newblock Predicting is not understanding: Recognizing and addressing underspecification in machine learning.
\newblock In \emph{ECCV}. Springer, 2022.

\bibitem[Teney et~al.(2024{\natexlab{a}})Teney, Lin, Oh, and Abbasnejad]{teney2024id}
Damien Teney, Yong Lin, Seong~Joon Oh, and Ehsan Abbasnejad.
\newblock {ID} and {OOD} performance are sometimes inversely correlated on real-world datasets.
\newblock \emph{NeurIPS}, 2024{\natexlab{a}}.

\bibitem[Teney et~al.(2024{\natexlab{b}})Teney, Nicolicioiu, Hartmann, and Abbasnejad]{teney2024neural}
Damien Teney, Armand~Mihai Nicolicioiu, Valentin Hartmann, and Ehsan Abbasnejad.
\newblock Neural redshift: Random networks are not random functions.
\newblock In \emph{CVPR}, 2024{\natexlab{b}}.

\bibitem[Tieleman(2012)]{tieleman2012lecture}
Tijmen Tieleman.
\newblock Lecture 6.5-rmsprop: Divide the gradient by a running average of its recent magnitude.
\newblock \emph{COURSERA: Neural networks for machine learning}, 4\penalty0 (2):\penalty0 26, 2012.

\bibitem[Tsoy and Konstantinov(2024)]{tsoy2024simplicity}
Nikita Tsoy and Nikola Konstantinov.
\newblock Simplicity bias of two-layer networks beyond linearly separable data.
\newblock \emph{arXiv preprint arXiv:2405.17299}, 2024.

\bibitem[Valle-Perez et~al.(2018)Valle-Perez, Camargo, and Louis]{valle2018deep}
Guillermo Valle-Perez, Chico~Q Camargo, and Ard~A Louis.
\newblock Deep learning generalizes because the parameter-function map is biased towards simple functions.
\newblock \emph{arXiv preprint arXiv:1805.08522}, 2018.

\bibitem[White et~al.(2023)White, Safari, Sukthanker, Ru, Elsken, Zela, Dey, and Hutter]{white2023neural}
Colin White, Mahmoud Safari, Rhea Sukthanker, Binxin Ru, Thomas Elsken, Arber Zela, Debadeepta Dey, and Frank Hutter.
\newblock Neural architecture search: Insights from 1000 papers.
\newblock \emph{arXiv preprint arXiv:2301.08727}, 2023.

\bibitem[Wolpert(2002)]{wolpert2002supervised}
David~H Wolpert.
\newblock The supervised learning no-free-lunch theorems.
\newblock \emph{Soft computing and industry: Recent applications}, 2002.

\bibitem[Xiao et~al.(2017)Xiao, Rasul, and Vollgraf]{xiao2017fashion}
Han Xiao, Kashif Rasul, and Roland Vollgraf.
\newblock Fashion-mnist: A novel image dataset for benchmarking machine learning algorithms.
\newblock \emph{arXiv preprint arXiv:1708.07747}, 2017.

\bibitem[Xie et~al.(2022)Xie, Takikawa, Saito, Litany, Yan, Khan, Tombari, Tompkin, Sitzmann, and Sridhar]{xie2022neural}
Yiheng Xie, Towaki Takikawa, Shunsuke Saito, Or Litany, Shiqin Yan, Numair Khan, Federico Tombari, James Tompkin, Vincent Sitzmann, and Srinath Sridhar.
\newblock Neural fields in visual computing and beyond.
\newblock In \emph{Computer Graphics Forum}. Wiley Online Library, 2022.

\bibitem[Xu et~al.(2019{\natexlab{a}})Xu, Zhang, Luo, Xiao, and Ma]{xu2019frequency}
Zhi-Qin~John Xu, Yaoyu Zhang, Tao Luo, Yanyang Xiao, and Zheng Ma.
\newblock Frequency principle: Fourier analysis sheds light on deep neural networks.
\newblock \emph{arXiv preprint arXiv:1901.06523}, 2019{\natexlab{a}}.

\bibitem[Xu et~al.(2019{\natexlab{b}})Xu, Zhang, and Xiao]{xu2019training}
Zhi-Qin~John Xu, Yaoyu Zhang, and Yanyang Xiao.
\newblock Training behavior of deep neural network in frequency domain.
\newblock In \emph{ICONIP}. Springer, 2019{\natexlab{b}}.

\bibitem[Xu et~al.(2024)Xu, Zhang, and Luo]{xu2024overview}
Zhi-Qin~John Xu, Yaoyu Zhang, and Tao Luo.
\newblock Overview frequency principle/spectral bias in deep learning.
\newblock \emph{Communications on Applied Mathematics and Computation}, 2024.

\bibitem[Yang et~al.(2022)Yang, Ajay, and Agrawal]{yang2022overcoming}
Ge Yang, Anurag Ajay, and Pulkit Agrawal.
\newblock Overcoming the spectral bias of neural value approximation.
\newblock \emph{arXiv preprint arXiv:2206.04672}, 2022.

\bibitem[Zhong and Andreas(2024)]{zhong2024algorithmic}
Ziqian Zhong and Jacob Andreas.
\newblock Algorithmic capabilities of random transformers.
\newblock \emph{arXiv preprint arXiv:2410.04368}, 2024.

\bibitem[Zhou et~al.(2023)Zhou, Bradley, Littwin, Razin, Saremi, Susskind, Bengio, and Nakkiran]{zhou2023algorithms}
Hattie Zhou, Arwen Bradley, Etai Littwin, Noam Razin, Omid Saremi, Josh Susskind, Samy Bengio, and Preetum Nakkiran.
\newblock What algorithms can transformers learn? a study in length generalization.
\newblock \emph{arXiv preprint arXiv:2310.16028}, 2023.

\bibitem[Zhou et~al.(2024)Zhou, Zhang, and Xu]{zhou2024rationale}
Zhangchen Zhou, Yaoyu Zhang, and Zhi-Qin~John Xu.
\newblock A rationale from frequency perspective for grokking in training neural network.
\newblock \emph{arXiv preprint arXiv:2405.17479}, 2024.

\end{thebibliography}
}

\clearpage
\setcounter{page}{1}
\appendix
\maketitlesupplementary

\section{Reviewers' FAQ}
This section contains interesting questions raised
during the review of this paper (paraphrased) and our answers.

\paragraph{Why use MLPs instead of CNNs or ViTs for example?}
The choice of \textbf{unstructured MLPs} is deliberate.
Since the primary goal is to discover 
optimal inductive biases via optimization,
it makes sense to start with architectures that impose little initial constraints.

\paragraph{Can the proposed method for learning activation functions be applied to other architectures?}
In principle yes, but the bi-level optimization is expensive.
We did not attempt to use it with large models.
This method is meant as an exploratory tool,
and the insights it delivered are much more fundamental.
They could serve in the design/selection of future architectures
independently of this 
optimization method.
For example,
\citet[Fig.\ 5]{teney2024neural}
already evaluated how various components (e.g.\ attention)
can nudge inductive biases in ways similar to activation functions.

\paragraph{Why is the scale of TV values different across datasets?}
TV values are not comparable across datasets because 
of the different distances between data points in input space.

\paragraph{Are learned activation functions more akin to pre-trained initialization than architecture choices?}
Not really,
because initializations can vanish with enough iterations of fine-tuning, while
the effects of activation functions remain.
However, it is true that parametrized activations carry more information than
typical architecture choices.

\section{Related Work}
\label{sec:relatedWork}

\paragraph{Inductive biases in deep learning}
are due to choices of architecture~\cite{goyal2022inductive} and of the learning algorithm (optimizer, objective, regularizers~\cite{kukavcka2017regularization}).
We focus on the former.
The simplicity bias has been studied from both aspects.
Most explanations attribute it to loss functions~\cite{pezeshki2021gradient}
and gradient descent
\cite{arora2019implicit,hermann2020shapes,lyu2021gradient,tachet2018learning}.
But work on untrained networks shows 
that it can be explained with architectures alone~\cite{de2019random,goldblum2023no,mingard2019neural,teney2024neural,valle2018deep}.
\citet{teney2024neural} showed that the choice of activation function can modulate the simplicity bias.
The \textbf{spectral bias}~\cite{rahaman2019spectral,kalimeris2019sgd}
or frequency principle~\cite{xu2019frequency}
is a related but different
effect related to training dynamics:
NNs approximate low-frequency components of the target function earlier during training with SGD.

\paragraph{Suitability of the simplicity bias.}
The tendency of NNs
to represent simple functions
is thought as the key why overparametrized networks avoid overfitting~\cite{arpit2017closer,poggio2018theory}.
\citet{schmidhuber1997discovering} even proposed to regularize a model's Kolmogorov complexity to improve generalization.
The preference for simplicity aligns with \textbf{Occam's razor},
a philosophical principle whose (absence of) justification has long been debated~\cite[Appendix~A]{mingard2023deep}.
\citet{domingos1999role} discussed 
arguments against Occam's razor for knowledge discovery.

\paragraph{Side-effects of the simplicity bias.}
The simplicity bias is responsible for \textbf{shortcut learning}~\cite{geirhos2020shortcut,puli2023don,teney2021evading}
and for amplifying performance disparities~\cite{bell2023simplicity}.
A vast literature addresses shortcut learning with 
alternative losses~\cite{puli2023don},
architectures~\cite{hosseini2018assessing},
diversification mechanisms~\cite{addepalli2022feature,teney2022predicting,teney2024id}, etc.
No study has however addressed its root cause,
which we pinpoint to architectural choices, activation functions in particular.
The simplicity bias is also detrimental in the use of NNs for \textbf{scientific computing} such as solving PDEs~\cite[Section 5.4]{xu2024overview}.
A solution relevant to activation functions was proposed 
in MscaleDNNs~\cite{liu2020multi} by restricting them to a compact support.
The simplicity bias makes it difficult for
\textbf{implicit neural representations}
to represent sharp image edges for example~\cite{ramasinghe2022frequency}.
The prevailing solution is to replace activation functions
with sines~\cite{sitzmann2020implicit},
Gaussians~\cite{ramasinghe2022beyond}, or wavelets~\cite{saragadam2023wire}.
Fourier features~\cite{shi2024improved}
are another solution, in fact
mathematically equivalent to periodic activations~\cite[Sect.~5]{xie2022neural}.
With \textbf{tabular data},
NNs are known to often perform poorly%
~\cite{dragoiclosing,grinsztajn2022tree}.
Solutions include Fourier features and numerical embeddings~\cite{gorishniy2022embeddings,li2024tree}
which can be seen as special cases of learned activation functions.
In \textbf{reinforcement learning},
a few studies have suggested that the spectral bias of typical architectures may be suboptimal~\cite{li2021functional}
and have experimented with Fourier features~\cite{yang2022overcoming}
and sine activations~\cite{mavor2024frequency}.
These examples support our message that the simplicity bias is not always desirable.
They also support the search for new activation functions to modulate it.


\paragraph{Activation functions}
are key for introducing non-linearities in NNs.
Many options were considered early on, e.g.\ 
sine activations in the Fourier Neural Networks from 1988~\cite{gallant1988there}.
ReLUs are often credited for enabling the rise of deep learning
by avoiding vanishing gradients~\cite{maas2013rectifier}.
However they are also essential in inducing the simplicity bias~\cite{teney2024neural}
which may be just as important. 
The research community has slowly converged towards smooth handcrafted variants of ReLUs
such as GeLUs~\cite{dubey2022activation,hendrycks2016gaussian,ramachandran2017swish}.
Some works proposed to \textbf{learn activation functions}
using extra parameters optimized alongside the weights of the network~\cite{alexandridis2024adaptive,apicella2019simple,apicella2021survey,bingham2020evolutionary,chelly2024trainable,ducotterd2024improving,jagtap2020adaptive,scardapane2019kafnets,sutfeld2020adaptive}.
See \citet{jagtap2023important} for a comprehensive review.
The goal is to better fit the training data
with an activation function that can evolve during training.
In contrast, we use meta learning to find an activation function that induces better inductive biases,
such that training with this \emph{fixed} activation provides better generalization.
This requires bi-level optimization, episodic training, and unbiased parametrization
that allows us to learn activations very different from existing ones.
\textbf{Kolmogorov-Arnold Networks}~\cite{liu2024kan}
parametrize the connections in a NN, which is equivalent to learning different activation functions
across channels and layers.
They use a parametrization as splines similar to ours.
Their benefits in physics-related problems likely result from the alterations to the inductive biases studied in this paper.
Our method differs from \textbf{neural architecture search}~\cite{white2023neural}
in its ability to discover novel activation functions from scratch, rather than selecting
from predefined candidates~\cite{sutfeld2020adaptive}
or restricted parametric functions~\cite{alexandridis2025adaptive}.

\section{Method for Learning Activation Functions}
\label{sec:learningAfsDetails}

This section provides details about the proposed method.

\paragraph{Novelty.}
Our method is designed to support an analysis of inductive biases and their effects in two steps.
\begin{enumerate}[topsep=0pt,itemsep=1pt]
\item \textbf{Learning an activation function} optimized for generalization on a specific dataset.
\item \textbf{Using this new fixed activation function} to train a network ``as usual'', such that the trained model can be analyzed and compared with any other e.g.\ a baseline ReLU architecture.
\end{enumerate}

\noindent
Our method is therefore very different from most existing works about learning activation functions
\cite{alexandridis2024adaptive,apicella2019simple,apicella2021survey,bingham2020evolutionary,chelly2024trainable,ducotterd2024improving,jagtap2020adaptive,scardapane2019kafnets,sutfeld2020adaptive}.
These usually train the model weights and activation function together
for the same objective i.e.\ fitting the training data.
In our formulation,
the activation function is trained for a different objective
i.e.\ maximizing \textit{generalization}.
We exploit this 
in \hyperref[sec:shortcut]{Section~\ref{sec:shortcut}} (Shortcut Learning)
by simulating in-domain (ID) and out-of-distribution (OOD) conditions.
Each setting then learns a different activation function
that prioritizes the learning of different features.

\paragraph{Parametrization as splines.}
We parametrize the learned activation functions as splines
such that we can learn function with arbitrary, irregular shapes if needed.
This contrasts with existing works on the learning of activation functions
that constrain the search
e.g.\ to combinations of existing activations~\cite{sutfeld2020adaptive},
a small MLP~\cite{apicella2019simple},
or other parametric functions~\cite{alexandridis2025adaptive}.
A parametrization as splines
was already used by
\citet{scardapane2019learning}
and in work concurrent to ours on Kolmogorov-Arnold networks~\cite{liu2024kan}.
Some technical details:
\begin{itemize}[topsep=0pt,itemsep=1pt]
\item
The parametrization takes three hyperparameters $n_\textrm{c}$, $a$, $b$.
\item
$n_\textrm{c}$ specifies the number of control points, typically 50.
\item
The control points are spread regularly in the $[a,b]$, typically $[-5,+5]$ to cover typical activation values.
\item
A spline then 
represents
piecewise linear segments
that interpolate the values specified in the parameters
$\bpsi := [\,g_\bpsi(a), \ldots g_\bpsi(b))\,] \in\mathbb{R}^{n_\textrm{c}}$.
Outside $[a,b]$, $g$ extrapolates the values of $g(a)$ and $g(b)$.
\item 
In our exploratory work, we compared this piecewise linear version
with cubic splines, which are smoother but computationally more expensive.
Both performed similarly.
We also compared it with 
a faster nearest-neighbor interpolation of control points. This performed much worse than the piecewise linear version.
\end{itemize}

\paragraph{Implementation of the algorithm.}
\setcounter{algorithm}{0}
We reproduce the complete procedure below.
The model $f_{\btheta,\bpsi}$
represents any chosen architecture
with weights/biases~$\btheta$
and activation functions parametrized by~$\bpsi$.
The gradient updates $\operatorname{GD}(\cdot,\cdot)$
are described as full-batch updates,
but they can be implemented with any optimizer e.g.\ mini-batch SGD or Adam.

The bi-level optimization is expensive since every
outer iteration trains the model from scratch.
We mitigate this as follows.
First, we train small-width models.
\hyperref[sec:tabular]{Section~\ref{sec:tabular}}
shows that the learned activations subsequently
transfer to wider models.
Second, we do not train the model to convergence in the inner loop.
Instead, we progressively increase the number of inner iterations.
This reduces the computational expense and makes the inner task progressively harder.
Third, second-order derivatives 
(i.e.\ backpropagating through the inner gradient updates)
are only computed over the last $t$ inner steps (typically $t{=}5$).
Our exploratory work found this to be better than a complete linearization (no second-order derivatives)
and vastly cheaper than backpropagating through the whole inner loop (which was not even testable at all because of the required GPU memory).

\paragraph{Optimization.}
The optimization of the activation function in
\hyperref[alg]{Algorithm~\ref{alg}}
proved to be a very difficult non-convex problem with many local minima.
We tried various optimizers for the gradient updates
on $\psi$ (SGD with and without momentum, RMSprop).
No option was consistently better.
We also tried to run multiple instances of the inner loop in parallel (with several models initialized differently) to stabilize the gradients $\nabla_\bpsi L$. However this usually provides worse solutions, indicating that exploration is indeed beneficial to avoid local minima.

A simple but effective workaround is to use vanilla gradient descent with restarts,
i.e.\ running \hyperref[alg]{Algorithm~\ref{alg}} with a different:
\begin{itemize}[topsep=0pt,itemsep=1pt]
\item random seed,
\item learning rate to update $\psi$ in $[0.01, 0.2]$,
\item number of control points $n_\textrm{c} \in [50, 400]$,
\item number of  inner steps backpropagated through $t \in [1, 50]$,
\item initialization as zeros or as a ReLU.
\end{itemize}
This is enough to learn slightly different activation functions. We then keep the best one according to its performance on the validation set after using it to retrain a model from scratch (as a fixed activation function) .




\section{Ablations of the Proposed Method}

We evaluate
below the design choices of the
proposed method to learn activation functions.
We perform these experiments on the image regression task with \fmnist
and 1-hidden layer MLPs
We report averages and standard deviations over 10 random seeds.
See \hyperref[sec:regressionExpDetails]{Section~\ref{sec:regressionExpDetails}}
for other experimental details.
See the captions of
\hyperref[tab:variance]{Tables~\ref{tab:variance}}--\ref{tab:outerObj}
for the takeaways of each experiment.

\begin{table}[h!]
  \renewcommand{\tabcolsep}{0.2em}
  \renewcommand{\arraystretch}{1.0}
    \caption{Evaluation of the variance across runs (over 10 random seeds and 4 restarts).
    It is quite similar for the baseline and the learned-activation models.
    The latter models obtain a higher accuracy on average.
    These results verify that the improvements from the learned activations are not simply due to running more trials with more chances of finding a ``lucky run''.
    We also show that the restarts
    (i.e.\ running the optimization with multiple hyperparameters, see \hyperref[sec:learningAfsDetails]{Section~\ref{sec:learningAfsDetails}})
    help find a better solution but are not indispensable to obtain an improvement over the baseline.
    \label{tab:variance}
    }
    \vspace{-6pt}
    \footnotesize
    \begin{tabularx}{\linewidth}{*{1}{>{\arraybackslash}X} c}
        \toprule
        Activation function & Accuracy (\%) \\
        \midrule
         ReLU baseline &  53.1 ~{\scriptsize $\pm$ 0.4}\\
         Learned, \underline{average} across restarts\,/\,hyperparameters & \underline{56.6} ~{\scriptsize $\pm$ 0.7}\\
         Learned, \textbf{best} across restarts\,/\,hyperparameters & \textbf{57.2} ~{\scriptsize $\pm$ 0.5}\\
        \bottomrule
    \end{tabularx}
\end{table}



\begin{table}[h!]
  \renewcommand{\tabcolsep}{0.2em}
  \renewcommand{\arraystretch}{1.0}
    \caption{Evaluation of different interpolation methods to represent learned activation functions.
    The nearest-neighbor interpolation is cheap to evaluate but performs the worst. The linear one (used in all our experiments) is almost identical to the cubic one (in appearance and performance) while being faster to evaluate.
    \label{tab:interp}
    }
    \vspace{-6pt}
    \footnotesize
    \begin{tabularx}{\linewidth}{*{1}{>{\arraybackslash}X} c}
        \toprule
        Activation function & Accuracy (\%) \\
        \midrule
         ReLU &  53.1 ~{\scriptsize $\pm$ 0.4}\\
         Learned, nearest-neighbor interpolation & 55.1 ~{\scriptsize $\pm$ 0.9}\\
         Learned, \underline{linear} spline (default) & \underline{57.2} ~{\scriptsize $\pm$ 0.5}\\
         Learned, \textbf{cubic} spline & \bfseries 57.3 ~{\scriptsize $\pm$ 0.7}\\
        \bottomrule
    \end{tabularx}
    \vspace{-12pt}
\end{table}

\begin{figure}[ht!]
  \centering
  \renewcommand{\tabcolsep}{1.3em}
  \renewcommand{\arraystretch}{1}
  \scriptsize
  \bfseries
  \begin{tabular}{ccc}
  \includegraphics[width=.25\columnwidth]{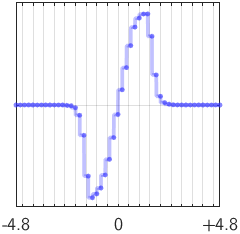} & 
  \includegraphics[width=.25\columnwidth]{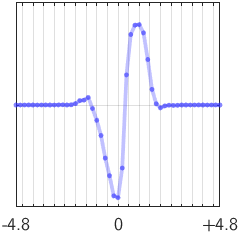} & 
  \includegraphics[width=.25\columnwidth]{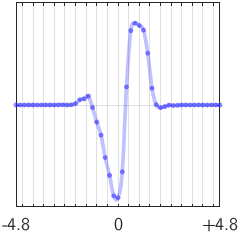}\\[3pt] 
    Nearest-neighbor & Linear spline & Cubic spline
  \end{tabular}
  \vspace{-4pt}
  \caption{\label{fig:interp}
    Activation functions learned with different interpolation methods.
    The linear and cubic ones are nearly identical.
  }
  \vspace{-8pt}
\end{figure}




\begin{table}[h!]
  \renewcommand{\tabcolsep}{0.2em}
  \renewcommand{\arraystretch}{1.0}
    \caption{Evaluation of different outer-loop objectives.
    The naive version simply optimizes the activation for minimum loss on the \emph{training} data, but this is suboptimal.
    Ideally, one would like to optimize the loss on the \emph{test} data (which would require cheating by accessing the test labels).
    We approximate it by optimizing on held-out \emph{validation} data,
    which the results show to be about as good
    (the last row would is expected to be the best without any evaluation noise).
    \label{tab:outerObj}}
    \vspace{-6pt}
    \footnotesize
    \begin{tabularx}{\linewidth}{*{1}{>{\arraybackslash}X} c}
        \toprule
        Activation function & Accuracy (\%) \\
        \midrule
         ReLU &  53.1 ~{\scriptsize $\pm$ 0.4}\\
         Learned to minimize loss on training data (naive) & 56.7 ~{\scriptsize $\pm$ 0.3}\\
         Learned to minimize loss on \textbf{validation} data (default) & \textbf{57.2} ~{\scriptsize $\pm$ 0.5}\\
         Learned to minimize loss on \underline{test} data (cheating) & \underline{56.9} ~{\scriptsize $\pm$ 0.5}\\
        \bottomrule
    \end{tabularx}
    \vspace{-12pt}
\end{table}


\clearpage
\onecolumn
\section{Experimental Details \& Additional Results}
\label{sec:results2}
\label{sec:expDetails}

\subsection{General Experimental Details}
When training MLPs on a given dataset, 
we first \textbf{tune standard hyperparameters} 
for the best validation accuracy using ReLUs 
(optimizer, batch size, learning rate).
We reuse these hyperparameters for all other experiments on this dataset,
i.e.\ we do not tune them again for the learned activation functions.
Every experiment uses \textbf{early stopping} i.e.\ we keep the model at the training step with the best validation accuracy.

All experiments were run on a single laptop (Dell XPS 15)
with an Nvidia RTX\,3050\,Ti (4\,GB of GPU memory).

\paragraph{Variance in the results.}
In order to make the analysis of results stable and consistently reproducible,
we use two interventions that greatly reduce the variance across seeds and training iterations.
First, the models are trained with large- or full-batch gradient descent (typically 4096 examples per mini-batch). This eliminates most of the variation across seeds.
Second, we use a simple stochastic weight averaging (SWA). That is, when evaluating a model, we use the average of the optimized weights over the last 50 training steps.
This consistently improves the accuracy of all models,
but it does not alter the training trajectories (by design)
and we verified that it does not alter the ranking of models.
The main advantage here
is that it greatly stabilizes the performance across training iterations,
i.e.\ the training curves are much smoother hence easier to analyze.

\subsection{Image Datasets}
\label{detailsImage}

\paragraph{Data.}
We use slightly cropped versions of the images in the original datasets.
This makes the data and models smaller and allows us to run a larger number of experiments with limited computational resources.
This makes the tasks slightly more difficult, hence the accuracies being lower baselines reported in prior work.
For \mnist, we crop 5\,pixels on every side. For \svhn, we crop 8\, pixels on each of the the left and right sides.



\paragraph{Architecture.}
We use fully-connected MLP.
Given that our goal is to evaluate the inductive biases induced by the choice of activation function,
MLPs minimizes the possible interactions with other architectural components that would complicate the analysis.

The only improvement over vanilla MLPs is the inclusion of \textbf{residual connections}. After each hidden layer, the output of the activation function is summed with the input to the layer (from before the application of weights and biases). This never hurts the accuracy, and helps when learning different layer-specific activations functions.

\begin{figure}[H]
  \vspace{145pt}
  \centering
  \renewcommand{\tabcolsep}{0.6em}
  \renewcommand{\arraystretch}{1.1}
  \scriptsize
  \bfseries
  \begin{tabular}{rcc}
  ~ & ~~~~~~~~~~~~Classification & ~~~~~~~~~~~~Regression\\
  \makecell[br]{\mnist \\ \vspace{3em}}&
  \includegraphics[clip, trim=0pt 23pt 0pt 0pt, width=.3\columnwidth]{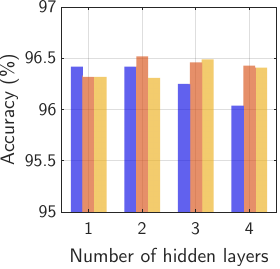}
  &
  \includegraphics[clip, trim=0pt 23pt 0pt 0pt, width=.3\columnwidth]{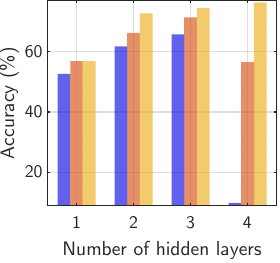}\\
  \makecell[br]{\fmnist \\ \vspace{3em}}&
  \includegraphics[clip, trim=0pt 23pt 0pt 0pt, width=.3\columnwidth]{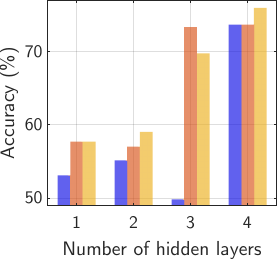}
  &
  \includegraphics[clip, trim=0pt 23pt 0pt 0pt, width=.3\columnwidth]{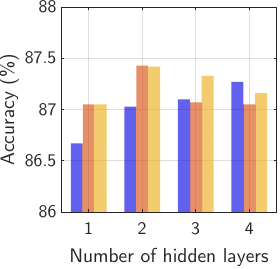}\\
  \makecell[br]{\svhn \\ \vspace{3em}}&
  \includegraphics[clip, trim=0pt 23pt 0pt 0pt, width=.3\columnwidth]{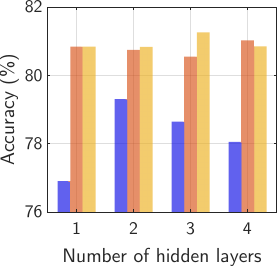}
  &
  \includegraphics[clip, trim=0pt 23pt 0pt 0pt, width=.3\columnwidth]{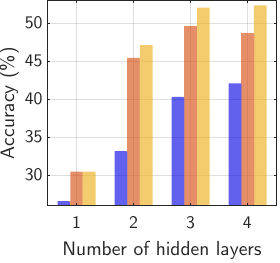}\\
  \makecell[br]{\cifar \\ \vspace{3em}}&
  \includegraphics[width=.3\columnwidth]{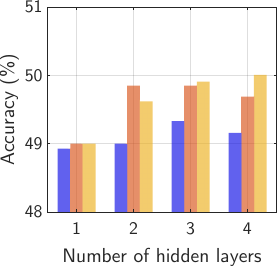}
  &
  \includegraphics[width=.3\columnwidth]{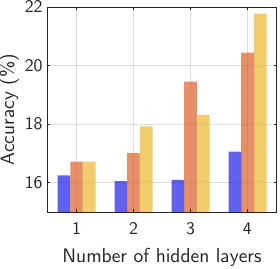}
  \end{tabular}\\[2pt]
  \begin{tabular}{l}
    \makecell[bl]{
      ~~\scriptsize\textcolor[HTML]{6161EF}{\raisebox{0.0em}{\scalebox{1.2}[1]{\ding{110}}}}
      ReLU activations\\
      ~~\scriptsize\textcolor[HTML]{E68F69}{\raisebox{0.0em}{\scalebox{1.2}[1]{\ding{110}}}}
      Learned act., ReLU init.\\
      ~~\scriptsize\textcolor[HTML]{F3CC6E}{\raisebox{0.0em}{\scalebox{1.2}[1]{\ding{110}}}}
      Learned act., zero init.
    }
  \end{tabular}
  \vspace{-6pt}
  \caption{\label{fig:imgNLayersAll}
  Image datasets, 
  results per number of layers.
  }
\end{figure}

For each dataset, we trained MLPs with 1 to 4 hidden layers, both with ReLUs and learned activation functions. Our main results retain the MLP whose depth is best for each activation function.
We provide in \hyperref[fig:imgNLayersAll]{Figure~\ref{fig:imgNLayersAll}}
the full results for every depth. We can see that the best number of layers is sometimes different across activation functions.



\paragraph{Regression tasks.}
\phantomsection
\label{sec:regressionExpDetails}
We use the same data as the image classification tasks.
The ground-truth regression targets are the class IDs $\{0,1,...,9\}$
that we normalized to $[-1,1]$.
I.e. we assign to the classes values regularly spread within $[-1,1]$.
This normalization is standard practice for regression models
to make the optimization numerically easier.
The MLP models output a single scalar with their last layer with no softmax or sigmoid.

\paragraph{Additional results.}
We provide below results 
on all four image datasets.
The main paper only includes results on \mnist for space reasons, but
similar observations can be made on the others.

\clearpage

\begin{multicols}{2}

\begin{figure}[H]
  \vspace{145pt}
  \renewcommand{\tabcolsep}{0.6em}
  \renewcommand{\arraystretch}{1.1}
  \scriptsize
  \bfseries
  \begin{tabular}{rcc}
  ~ & Classification & Regression\\
  \makecell[br]{\mnist \\ \vspace{3em}}&
  \includegraphics[width=.25\columnwidth]{visualizations-mnist-ranged-cropped-bestExps/af-file0016-47--2-va-sn0-nTr1.00.png}
  &
  \includegraphics[width=.25\columnwidth]{visualizations-mnist-ranged-cropped-regression-bestExps/af-file0018-44--4-va-sn0-nTr1.00.png}\\
  \makecell[br]{\fmnist \\ \vspace{3em}}&
  \includegraphics[width=.25\columnwidth]{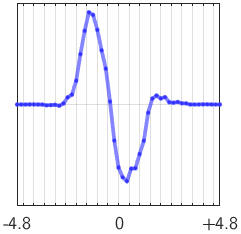}
  &
  \includegraphics[width=.25\columnwidth]{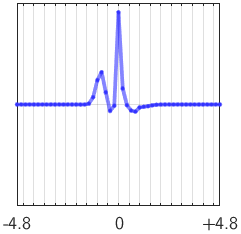}\\
  \makecell[br]{\svhn \\ \vspace{3em}}&
  \includegraphics[width=.25\columnwidth]{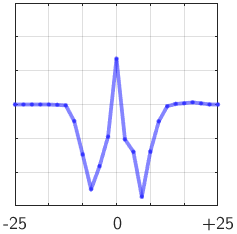}
  &
  \includegraphics[width=.25\columnwidth]{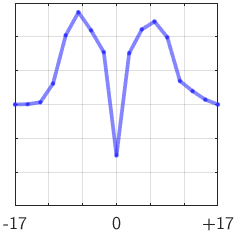}\\
  \makecell[br]{\cifar \\ \vspace{3em}}&
  \includegraphics[width=.25\columnwidth]{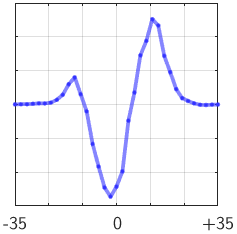}
  &
  \includegraphics[width=.25\columnwidth]{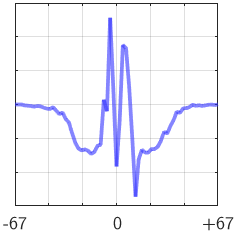}
  \end{tabular}
  \vspace{-8pt}
  \caption{\label{fig:imgActAll}
  Activation functions learned for image datasets treated as classification or regression tasks.
  The activation functions learned for regression contain more irregularities.
  These help networks represent complex functions with sharp transitions.
  }
  \vspace{-8pt}
\end{figure}


\end{multicols}

\clearpage

\begin{figure*}[hpt!]
  \renewcommand{\tabcolsep}{0.9em}
  \renewcommand{\arraystretch}{1.2}
  \scriptsize
  \begin{tabular}{rccccc}
   ~ & \multicolumn{2}{c}{~~~Test vs.\ training accuracy} & ~ &  \multicolumn{2}{c}{~~~Test accuracy vs.\ complexity}\\
   ~ & ~~~~Classification & ~~~~Regression & ~ & ~~~~Classification & ~~~~Regression\\[2pt]
  \makecell[br]{\mnist \\ \vspace{5.5em}}&
  \includegraphics[height=.17\textwidth]{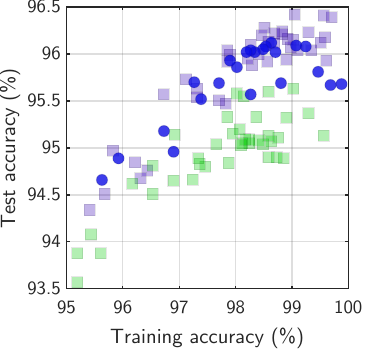}&
  \includegraphics[height=.17\textwidth]{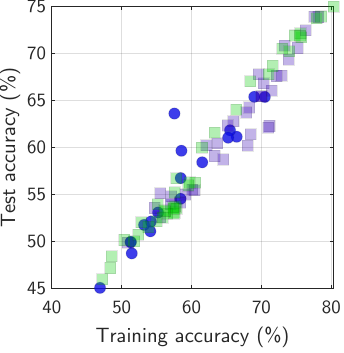}
  &~~~~~~~~~~~&
  \includegraphics[height=.17\textwidth]{visualizations-imgCls/mnist-ranged-cropped_007.pdf}&
  \includegraphics[height=.17\textwidth]{visualizations-imgReg/mnist-ranged-cropped-regression_007.pdf}\\
  \makecell[br]{\fmnist \\ \vspace{5.5em}}&
  \includegraphics[height=.17\textwidth]{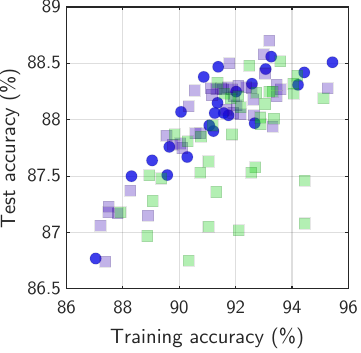}&
  \includegraphics[height=.17\textwidth]{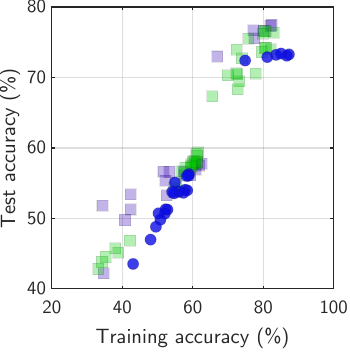}
  &~~~~~~~~~~~&
  \includegraphics[height=.17\textwidth]{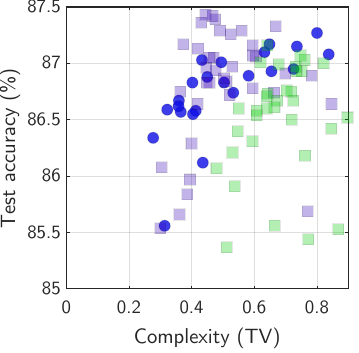}&
  \includegraphics[height=.17\textwidth]{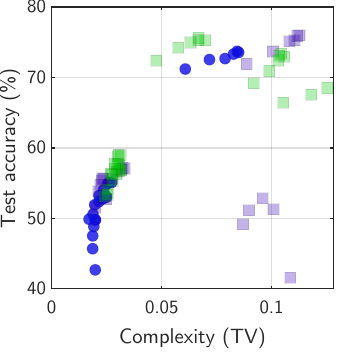}\\
  \makecell[br]{\svhn \\ \vspace{5.5em}}&
  \includegraphics[height=.17\textwidth]{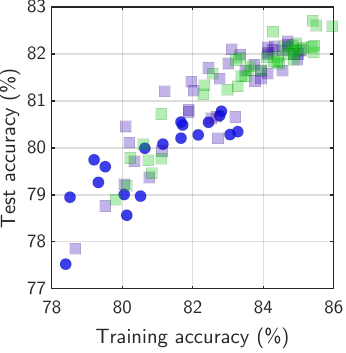}&
  \includegraphics[height=.17\textwidth]{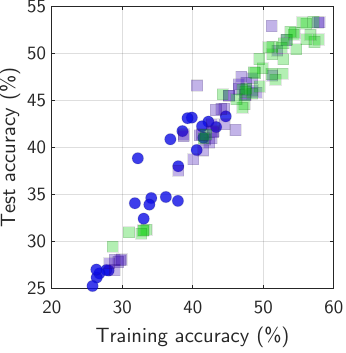}
  &~~~~~~~~~~~&
  \includegraphics[height=.17\textwidth]{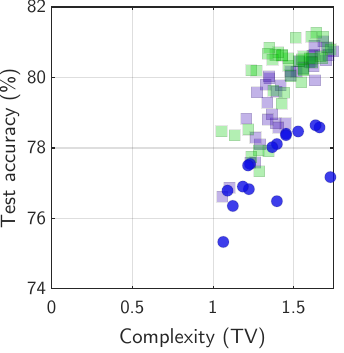}&
  \includegraphics[height=.17\textwidth]{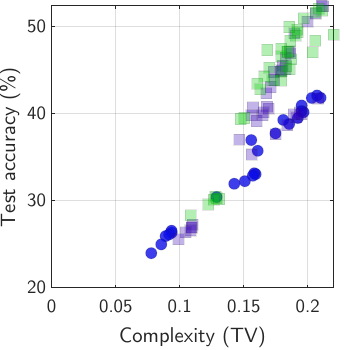}\\
  \makecell[br]{\cifar \\ \vspace{5.5em}}&
  \includegraphics[height=.17\textwidth]{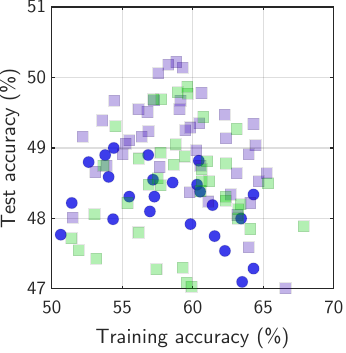}&
  \includegraphics[height=.17\textwidth]{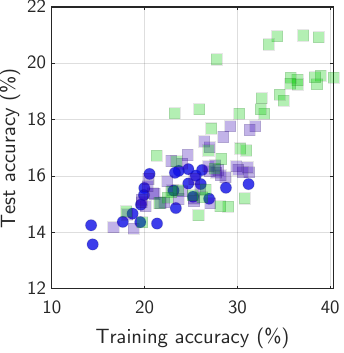}
  &~~~~~~~~~~~&
  \includegraphics[height=.17\textwidth]{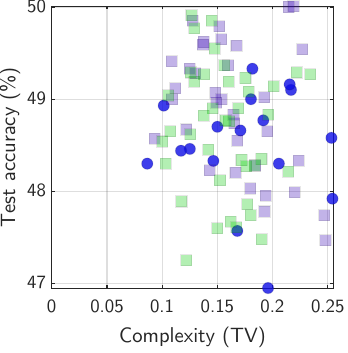}&
  \includegraphics[height=.17\textwidth]{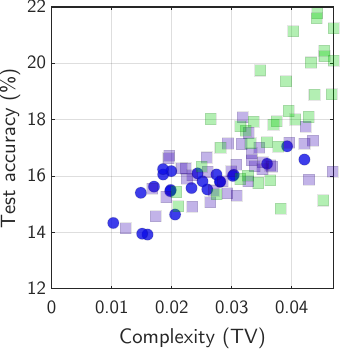}\\
  \end{tabular}
  \vspace{-6pt}
  \caption{
    \label{fig:imgPlotsAll}
    Analysis of models trained on image datasets.
    Each marker represents a model with different hyperparameters or number of training steps,
    and ReLUs
    (\textcolor[HTML]{3D3DEB}{\raisebox{0.0em}{\scalebox{0.75}[0.75]{\ding{108}}}})
    or learned activations with initialization as ReLUs
    (\textcolor[HTML]{C1B4E8}{\raisebox{0.0em}{\scalebox{0.75}[0.75]{\ding{110}}}})    
    or as zeros
    (\textcolor[HTML]{B3F0B3}{\raisebox{0.0em}{\scalebox{0.75}[0.75]{\ding{110}}}}).
        (Left)~tr/test acc, models with learned activations have better accuracy than ReLUs, especially those learned from a random initialization.
  }
\end{figure*}

\clearpage
\onecolumn
\subsection{Transfer of Learned Activation Functions across Image Regression Datasets}
\label{sec:transferImgReg}

We provide below additional results on the transfer of learned activations across datasets, using the image regression tasks
and 4-hidden layer MLPs.
As in most experiments, we train a dataset-specific activation on each dataset (\mnist, \fmnist, \cifar, \svhn) then use each of them to train a different model on each dataset.
This gives a $4\!\times\!4$ matrix of results
(middle rows in \hyperref[tab:transferImgReg]{Table~\ref{tab:transferImgReg}}).
We also attempt to learn an activation function on all datasets simultaneously (last row).
See the table caption of the observations.



\begin{table}[h!]
  \renewcommand{\tabcolsep}{0.5em}
  \renewcommand{\arraystretch}{1.0}
    \caption{Transfer of learned activation functions across image regression datasets.
    The diagonal elements (\textbf{gray cells}) correspond to activation functions optimized for a specific dataset then used to train a model on the same dataset.
    These obviously work well, 
    but other combinations also surpass the ReLU baseline, which indicates positive transfer across datasets.
    The one learned on all datasets (\textbf{last row}) only improves over ReLUs
    on the two harder datasets (\svhn, \cifar)
    and the improvements are (expectedly) smaller than with dataset-specific solutions.
    Further work may be needed to better balance multiple tasks when learning an activation function for multiple datasets.
    \label{tab:transferImgReg}
    }
    \vspace{-6pt}
    \centering
    \footnotesize
    \begin{tabularx}{.9\linewidth}{*{1}{>{\arraybackslash}X} ccccccc}
        \toprule
        ~ & ~~~ & \multicolumn{4}{c}{Accuracy (\%) of models trained on} & ~~~~~~~ & Average $\Delta$\,accuracy \\
        Activation function && ~~~~\mnist~~~~ & \fashionm & ~~~~~\svhn~~~~~ & ~~~~~\cifar~~~~~ & & compared to ReLU \\
        \midrule
        ReLU &&  \underline{76.7}  &  73.9  &  42.9  &  16.1  &&  \,\phantom{$+$}0.0	\phantom{~{\scriptsize $\pm$ ~0.0}} \\
        Learned on \mnist &&  \cellcolor{gray!20} \textbf{79.7}  &  73.0  &  41.0  &  18.2  &&  \textbf{\,$+$0.6}	~{\scriptsize $\pm$ ~2.3} \\
        Learned on \fmnist &&  64.3  &  \cellcolor{gray!20} \textbf{75.1}  &  39.7  &  \textbf{23.7}  &&  $-$1.7	~{\scriptsize $\pm$ ~8.4} \\
        Learned on \svhn &&  61.0  &  73.2  &  \cellcolor{gray!20} \textbf{54.1}  &  19.2  &&  $-$0.5	~{\scriptsize $\pm$ 11.3} \\
        Learned on \cifar &&  57.6  &  \underline{74.5}  &  41.0  &  \cellcolor{gray!20} \underline{22.6}  &&  $-$3.5	~{\scriptsize $\pm$ 11.0} \\
        Learned on all datasets simultaneously &&  76.2  &  72.8  &  \underline{45.0}  &  17.4  &&  \underline{\,$+$0.4}	~{\scriptsize $\pm$ ~1.5} \\
        \bottomrule
    \end{tabularx}
\end{table}


\begin{figure}[hpt!]
  \centering
  \includegraphics[width=.9\columnwidth]{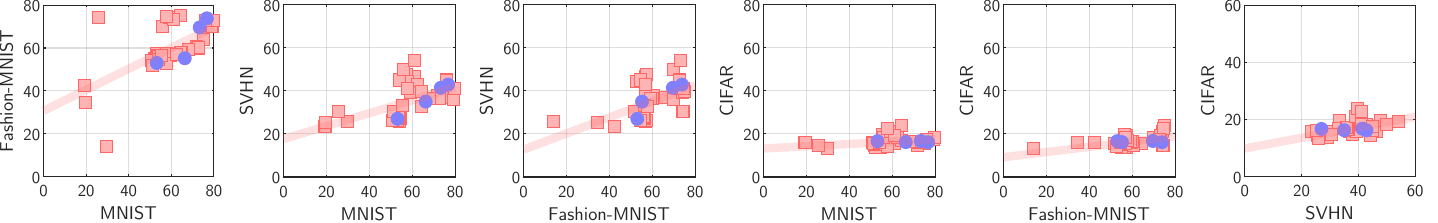}
  \vspace{-6pt}
  \caption{\label{fig:transferImgReg}
  Transfer across image regression datasets.
  Each marker represents an
  MLP architecture with 1 to 4 hidden layers,
  with ReLUs~(\textcolor[HTML]{8080FF}{\raisebox{0.0em}{\scalebox{0.75}[0.75]{\ding{108}}}})
  or activation functions
  learned on each of the four datasets~(\textcolor[HTML]{FF6666}{\raisebox{0.0em}{\scalebox{0.75}[0.75]{\ding{110}}}}).
  We plot the accuracy of each architecture on pairs of datasets
  to show that improvements often correlate across datasets (the line represents the best linear fit to the \textcolor[HTML]{FF6666}{\raisebox{0.0em}{\scalebox{0.75}[0.75]{\ding{110}}}}).
  }
\end{figure}

\clearpage
\onecolumn
\subsection{Tabular Data}
\label{sec:results2Tabular}
\label{detailsTabular}

\begin{figure*}[ht!]
  \centering
  \scriptsize
  \bfseries
  \renewcommand{\tabcolsep}{0.25em}
  \renewcommand{\arraystretch}{1}
  \begin{tabularx}{\textwidth}{*{5}{>{\centering\arraybackslash}X}}
  \includegraphics[height=.15\textwidth]{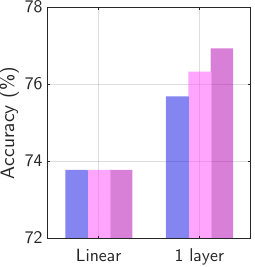}&
  \includegraphics[height=.15\textwidth]{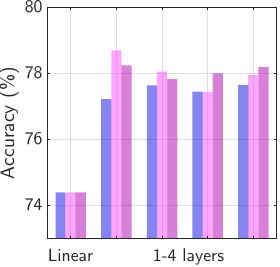}&
  \includegraphics[height=.15\textwidth]{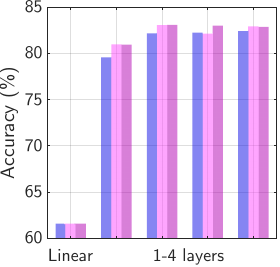}&
  \includegraphics[height=.15\textwidth]{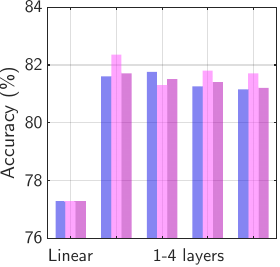}&
  \raisebox{3.6em}{\includegraphics[width=.15\textwidth]{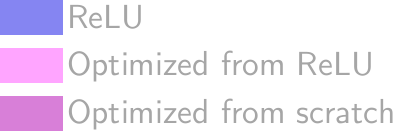}}\\
  Average (all tabular datasets) &
  \phantom{abc} \textsc{electricity} &
  \phantom{abc} \textsc{coverType} &
  \phantom{abc} \textsc{magicTelescope} & ~
  \end{tabularx}
  \vspace{-4pt}
  \caption{\label{fig:tabNLayers}
  (Left) On the tabular datasets, the activation functions learned from scratch (initialized with zeros)
  are usually better than from an initialization as ReLUs.
  But this varies across datasets and the opposite is sometimes true (right).
  Models with learned activation also often perform best with fewer layers than with ReLUs,
  such as on the three datasets pictured.
  }
  \vspace{-4pt}
\end{figure*}

\paragraph{Implementation details}
\begin{itemize}[topsep=0pt,itemsep=4pt]
\item \textbf{Nearest-neighbor (k-NN)}. We use Matlab's implementation
\texttt{fitcknn\!()} with Bayesian hyperparameter optimization for the number of neighbors and the distance measure (L1 or L2).

\item \textbf{Boosted trees}. We use Matlab's implementation
\texttt{fitcensemble\!()} with Bayesian hyperparameter optimization of standard hyperparameters.
All the tabular datasets that we use are binary classification tasks, and the classification trees therefore produce discrete class predictions.
To make the visualizations of ``soft predictions'' as in
\hyperref[fig:tabBoundaries]{Figure~\ref{fig:tabBoundaries}c},
we also train \emph{regression} trees with
\texttt{fit\underline{r}ensemble\!()}, using the class labels in $\{0,1\}$ as regression targets.
The output of these regression trees is then more comparable
to the outputs (logits) from the MLP models.

\item \textbf{Linear model}. We implement this baseline with the same code as our MLP models but with no hidden layer.

\item \textbf{MLP models}. Our models use 1 to 4 hidden layers, a width of 256, and they are trained with RMSprop~\cite{tieleman2012lecture} with large mini-batches of 4096 examples to provide stable and consistent results.
The number of layers is selected for the best validation accuracy for each type of activation function.
The learning rate is also selected for best validation accuracy, but only once with ReLUs then reused for other activation functions.

The performance of our models 
would likely improve with additional hyperparameter tuning. The \textbf{width} alone has a large impact on accuracy, as evaluated in \hyperref[fig:tabBoundaries]{Figures~\ref{fig:tabDim}} and \ref{fig:tabNLayers}.
Our goal is not best absolute performance so we did not expend resources in hyperparameter tuning
and focused on \textbf{like-for-like comparisons} (i.e.\ only changing the activation function).
If anything, our MLP models (and those with learned activation functions in particular) are at a disadvantage compared to the baselines.

\item The \textbf{TanH activation functions with a prefactor}
follow \citet{teney2024neural}. They are simple TanH functions with a multiplier:
$\operatorname{tanh}(\alpha x)$.
The scalar $\alpha \in [0.01, 8]$ is tuned for the best validation accuracy and shared across layers.
The learning rate $\lambda$, which is originally tuned on the ReLU model as mentioned above, is adjusted 
as $\lambda \!\leftarrow\! \lambda / \alpha$.

\item \textbf{Data normalization}.
For every tabular dataset, we normalize the data (shift and scale) such that every input dimension (``column'' in tabular terms) occupies the $[-1,1]$ range.
We experimented with other options: a normalization to unit standard deviation, and a quantile normalization to approximate a Gaussian or uniform distribution for every dimension.
However they produced disparate results across our 16 datasets, so we settled with the simplest option
to keep things consistent.
If absolute performance is the objective, this should be optimized for each dataset.
It has a large effect on the accuracy of MLPs, but not of trees nor k-NN classifiers. So again, our models are likely at a disadvantage compared to the baselines.
\end{itemize}


\paragraph{Details on the visualizations}
\begin{itemize}
\item
In 
\hyperref[fig:tabBoundaries]{Figure~\ref{fig:tabBoundaries}c},
the grayscale images are produced
by evaluating each model on $200\!\times\!200\!=\!40,000$
points in a 2D slice of the input space defined as follows.
We first select one training example $\bx$ at random.
We then select two input dimensions $m$, $n$ at random.
We create every point of the slice by replacing the $m$th and $n$th values (the scalars $x[m]$ and $x[n]$).
of $\bx$ by every possible value 
in a grid of $200\!\times\!200$ values over $[-1,1]\times[-1,1]$.
Since our data is normalized such that every dimension covers $[-1,1]$, we now have a slice of inputs in a realistic range.
This also explains why the training examples, marked by \coloredBullet in \hyperref[fig:tabBoundaries]{Figure~\ref{fig:tabBoundaries}c}
are not centered in the images. They would be centered only if $x[m]\!=\!x[n]\!=\!0$.

The values plotted as a grayscale image are the network's output before a softmax/sigmoid activation.
These values are not bounded to a specific range, so we scale them in each image to fill the black $\rightarrow$ white range.

\item
In \hyperref[fig:tabBoundaries]{Figure~\ref{fig:tabBoundaries}a and b},
the loss and complexity landscapes are produced by
evaluating models over a
$50\!\times\!50$ grid covering a 2D slice of the parameter space (weights and biases).
The slices are chosen to align with the first two principal directions of the trajectory.
We obtain them by computing the PCA of a matrix made of the parameters from a number of checkpoints recorded over the training of the model.
The $50\!\times\!50$ sets of parameters are obtained by applying perturbations to the trained model along these two directions.
For each such set of parameters, we evaluate the model's training loss
and its complexity
(\hyperref[sec:tv]{Section~\ref{sec:tv}})
to make the loss and complexity landscapes.
The range of loss/complexity values is consistent across all the visualization of a given dataset
(i.e.\ a given color represents the same level of complexity across all 
plots in
\hyperref[fig:tabBoundaries]{Figure~\ref{fig:tabBoundaries}b} for example).
\end{itemize}

\paragraph{Intuition for the ``input activation functions'' (IAFs).}
The IAFs are activation functions that are applied directly on the input data, before it is passed to a standard MLP.
The key is that these IAFs are applied independently to each dimension, such that they can
implement a different behavior for each dimension (or ``column'' of the data).
This is particularly useful for tabular data because every dimension can represent a different type of information.
In comparison, once the data is passed through the first layer of an MLP, the dimensions are all mixed together, and the subsequent activation function(s) are applied similarly to every dimension of the hidden representations.

The property of tabular datasets of requiring little or sparse feature interaction
is well known and has been exploited in prior architectures designed for tabular data, see
e.g.\ \citet{gorishniy2022embeddings}.
This property is also a likely reason why decision trees are well suited to tabular data,
since they implement decision boundaries aligned with dimensions of the data.

\begin{figure*}[ht!]
  \centering
  \renewcommand{\tabcolsep}{0em}
  \renewcommand{\arraystretch}{1}
  \scriptsize
  \begin{tabularx}{\textwidth}{*{8}{>{\centering\arraybackslash}X}}
    \textsc{credit}&
    \textsc{electricity}&
    \textsc{covertype}&
    \textsc{pol}&
    \textsc{house16h}&
    \textsc{magictelescope}&
    \textsc{bankMarketing}&
    \textsc{miniboone}\\
  \includegraphics[width=.123\textwidth]{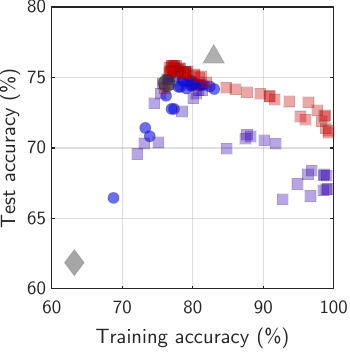}&
  \includegraphics[width=.123\textwidth]{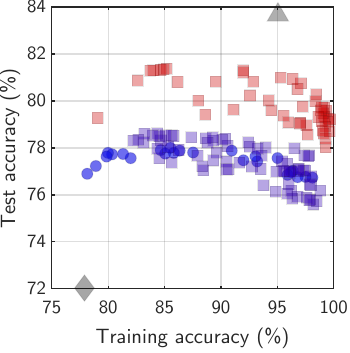}&
  \includegraphics[width=.123\textwidth]{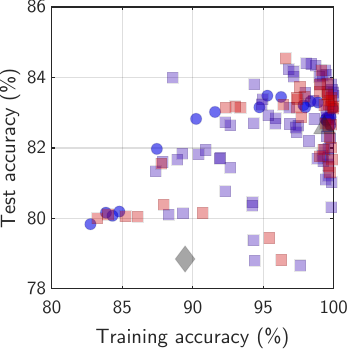}&
  \includegraphics[width=.123\textwidth]{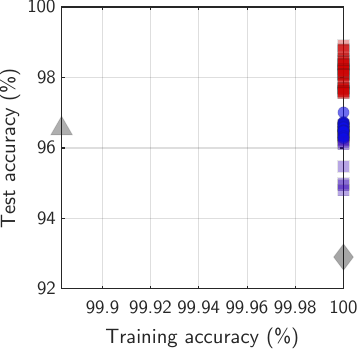}&
  \includegraphics[width=.123\textwidth]{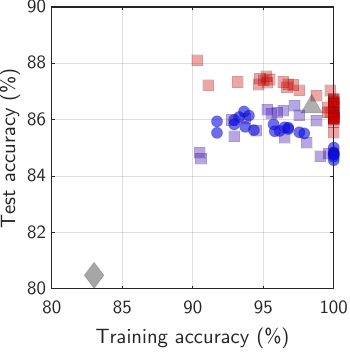}&
  \includegraphics[width=.123\textwidth]{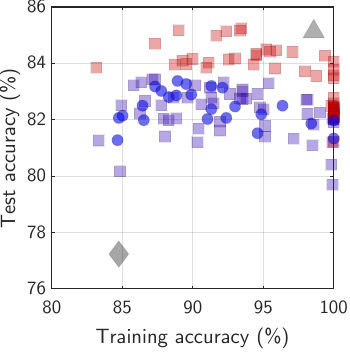}&
  \includegraphics[width=.123\textwidth]{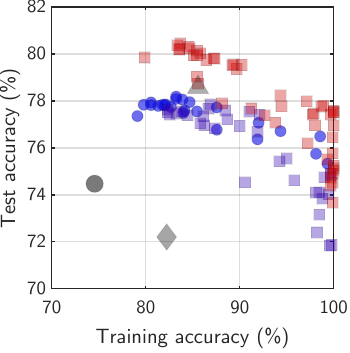}&
  \includegraphics[width=.123\textwidth]{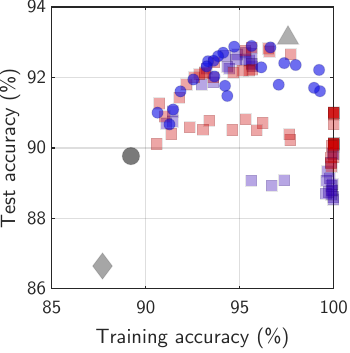}\\[5pt]
    \textsc{higgs}&
    \textsc{eyeMovements}&
    \textsc{diabetes130us}&
    \textsc{jannis}&
    \textsc{defaultOfCC}&
    \textsc{bioresponse}&
    \textsc{california}&
    \textsc{heloc}\\
  \includegraphics[width=.123\textwidth]{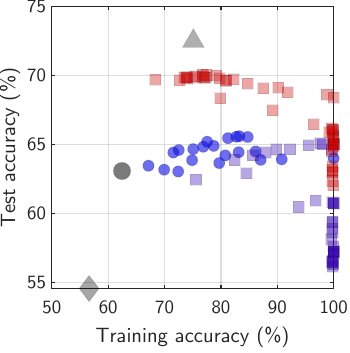}&
  \includegraphics[width=.123\textwidth]{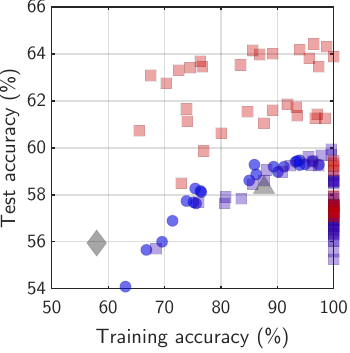}&
  \includegraphics[width=.123\textwidth]{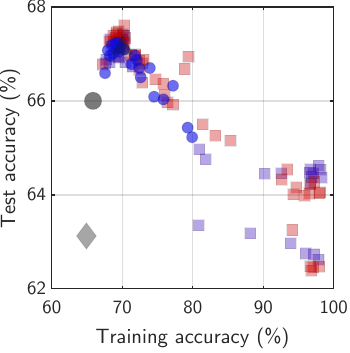}&
  \includegraphics[width=.123\textwidth]{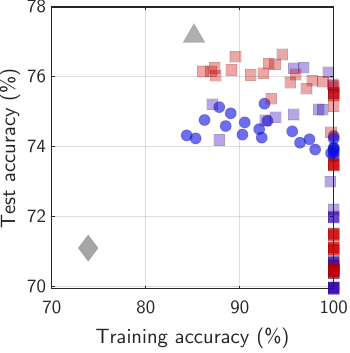}&
  \includegraphics[width=.123\textwidth]{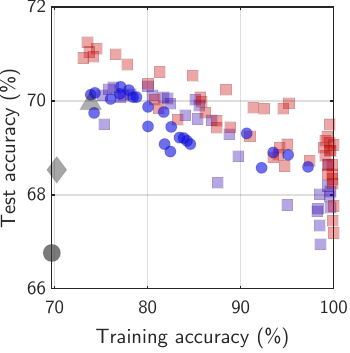}&
  \includegraphics[width=.123\textwidth]{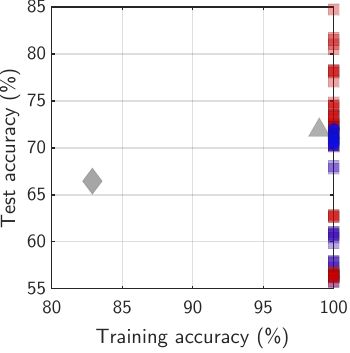}&
  \includegraphics[width=.123\textwidth]{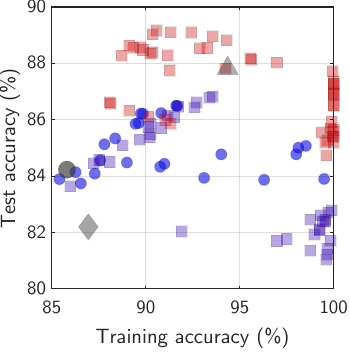}&
  \includegraphics[width=.123\textwidth]{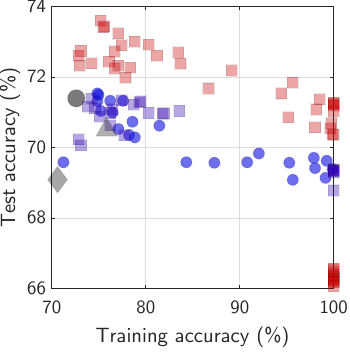}\\
  \end{tabularx}
  \vspace{-4pt}
  \caption{\label{fig:tabPlotsAll1}
    \textbf{Training vs.\ test accuracy} for all tabular datasets.
    The accuracy of ReLUs is generally surpassed
    by TanHs with the right prefactor.
    The accuracy is almost always best with the learned IAFs.
    As expected, these better models also show a clearly higher complexity.
    We also observe that the k-NN/trees/learned AFs models
    lie outside the pareto front of the ReLU models.
    In other words, they exhibit a different relation between training and test accuracy,
    which indicates that they clearly posses different inductive biases.
  }
\end{figure*}

\begin{figure*}[ht!]
  \centering
  \renewcommand{\tabcolsep}{0em}
  \renewcommand{\arraystretch}{1}
  \scriptsize
  \begin{tabularx}{\textwidth}{*{8}{>{\centering\arraybackslash}X}}
    \textsc{credit}&
    \textsc{electricity}&
    \textsc{covertype}&
    \textsc{pol}&
    \textsc{house16h}&
    \textsc{magictelescope}&
    \textsc{bankMarketing}&
    \textsc{miniboone}\\
  \includegraphics[width=.123\textwidth]{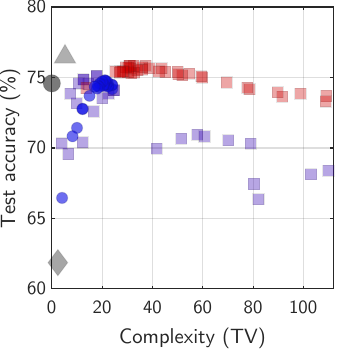}&
  \includegraphics[width=.123\textwidth]{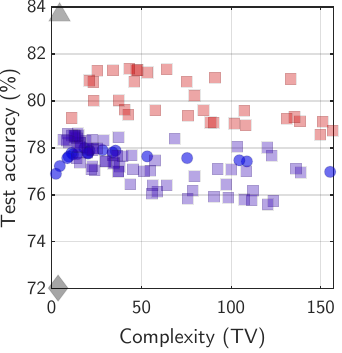}&
  \includegraphics[width=.123\textwidth]{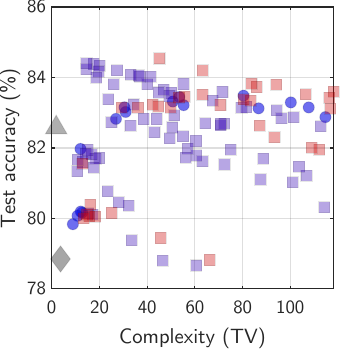}&
  \includegraphics[width=.123\textwidth]{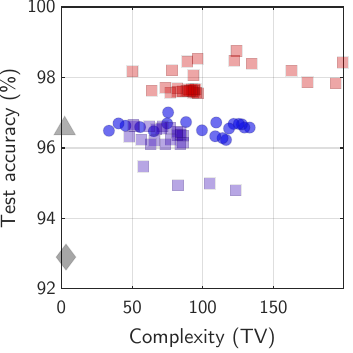}&
  \includegraphics[width=.123\textwidth]{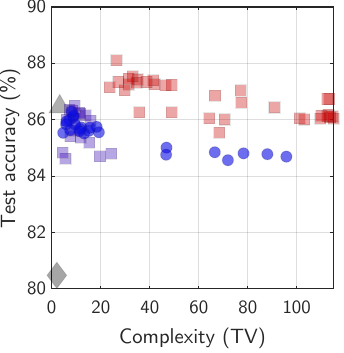}&
  \includegraphics[width=.123\textwidth]{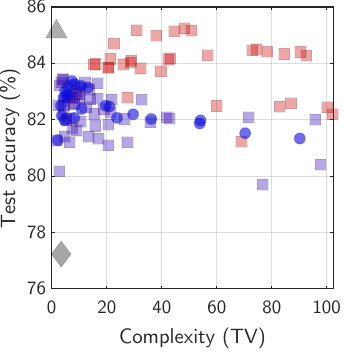}&
  \includegraphics[width=.123\textwidth]{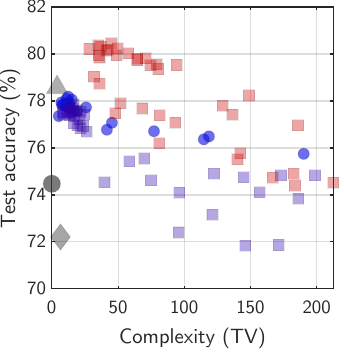}&
  \includegraphics[width=.123\textwidth]{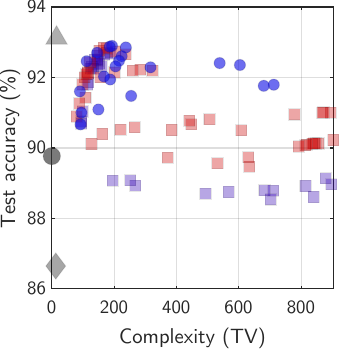}\\[5pt]
    \textsc{higgs}&
    \textsc{eyeMovements}&
    \textsc{diabetes130us}&
    \textsc{jannis}&
    \textsc{defaultOfCC}&
    \textsc{bioresponse}&
    \textsc{california}&
    \textsc{heloc}\\
  \includegraphics[width=.123\textwidth]{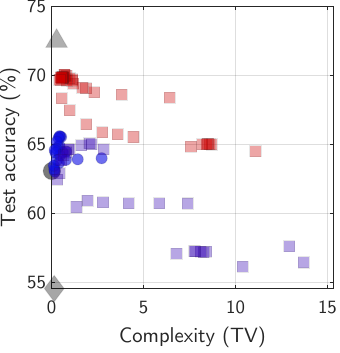}&
  \includegraphics[width=.123\textwidth]{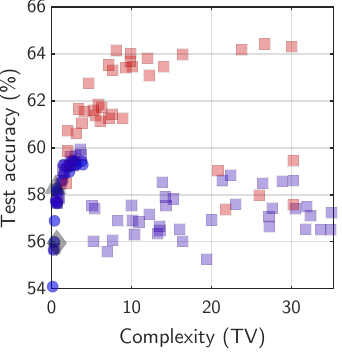}&
  \includegraphics[width=.123\textwidth]{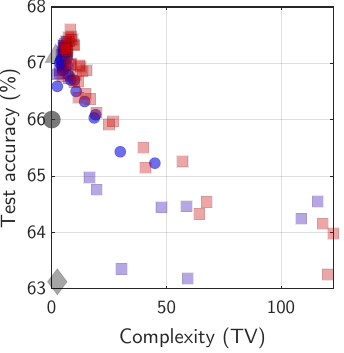}&
  \includegraphics[width=.123\textwidth]{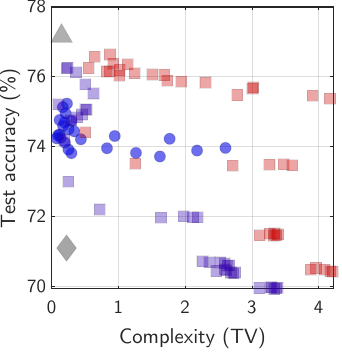}&
  \includegraphics[width=.123\textwidth]{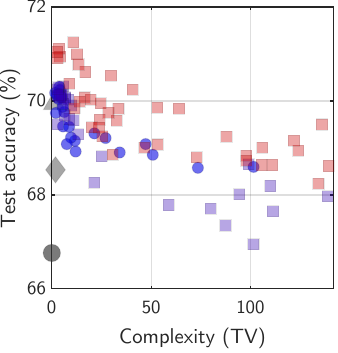}&
  \includegraphics[width=.123\textwidth]{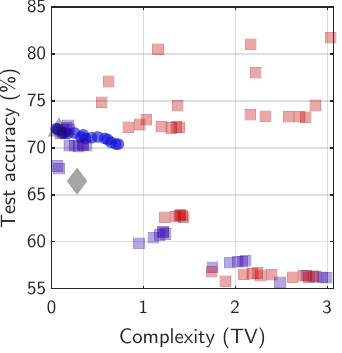}&
  \includegraphics[width=.123\textwidth]{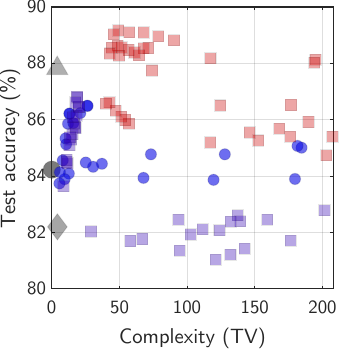}&
  \includegraphics[width=.123\textwidth]{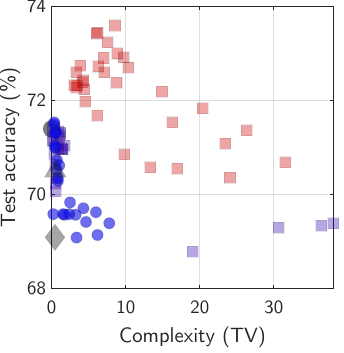}\\
  \end{tabularx}
  \vspace{-4pt}
  \caption{\label{fig:tabPlotsAll2}
  \textbf{Test accuracy vs.\ complexity} for all tabular datasets. As highlighted in \hyperref[fig:tabPlots]{Figure~\ref{fig:tabPlots}},
  the accuracy peaks at different complexity levels across datasets.
  This explains why \emph{dataset-specific} activation functions (and inductive biases)
  outperform the baselines.}
\end{figure*}

\paragraph{Additional results.}
See the figure below and their captions for details and observations.
In
\hyperref[fig:tabPlotsAll1]{Figures~\ref{fig:tabPlotsAll1}} and \ref{fig:tabPlotsAll2},
each marker represents a model with different hyperparameters, number of training steps,
and ReLUs\;%
(\textcolor[HTML]{5C5CFD}{\raisebox{0.0em}{\scalebox{0.75}[0.75]{\ding{108}}}})
or learned activations initilized as ReLUs\;%
(\textcolor[HTML]{B8A6E4}{\raisebox{0.0em}{\scalebox{0.75}[0.75]{\ding{110}}}})    
or as zeros\;%
(\textcolor[HTML]{EDA6A6}{\raisebox{0.0em}{\scalebox{0.75}[0.75]{\ding{110}}}}).
The k-NN and tree models are represented as
\textcolor[HTML]{AAAAAA}{\raisebox{0.0em}{\scalebox{0.75}[1.1]{\ding{117}}}}
and
\textcolor[HTML]{AAAAAA}{\raisebox{0.0em}{\scalebox{0.75}[0.75]{\ding{115}}}}.

\begin{figure*}[ht!]
  \centering
  \includegraphics[width=.99\textwidth]{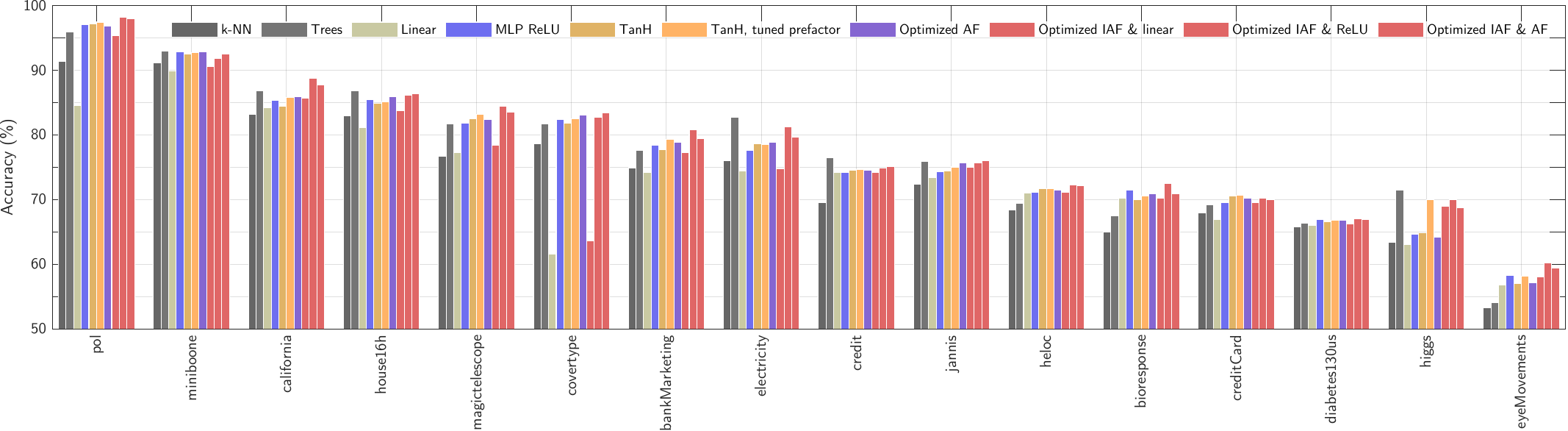}
  \vspace{-8pt}
  \caption{\label{fig:tabAllDatasets}
  Comparison of model types on every tabular dataset~\cite{grinTabBenchmark},
  approximately sorted by decreasing performance. 
  In almost all cases, the ReLU baseline is surpassed by optimized activation functions (TanH with prefactor, learned AFs, and learned IAFs).
  }
  \vspace{-4pt}
\end{figure*}

\setlength{\fboxrule}{0.2pt} 
\begin{figure*}[ht!]
\centering
\renewcommand{\tabcolsep}{.25em}
\renewcommand{\arraystretch}{1.5}
\scriptsize
\begin{tabularx}{\textwidth}{*{6}{>{\centering\arraybackslash}X}}

\makecell{
\includegraphics[clip, trim=2pt 11.1pt 6.5pt 0pt, height=.06\textwidth]{visualizations-tabCls2-ranged/af-file0001-100-1-te-sn0-0256-nTr0.10.png}%
 \reflectbox{\rotatebox[origin=c]{180}{\includegraphics[height=.06\textwidth]{visualizations-tabCls2-ranged/ll-file0001-100-1-te-sn0-0256-nTr0.10-vis1.png}}}%
}&
\makecell{
\includegraphics[clip, trim=2pt 11.1pt 6.5pt 0pt, height=.06\textwidth]{visualizations-tabCls2-ranged/af-file0002-189-1-te-sn0-0256-nTr0.10.png}%
\reflectbox{\rotatebox[origin=c]{180}{\includegraphics[height=.06\textwidth]{visualizations-tabCls2-ranged/ll-file0002-189-1-te-sn0-0256-nTr0.10-vis1.png}}}%
}&
\makecell{
\includegraphics[clip, trim=2pt 11.1pt 6.5pt 0pt, height=.06\textwidth]{visualizations-tabCls2-ranged/af-file0004-48-1-va-sn0-nTr1.00.png}%
\reflectbox{\rotatebox[origin=c]{180}{\includegraphics[height=.06\textwidth]{visualizations-tabCls2-ranged/ll-file0004-148-1-va-sn0-0256-nTr0.10-vis1.png}}}%
}& 
\makecell{
\includegraphics[clip, trim=2pt 11.1pt 6.5pt 0pt, height=.06\textwidth]{visualizations-tabCls2-ranged/af-file0005-35-1-va-sn0-nTr1.00}%
\reflectbox{\rotatebox[origin=c]{180}{\includegraphics[height=.06\textwidth]{visualizations-tabCls2-ranged/ll-file0005-135-1-va-sn0-0256-nTr0.10-vis1.png}}}%
}&
\multicolumn{2}{l}{
\raisebox{1.7em}{\textcolor[HTML]{666666}{\textbf{$\leftarrow$~~Activation function \& loss landscape}}}
}\\[-2pt]

\makecell{
\includegraphics[height=.06\textwidth]{visualizations-tabCls2-ranged/cl-file0001-100-1-te-sn0-0256-nTr0.10-vis1-1.png}%
\includegraphics[height=.06\textwidth]{visualizations-tabCls2-ranged/cl-file0001-100-1-te-sn0-0256-nTr0.10-vis4-1.png}%
}&
\makecell{
\includegraphics[height=.06\textwidth]{visualizations-tabCls2-ranged/cl-file0002-189-1-te-sn0-0256-nTr0.10-vis1-1.png}%
\includegraphics[height=.06\textwidth]{visualizations-tabCls2-ranged/cl-file0002-189-1-te-sn0-0256-nTr0.10-vis4-1.png}%
}&
\makecell{
\includegraphics[height=.06\textwidth]{visualizations-tabCls2-ranged/cl-file0004-148-1-va-sn0-0256-nTr0.10-vis1-1.png}%
\includegraphics[height=.06\textwidth]{visualizations-tabCls2-ranged/cl-file0004-148-1-va-sn0-0256-nTr0.10-vis4-1.png}%
}&
\makecell{
\includegraphics[height=.06\textwidth]{visualizations-tabCls2-ranged/cl-file0005-135-1-va-sn0-0256-nTr0.10-vis1-1.png}%
\includegraphics[height=.06\textwidth]{visualizations-tabCls2-ranged/cl-file0005-135-1-va-sn0-0256-nTr0.10-vis4-1.png}%
}&
\multicolumn{2}{l}{
\raisebox{1.7em}{\textcolor[HTML]{666666}{\textbf{$\leftarrow$~~Complexity landscape}}}
}\\[-2pt]

\makecell{\includegraphics[height=.119\textwidth]{visualizations-tabCls2-ranged/gridSqr-file0001-100-1-te-sn0-0256-nTr0.10.png}}&
\makecell{\includegraphics[height=.119\textwidth]{visualizations-tabCls2-ranged/gridSqr-file0002-189-1-te-sn0-0256-nTr0.10.png}}&
\makecell{\includegraphics[height=.119\textwidth]{visualizations-tabCls2-ranged/gridSqr-file0004-148-1-va-sn0-0256-nTr0.10.png}}&
\makecell{\includegraphics[height=.119\textwidth]{visualizations-tabCls2-ranged/gridSqr-file0005-135-1-va-sn0-0256-nTr0.10.png}}&
\makecell{\includegraphics[height=.119\textwidth]{visualizations-tabCls2-ranged/gridSqr-file0007-532-0-te-sn0-nTr0.10.png}}&
\makecell{\includegraphics[height=.119\textwidth]{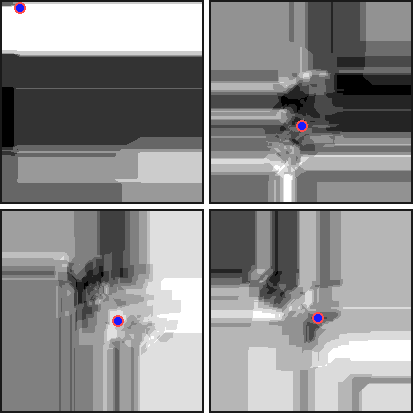}}
\\[12pt]

\makecell{
\includegraphics[clip, trim=2pt 11.1pt 6.5pt 0pt, height=.06\textwidth]{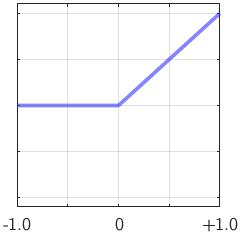}%
 \reflectbox{\rotatebox[origin=c]{180}{\includegraphics[height=.06\textwidth]{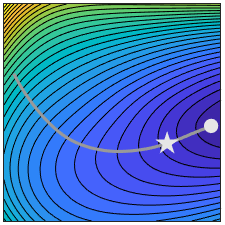}}}%
}&
\makecell{
\includegraphics[clip, trim=2pt 11.1pt 6.5pt 0pt, height=.06\textwidth]{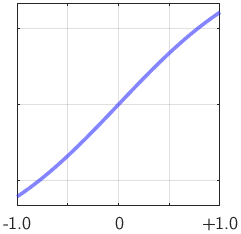}%
\reflectbox{\rotatebox[origin=c]{180}{\includegraphics[height=.06\textwidth]{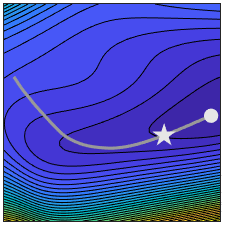}}}%
}&
\makecell{
\includegraphics[clip, trim=2pt 11.1pt 6.5pt 0pt, height=.06\textwidth]{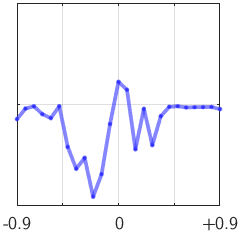}%
\reflectbox{\rotatebox[origin=c]{180}{\includegraphics[height=.06\textwidth]{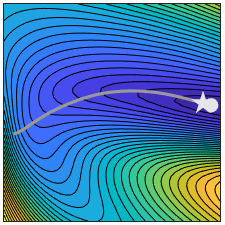}}}%
}& 
\makecell{
\includegraphics[clip, trim=2pt 11.1pt 6.5pt 0pt, height=.06\textwidth]{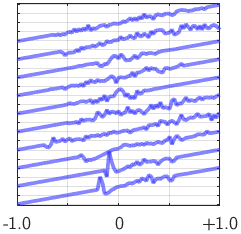}%
\reflectbox{\rotatebox[origin=c]{180}{\includegraphics[height=.06\textwidth]{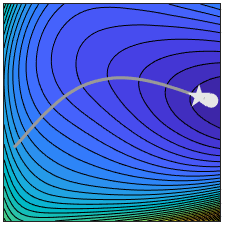}}}%
}&
\multicolumn{2}{l}{
\raisebox{1.7em}{\textcolor[HTML]{666666}{\textbf{$\leftarrow$~~Activation function \& loss landscape}}}
}\\[-2pt]

\makecell{
\includegraphics[height=.06\textwidth]{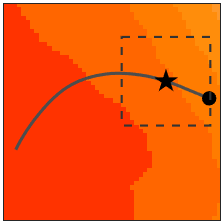}%
\includegraphics[height=.06\textwidth]{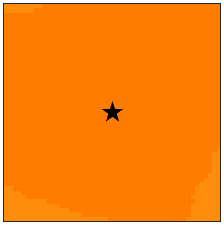}%
}&
\makecell{
\includegraphics[height=.06\textwidth]{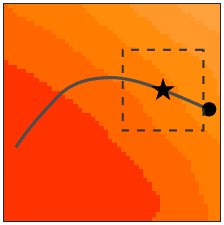}%
\includegraphics[height=.06\textwidth]{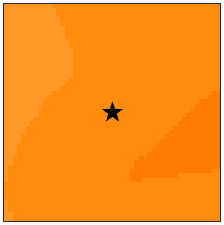}%
}&
\makecell{
\includegraphics[height=.06\textwidth]{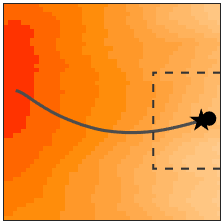}%
\includegraphics[height=.06\textwidth]{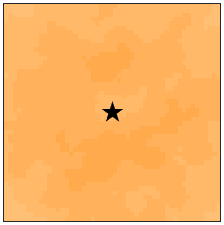}%
}&
\makecell{
\includegraphics[height=.06\textwidth]{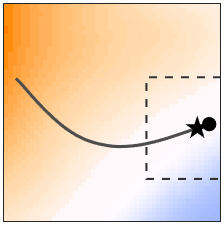}%
\includegraphics[height=.06\textwidth]{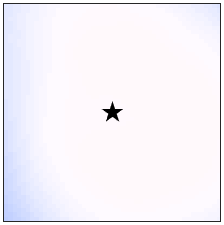}%
}&
\multicolumn{2}{l}{
\raisebox{1.7em}{\textcolor[HTML]{666666}{\textbf{$\leftarrow$~~Complexity landscape}}}
}\\[-2pt]

\makecell{\includegraphics[height=.119\textwidth]{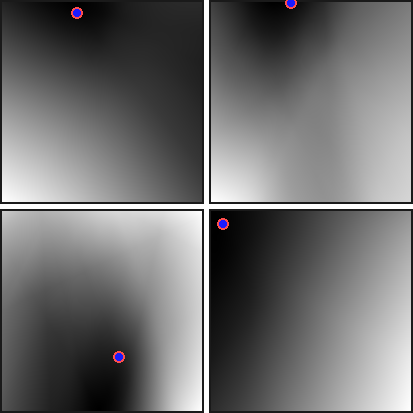}}&
\makecell{\includegraphics[height=.119\textwidth]{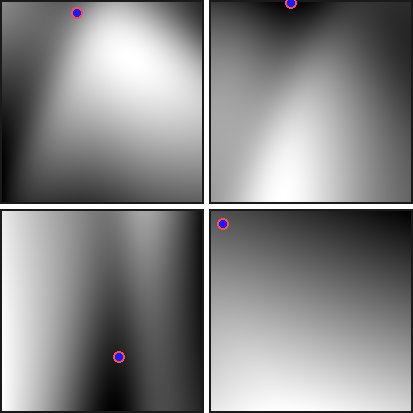}}&
\makecell{\includegraphics[height=.119\textwidth]{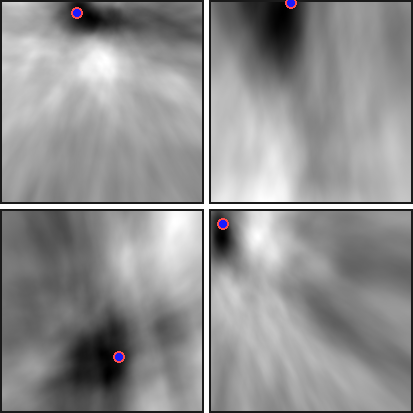}}&
\makecell{\includegraphics[height=.119\textwidth]{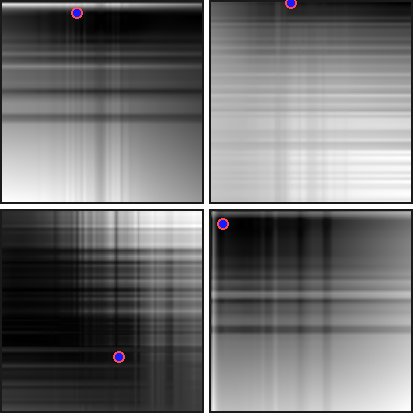}}&
\makecell{\includegraphics[height=.119\textwidth]{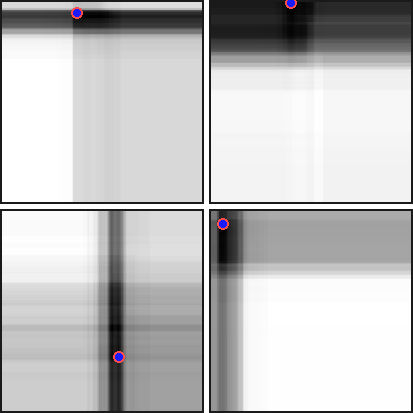}}&
\makecell{\includegraphics[height=.119\textwidth]{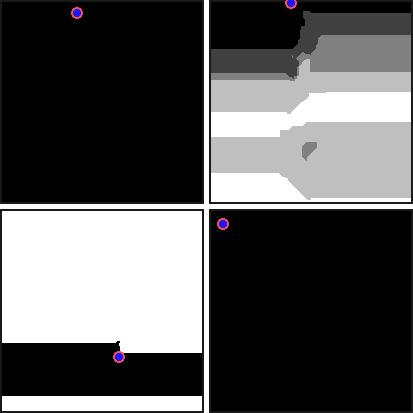}}
\\[12pt]

\makecell{
\includegraphics[clip, trim=2pt 11.1pt 6.5pt 0pt, height=.06\textwidth]{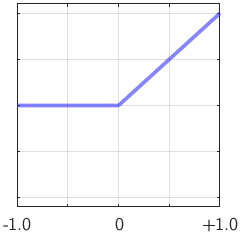}%
 \reflectbox{\rotatebox[origin=c]{180}{\includegraphics[height=.06\textwidth]{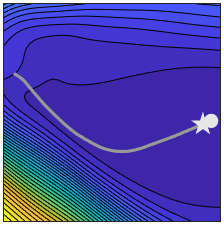}}}%
}&
\makecell{
\includegraphics[clip, trim=2pt 11.1pt 6.5pt 0pt, height=.06\textwidth]{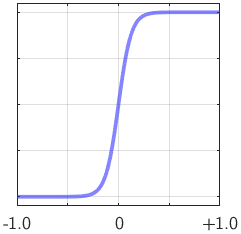}%
\reflectbox{\rotatebox[origin=c]{180}{\includegraphics[height=.06\textwidth]{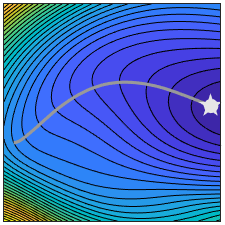}}}%
}&
\makecell{
\includegraphics[clip, trim=2pt 11.1pt 6.5pt 0pt, height=.06\textwidth]{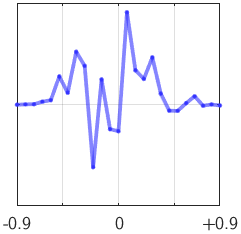}%
\reflectbox{\rotatebox[origin=c]{180}{\includegraphics[height=.06\textwidth]{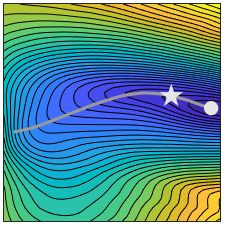}}}%
}& 
\makecell{
\includegraphics[clip, trim=2pt 11.1pt 6.5pt 0pt, height=.06\textwidth]{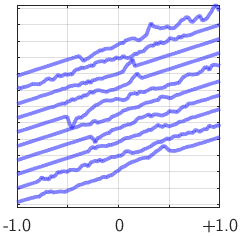}%
\reflectbox{\rotatebox[origin=c]{180}{\includegraphics[height=.06\textwidth]{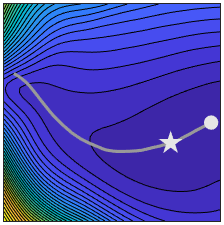}}}%
}&
\multicolumn{2}{l}{
\raisebox{1.7em}{\textcolor[HTML]{666666}{\textbf{$\leftarrow$~~Activation function \& loss landscape}}}
}\\[-2pt]

\makecell{
\includegraphics[height=.06\textwidth]{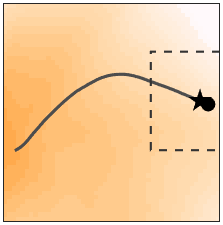}%
\includegraphics[height=.06\textwidth]{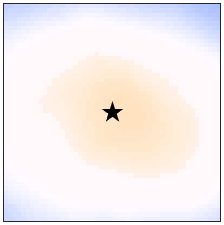}%
}&
\makecell{
\includegraphics[height=.06\textwidth]{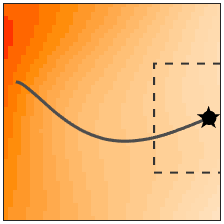}%
\includegraphics[height=.06\textwidth]{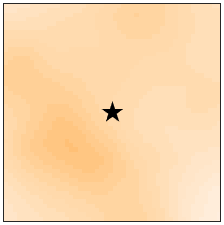}%
}&
\makecell{
\includegraphics[height=.06\textwidth]{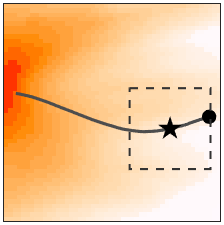}%
\includegraphics[height=.06\textwidth]{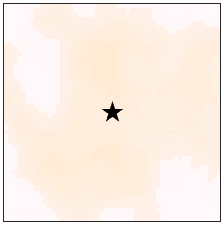}%
}&
\makecell{
\includegraphics[height=.06\textwidth]{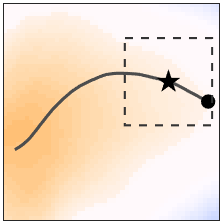}%
\includegraphics[height=.06\textwidth]{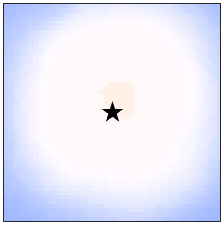}%
}&
\multicolumn{2}{l}{
\raisebox{1.7em}{\textcolor[HTML]{666666}{\textbf{$\leftarrow$~~Complexity landscape}}}
}\\[-2pt]

\makecell{\includegraphics[height=.119\textwidth]{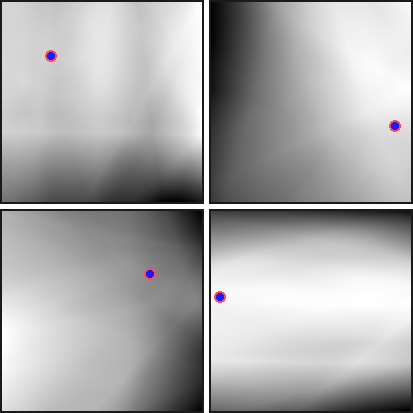}}&
\makecell{\includegraphics[height=.119\textwidth]{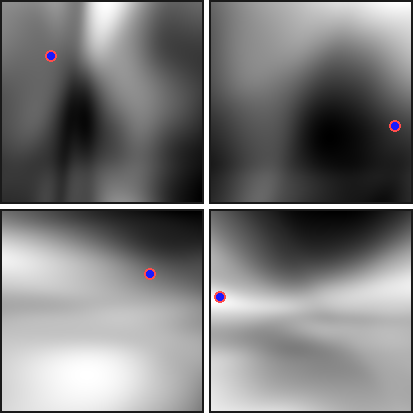}}&
\makecell{\includegraphics[height=.119\textwidth]{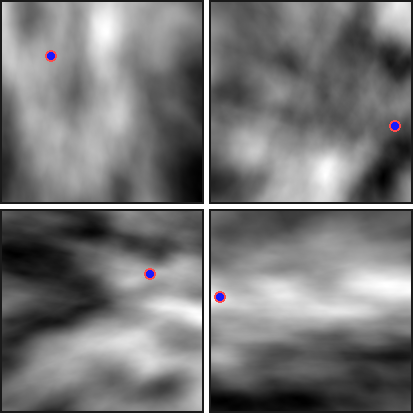}}&
\makecell{\includegraphics[height=.119\textwidth]{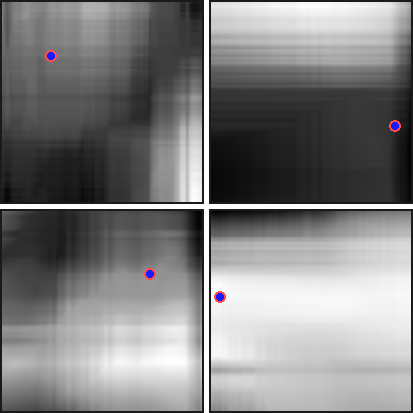}}&
\makecell{\includegraphics[height=.119\textwidth]{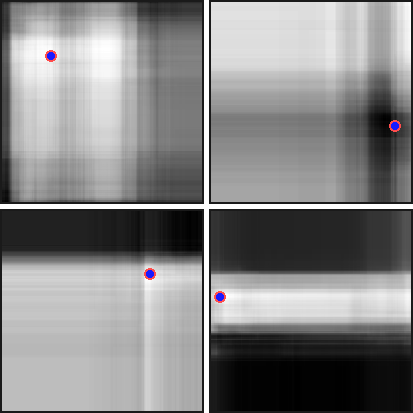}}&
\makecell{\includegraphics[height=.119\textwidth]{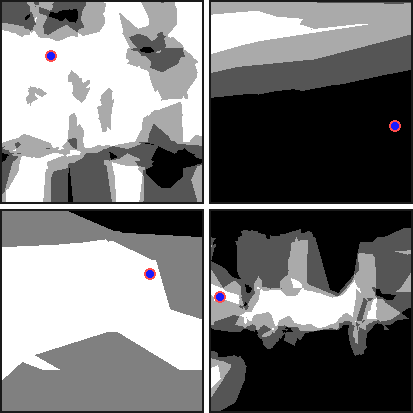}}
\\[0pt]

\textbf{MLP,~~ReLU}&
\textbf{MLP,~~TanH w/ prefactor}&
\textbf{MLP,~~learned AF}&
\textbf{MLP,~~learned IAFs}&
\textbf{Boosted decision trees}&
\textbf{k-NN}
\end{tabularx}
  \vspace{-8pt}
  \caption{\label{fig:tabBoundaries2}
  Models trained on three
  tabular datasets:
    \textsc{electricity}
    \textsc{magicTelescope}, and
    \textsc{coverType}~\cite{grinTabBenchmark}.
  See 
  \hyperref[fig:tabBoundaries]{Figure~\ref{fig:tabBoundaries}}
  in the main paper
  for details on the meaning of these visualizations.
  The observations are similar across datasets.
  }
  \vspace{-9pt}
\end{figure*}
\setlength{\fboxrule}{0.4pt} 

\clearpage

\subsection{Shortcut Learning}

\paragraph{Data.}
The collages datasets are built using all 10 classes from the original datasets.
This is a more difficult task that prior work~\cite{shah2020pitfalls,teney2021evading}
that only used 2 classes from each dataset.
Our training set uses random combinations of training images from the original dataset. Ditto for the validation and test sets. When no validation data is defined, we hold out a fraction of the training set of the same size as the test set.

\paragraph{Architecture.}
Our models are 1-layer fully-connected MLPs of width 32, trained with large-batch SGD (4096 examples per mini-batch) and a learning rate of 0.01.
Only the activation function varies across experiments.

\begin{figure*}[h!]
  \centering
  \includegraphics[clip, trim=0pt 0pt 225pt 0pt, height=.15\columnwidth]{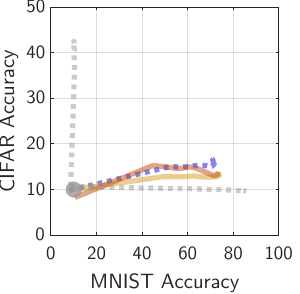}~~
  \includegraphics[height=.15\columnwidth]{visualizations-collages/plot-sn2.pdf}
  \vspace{-4pt}
  \caption{\label{fig:shortcut2}
  Experiments on shortcut learning (\mnist/\cifar collages) \textbf{without (left)} and \textbf{with (right)} spectral normalization. The training trajectory with the activation optimized for \cifar clearly differs from the baseline when using spectral normalization.}
\end{figure*}

\paragraph{Spectral normalization.}
Our most successful experiments on shortcut learning use \textbf{spectral normalization}~\cite{gogianu2021spectral,rosca2020case} on all layers when training and using the learned activation functions.
The motivation comes from~\citet{teney2024neural} who showed that the magnitude of the weights in a layer, together with the choice of activation function, influences the level of ``preferred complexity'' of the network.
We therefore hypothesized that the level of ``preferred complexity'' of a learned activation
would be more stable (invariant to weight magnitudes) if these could be constrained in a narrow range.
Spectral normalization is one way to constrain the magnitude of the linear transformation.
We compare in \hyperref[fig:shortcut2]{Figure~\ref{fig:shortcut2}} the same experiments performed without and with spectral normalization. We see that the ability of the learned activations to steer the model is slightly better
with spectral normalization (clearer differences in the training trajectories).

\paragraph{Additional results.}
We repeat our experiments with collages made from \mnist/\svhn. The training trajectories are not as distinct as with \mnist/\cifar, but the models do also achieve different top accuracies on the two datasets.

\begin{figure*}[h!]
  \renewcommand{\tabcolsep}{0.2em}
  \renewcommand{\arraystretch}{1}
  \centering
  \begin{tabular}{ccccc}
    \makecell[b]{
    \includegraphics[height=.0365\textwidth]{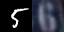}\\[-2pt]
    \includegraphics[height=.0365\textwidth]{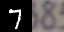}\\[3.5pt]
    \footnotesize Ambiguous\\[-3pt]\footnotesize training images}&
    \makecell[b]{
    \includegraphics[height=.084\textwidth]{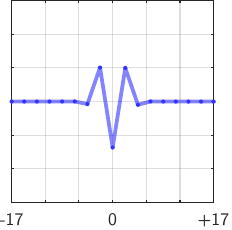}\\
    \footnotesize Optimized\\[-3pt]\footnotesize for \mnist}&
    \makecell[b]{
    \includegraphics[height=.084\textwidth]{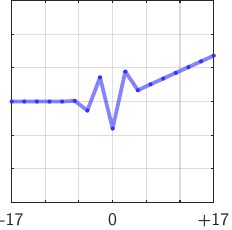}\\
    \footnotesize Optimized\\[-3pt]\footnotesize for \svhn}&
    \makecell[br]{\includegraphics[height=.12\textwidth]{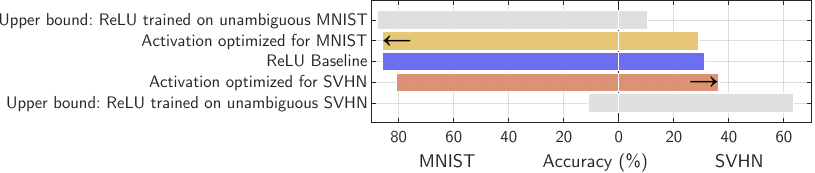}\vspace{-1pt}}&
    \makecell[br]{\includegraphics[clip, trim=0pt 0pt 225pt 0pt, height=.12\textwidth]{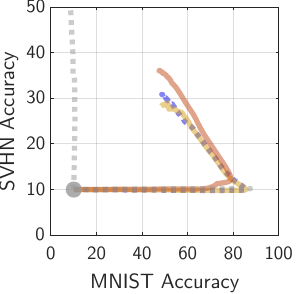}\vspace{0.02pt}}
  \end{tabular}\vspace{-4pt}
  \caption{\label{fig:shortcut3}
  Experiments on shortcut learning with \mnist/\svhn collages.
  Similar effects are obtained as with \mnist/\cifar (\hyperref[fig:shortcut]{Figure~\ref{fig:shortcut}}).
  }
\end{figure*}

\clearpage
\onecolumn
\subsection{Algorithmic Tasks and Grokking}
\label{detailsGrokking}

\paragraph{Data.}
We visualize in \hyperref[fig:grokking]{Figure~\ref{fig:grokking}}
the target functions of the algorithmic tasks used to investigate grokking
used in prior work~\cite{power2022grokking}.
Each axis of the visualizations corresponds to one of the two discrete-valued operands.
Grayscale values correspond to the target function's output, scaled to fit within the
black~$\rightarrow$~white range.
From the point of view of a network, operands and output are represented
as one-hot vectors. For example, for the task
$a\!+\!b \modulo 27$, each operands can take 27 different values.
Each is represented by a one-hot vector of length 27. The two are concatenated such that the input to the network is a vector of size
$2\!\times\!27\,{=}\,54$. The output of the network is a classification
over the $27$ possible values.
For every task, we generate all possible data (i.e.\ every combination of values of the two operands)
and make random 80/20 training/test splits.

\paragraph{Architecture.}
All networks in this section are 1-hidden layer MLPs of width 256,
trained with an MSE loss and large-batch ($4096$) gradient descent, no weight decay, learning rate of $1.0$, for max.\ $6e4$ training steps.

\begin{figure*}[bht!]
  \centering
  \renewcommand{\tabcolsep}{0em}
  \renewcommand{\arraystretch}{1}
  \begin{tabularx}{\textwidth}{*{11}{>{\centering\arraybackslash}X}}
    \makecell{\scriptsize $a\!+\!b$\\[-3pt]\scriptsize$(\modulo 27)$}&
    \makecell{\scriptsize $a\!-\!b$\\[-3pt]\scriptsize$(\modulo 27)$}&
    \makecell{\scriptsize $ab$\\[-3pt]\scriptsize$(\modulo 27)$}&
    \makecell{\scriptsize If $b$ odd, $ab$\\[-3pt]
                 \scriptsize else $a\!+\!b$}&
    \makecell{\scriptsize $a^2 + b^2$\\[-3pt]\scriptsize$(\modulo 27)$}&
    \makecell{\scriptsize $a^2 + ab + b^2$\\[-3pt]\scriptsize$(\modulo 27)$}&
    \makecell{\scriptsize $a^2 + ab + b^2$\\[-3pt]\scriptsize$+ a~(\modulo 27)$}&
    \makecell{\scriptsize $a^3 + ab$\\[-3pt]\scriptsize$(\modulo 27)$}&
    \makecell{\scriptsize $a.b$\\[-3pt]\scriptsize in $S_4$}&
    \makecell{\scriptsize $a.b.a$\\[-3pt]\scriptsize in $S_4$}&
    \makecell{\scriptsize $ a.b.a^{-1}$\\[-3pt]\scriptsize in $S_4$}\\[10pt]
    \makecell{\includegraphics[height=.071\textwidth]{visualizations-grok/grok-a27-0.8.png}}&
    \makecell{\includegraphics[height=.071\textwidth]{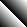}}&
    \makecell{\includegraphics[height=.071\textwidth]{visualizations-grok/grok-c27-0.8.png}}&
    \makecell{\includegraphics[height=.071\textwidth]{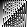}}&
    \makecell{\includegraphics[height=.071\textwidth]{visualizations-grok/grok-e27-0.8.png}}&
    \makecell{\includegraphics[height=.071\textwidth]{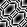}}&
    \makecell{\includegraphics[height=.071\textwidth]{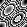}}&
    \makecell{\includegraphics[height=.071\textwidth]{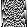}}&
    \makecell{\includegraphics[height=.071\textwidth]{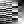}}&
    \makecell{\includegraphics[height=.071\textwidth]{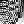}}&
    \makecell{\includegraphics[height=.071\textwidth]{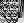}}\\[1pt]
    \makecell{\scriptsize$(\modulo 53)$}&
    \makecell{\scriptsize$(\modulo 53)$}&
    \makecell{\scriptsize$(\modulo 53)$}&
    \makecell{\scriptsize$(\modulo 53)$}&
    \makecell{\scriptsize$(\modulo 53)$}&
    \makecell{\scriptsize$(\modulo 53)$}&
    \makecell{\scriptsize$(\modulo 53)$}&
    \makecell{\scriptsize$(\modulo 53)$}&
    \makecell{\scriptsize in $S_5$}&
    \makecell{\scriptsize in $S_5$}&
    \makecell{\scriptsize in $S_5$}\\[0pt]
    \makecell{\includegraphics[height=.071\textwidth]{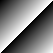}}&
    \makecell{\includegraphics[height=.071\textwidth]{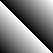}}&
    \makecell{\includegraphics[height=.071\textwidth]{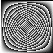}}&
    \makecell{\includegraphics[height=.071\textwidth]{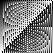}}&
    \makecell{\includegraphics[height=.071\textwidth]{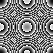}}&
    \makecell{\includegraphics[height=.071\textwidth]{visualizations-grok/grok-f53-0.9.png}}&
    \makecell{\includegraphics[height=.071\textwidth]{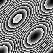}}&
    \makecell{\includegraphics[height=.071\textwidth]{visualizations-grok/grok-h53-0.9.png}}&
    \makecell{\includegraphics[height=.071\textwidth]{visualizations-grok/grok-i120-0.8.png}}&
    \makecell{\includegraphics[height=.071\textwidth]{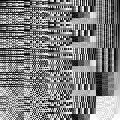}}&
    \makecell{\includegraphics[height=.071\textwidth]{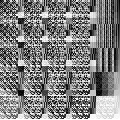}}\\
  \end{tabularx}
  \caption{\label{fig:grokkingData}
  Algorithmic tasks used to investigate grokking, also used by \citet{power2022grokking}.
  Each task is defined as an operation over two discrete-valued operands,
  passed to the model as 
  one-hot encodings.
  We visualize the target function of each task by plotting its value
  over all possible values of the operands (corresponding to the X/Y axes of each image).
  $S_n$ is the group of permutations of $n$ elements ($|S_4|\!=\!24$, $|S_5|\!=\!120$).
  }
\end{figure*}

\begin{figure*}[pt!]
  \centering
  \renewcommand{\tabcolsep}{0em}
  \renewcommand{\arraystretch}{1.3}
  \begin{tabularx}{\textwidth}{*{2}{>{\centering\arraybackslash}X}}
\makecell{
    \begin{overpic}[height=.09\textwidth]{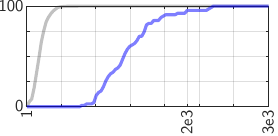}
        \put(54, 52){\makebox(0,0){{\scriptsize{ReLU}}}}
    \end{overpic}%
    \begin{overpic}[height=.09\textwidth]{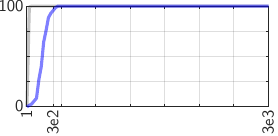}
        \put(54, 52){\makebox(0,0){{\scriptsize{Learned activation function}}}}
    \end{overpic}
    \raisebox{.018\textwidth}{\includegraphics[height=.068\textwidth]{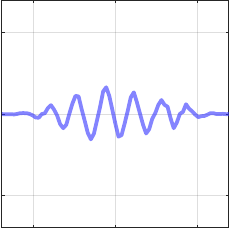}}
}
&
\makecell{
    \begin{overpic}[height=.09\textwidth]{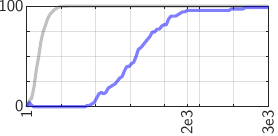}
        \put(54, 52){\makebox(0,0){{\scriptsize{ReLU}}}}
    \end{overpic}%
    \begin{overpic}[height=.09\textwidth]{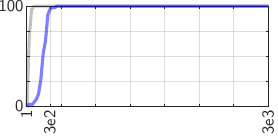}
        \put(54, 52){\makebox(0,0){{\scriptsize{Learned activation function}}}}
    \end{overpic}
    \raisebox{.018\textwidth}{\includegraphics[height=.068\textwidth]{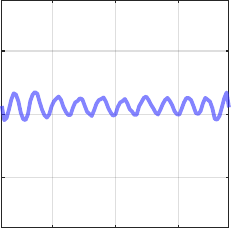}}
}
\\
\makecell{
    \begin{overpic}[height=.09\textwidth]{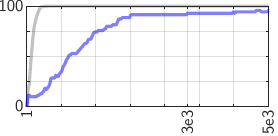}
    \end{overpic}%
    \begin{overpic}[height=.09\textwidth]{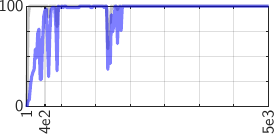}
    \end{overpic}
    \raisebox{.018\textwidth}{\includegraphics[height=.068\textwidth]{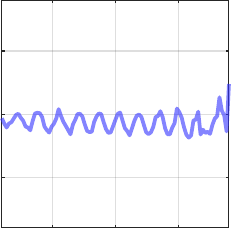}}
}
&
\makecell{
    \begin{overpic}[height=.09\textwidth]{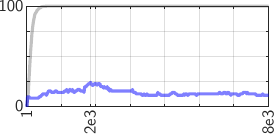}
    \end{overpic}%
    \begin{overpic}[height=.09\textwidth]{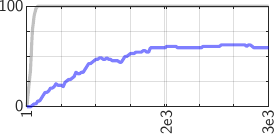}
    \end{overpic}
    \raisebox{.018\textwidth}{\includegraphics[height=.068\textwidth]{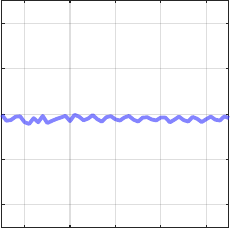}}
}
\\
\makecell{
    \begin{overpic}[height=.09\textwidth]{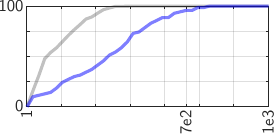}
    \end{overpic}%
    \begin{overpic}[height=.09\textwidth]{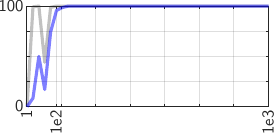}
    \end{overpic}
    \raisebox{.018\textwidth}{\includegraphics[height=.068\textwidth]{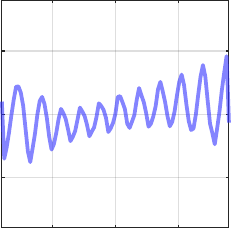}}
}
&
\makecell{
    \begin{overpic}[height=.09\textwidth]{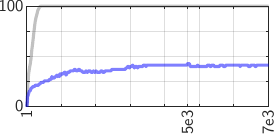}
    \end{overpic}%
    \begin{overpic}[height=.09\textwidth]{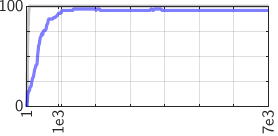}
    \end{overpic}
    \raisebox{.018\textwidth}{\includegraphics[height=.068\textwidth]{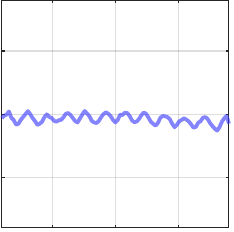}}
}
\\
\makecell{
    \begin{overpic}[height=.09\textwidth]{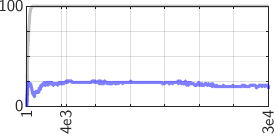}
    \end{overpic}%
    \begin{overpic}[height=.09\textwidth]{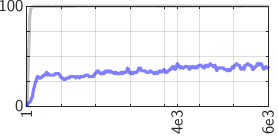}
    \end{overpic}
    \raisebox{.018\textwidth}{\includegraphics[height=.068\textwidth]{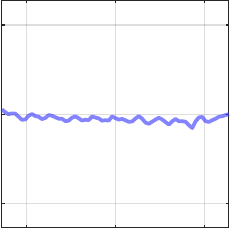}}
}
&
\makecell{
    \begin{overpic}[height=.09\textwidth]{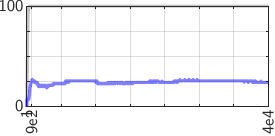}
    \end{overpic}%
    \begin{overpic}[height=.09\textwidth]{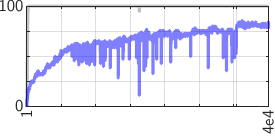}
    \end{overpic}
    \raisebox{.018\textwidth}{\includegraphics[height=.068\textwidth]{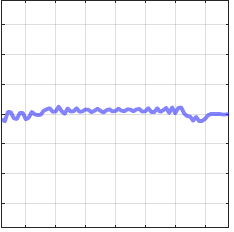}}
}
\\
\makecell{
    \begin{overpic}[height=.09\textwidth]{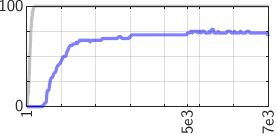}
    \end{overpic}%
    \begin{overpic}[height=.09\textwidth]{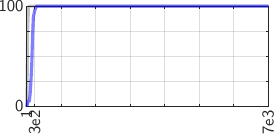}
    \end{overpic}
    \raisebox{.018\textwidth}{\includegraphics[height=.068\textwidth]{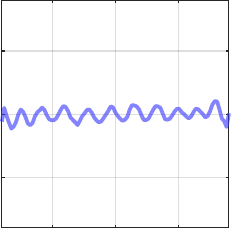}}
}
&
\makecell{
    \begin{overpic}[height=.09\textwidth]{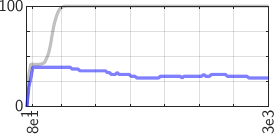}
    \end{overpic}%
    \begin{overpic}[height=.09\textwidth]{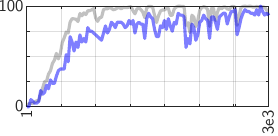}
    \end{overpic}
    \raisebox{.018\textwidth}{\includegraphics[height=.068\textwidth]{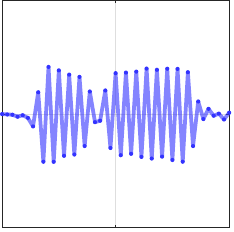}}
}
\\
\makecell{
    \begin{overpic}[height=.09\textwidth]{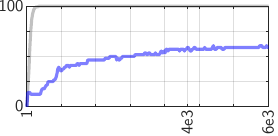}
    \end{overpic}%
    \begin{overpic}[height=.09\textwidth]{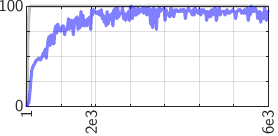}
    \end{overpic}
    \raisebox{.018\textwidth}{\includegraphics[height=.068\textwidth]{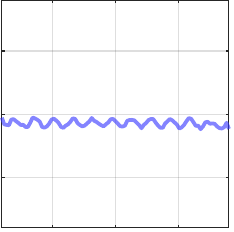}}
}
&
~~
\\
\makecell{
    \begin{overpic}[height=.09\textwidth]{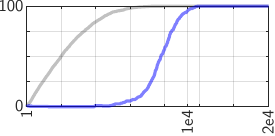}
    \end{overpic}%
    \begin{overpic}[height=.09\textwidth]{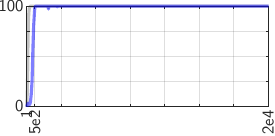}
    \end{overpic}
    \raisebox{.018\textwidth}{\includegraphics[height=.068\textwidth]{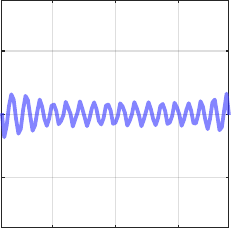}}
}
&
\makecell{
    \begin{overpic}[height=.09\textwidth]{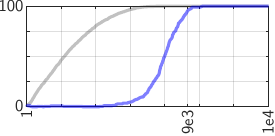}
    \end{overpic}%
    \begin{overpic}[height=.09\textwidth]{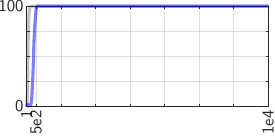}
    \end{overpic}
    \raisebox{.018\textwidth}{\includegraphics[height=.068\textwidth]{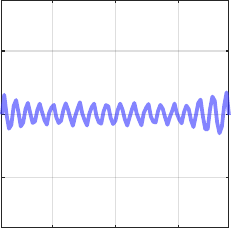}}
}
\\
\makecell{
    \begin{overpic}[height=.09\textwidth]{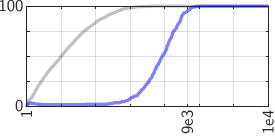}
    \end{overpic}%
    \begin{overpic}[height=.09\textwidth]{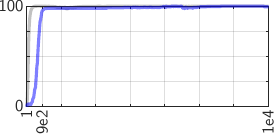}
    \end{overpic}
    \raisebox{.018\textwidth}{\includegraphics[height=.068\textwidth]{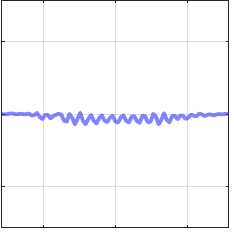}}
}
&
\makecell{
    \begin{overpic}[height=.09\textwidth]{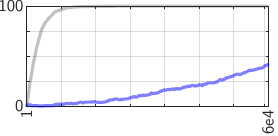}
    \end{overpic}%
    \begin{overpic}[height=.09\textwidth]{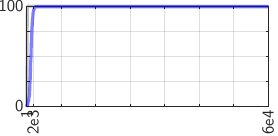}
    \end{overpic}
    \raisebox{.018\textwidth}{\includegraphics[height=.068\textwidth]{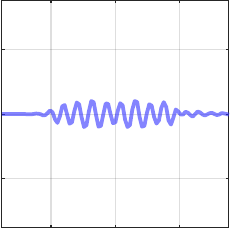}}
}
\\
\makecell{
    \begin{overpic}[height=.09\textwidth]{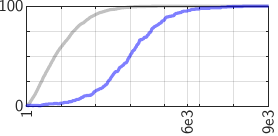}
    \end{overpic}%
    \begin{overpic}[height=.09\textwidth]{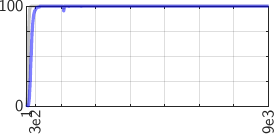}
    \end{overpic}
    \raisebox{.018\textwidth}{\includegraphics[height=.068\textwidth]{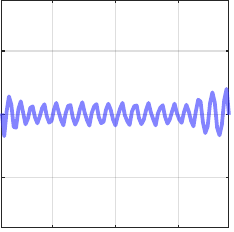}}
}
&
\makecell{
    \begin{overpic}[height=.09\textwidth]{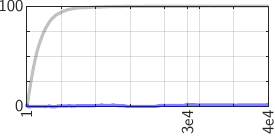}
    \end{overpic}%
    \begin{overpic}[height=.09\textwidth]{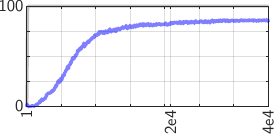}
    \end{overpic}
    \raisebox{.018\textwidth}{\includegraphics[height=.068\textwidth]{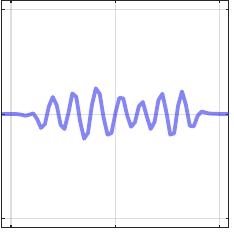}}
}
\\
\makecell{
    \begin{overpic}[height=.09\textwidth]{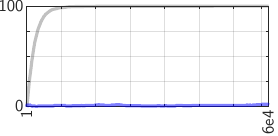}
    \end{overpic}%
    \begin{overpic}[height=.09\textwidth]{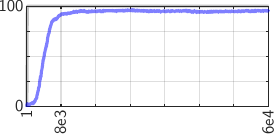}
    \end{overpic}
    \raisebox{.018\textwidth}{\includegraphics[height=.068\textwidth]{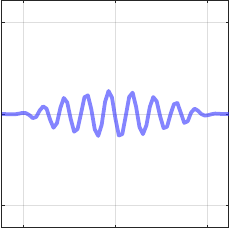}}
}
&
\makecell{
    \begin{overpic}[height=.09\textwidth]{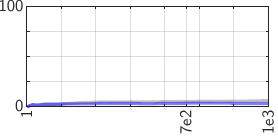}
    \end{overpic}%
    \begin{overpic}[height=.09\textwidth]{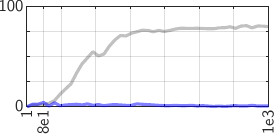}
    \end{overpic}
    \raisebox{.018\textwidth}{\includegraphics[height=.068\textwidth]{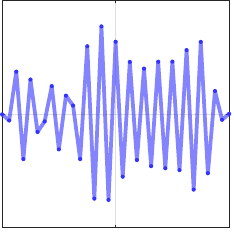}}
}
\\
\makecell{
    \begin{overpic}[height=.09\textwidth]{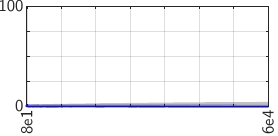}
    \end{overpic}%
    \begin{overpic}[height=.09\textwidth]{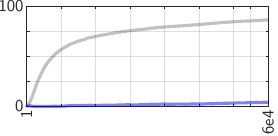}
    \end{overpic}
    \raisebox{.018\textwidth}{\includegraphics[height=.068\textwidth]{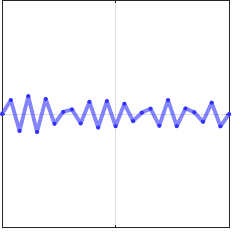}}
}
&
\makecell{
    \begin{overpic}[height=.09\textwidth]{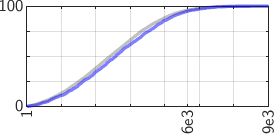}
    \end{overpic}%
    \begin{overpic}[height=.09\textwidth]{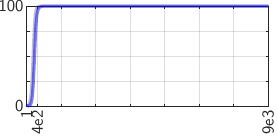}
    \end{overpic}
    \raisebox{.018\textwidth}{\includegraphics[height=.068\textwidth]{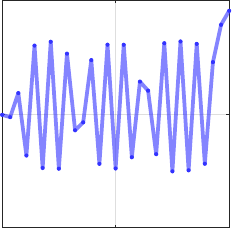}}
}
\\
\makecell{
    \begin{overpic}[height=.09\textwidth]{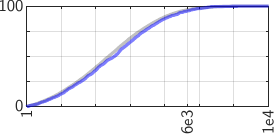}
    \end{overpic}%
    \begin{overpic}[height=.09\textwidth]{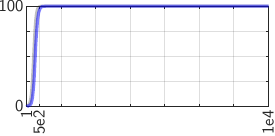}
    \end{overpic}
    \raisebox{.018\textwidth}{\includegraphics[height=.068\textwidth]{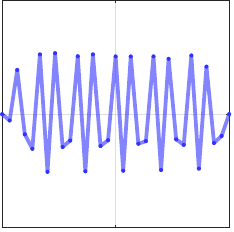}}
}
&
~~
\\
  \end{tabularx}
  \caption{\label{fig:grokkingFullResults}
  Training curves for all tasks from
  \hyperref[fig:grokkingData]{Figure~\ref{fig:grokkingData}}
  (same order, left-to-right then top-to-bottom).
  For each task, we show the accuracy across training iterations
  (\textcolor[HTML]{777777}{\raisebox{0.225em}{\scalebox{1.1}[0.125]{\ding{110}}}}
  training accuracy,
  \textcolor[HTML]{7F7FFF}{\raisebox{0.19em}{\scalebox{1.1}[0.25]{\ding{110}}}}
  test accuracy)
  for models with ReLUs and learned activations,
  and the learned activation function itself
  over $[-1,1]$.
  In almost all cases, the learned activation functions converge faster and/or to a higher test accuracy than ReLUs.
  }
\end{figure*}

\clearpage
\section{Measure of Complexity based on Total Variation}
\label{sec:tv}

\begin{figure}[h!]
  \centering
  \includegraphics[width=.25\columnwidth]{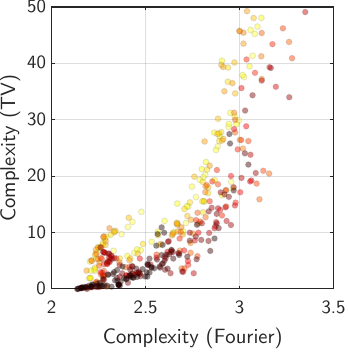}
  \vspace{-8pt}
  \caption{\label{fig:tvVsFourier}
  Comparison of the measure of complexity based on
  total variation (TV) 
  vs.\ a Fourier-based measure of complexity from prior work~\cite{teney2024neural}.
  We plot the values for a large number of models trained on the
  \textsc{electricity}
  tabular dataset~\cite{grinTabBenchmark}.
  The models use a TanH activation with a prefactor ranging from 0.1 to 8. The shade of the markers corresponds to the value of the prefactor (darker $\approx$ smaller).
  }
\end{figure}

\paragraph{Validation against Fourier complexity.}
To validate the proposed measure of complexity based on total variation (TV),
we compare it against a Fourier-based measure from prior work~\cite{teney2024neural}.
We plot the two for a large number of models in
\hyperref[fig:tvVsFourier]{Figure~\ref{fig:tvVsFourier}}.
They are very closely correlated. The TV is discriminative for both small and large values,
its evaluation is numerically more stable, and it is more straightforward to implement.
We made similar observations with other models and other datasets.

\paragraph{Implementation.}
The TV complexity involves a few implementation choices.
Most are not critical as long as
they are consistent across values being compared.
We provide precise hyperparameter values that we used
but they are easy to tune. One can simply evaluate the TV of some models multiple times
(with different random seeds for the choice of paths)
and verify that the variance is small.

\begin{figure}[h!]
  \centering
  \includegraphics[width=.7\columnwidth]{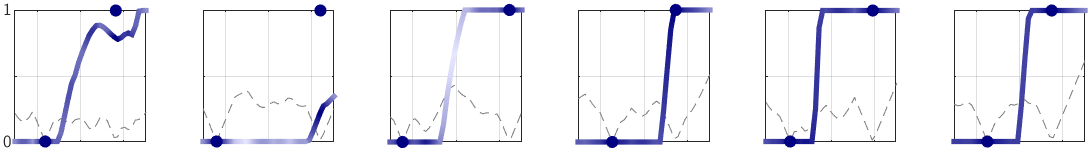}\\[4pt]
  \includegraphics[width=.7\columnwidth]{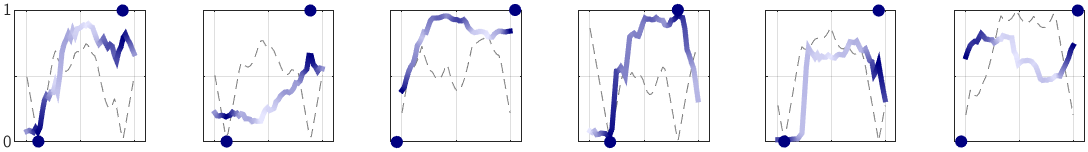}
  \vspace{-4pt}
  \caption{\label{fig:tvPaths}
  Examples of 1D paths (along the X axis) used to compute the TV complexity.
  The Y axis represents the model output. 
  These examples correspond to a model trained on a tabular classification dataset with ground truth labels in $\{0,1\}$.
  The blue dots
  (\textcolor[HTML]{000080}{\raisebox{-.02em}{\scalebox{1.35}[1.35]{$\bullet$}}})
  represent the paths' end points, which are training examples picked at random, and their ground truth values.
  The blue lines
  (\textcolor[HTML]{000080}{\raisebox{0.18em}{\scalebox{1.2}[0.3]{\ding{110}}}})
  represent the output 
  of the model (capped to $[0,1]$ for visualization).
  Note that this model does not perfectly interpolate the training points, i.e.\ the line does not always pass through the blue points.
  Dashed lines in the background 
  represent the distance to the closest point in the dataset, for debugging purposes.
  }
\end{figure}

\begin{itemize}[topsep=0pt,itemsep=1pt]
\item Number of linear paths: $200$. This simply needs to be high enough to probe the function along many dimensions.
\item Number of points on each path: $100$. This simply need to be high enough to capture the resolution of the variations in the function (see \hyperref[fig:tvPaths]{Figure~\ref{fig:tvPaths}}).
\item The two points defining each path are chosen as two points from the training set with \textbf{different labels}. One can also include points with the same label, but the path between them often is a constant line that does not bring any information.
\item We account for the fact that the function may not perfectly fit the ground truth values by first subtracting, from the evaluated path (blue line in \hyperref[fig:tvPaths]{Figure~\ref{fig:tvPaths}}),
the straight path connecting the predicted values at the two end points.
What we measure is therefore the \textbf{deviation from a piecewise linear model}.
Therefore, by design, the TV complexity of a linear model is~$0$.
\item Conceptually, it could make sense to normalize the TV of every path by the distance between its end points, because more variations could be expected along a longer path. In practice however, this would make no difference as long as the complexity values being compared are measured on the same set of paths/end points, or even just many paths from the same distribution of end points (i.e.\ the same dataset).
\end{itemize}


\end{document}